\documentclass[a4paper,11pt]{article}
\usepackage[T1]{fontenc}
\usepackage[utf8]{inputenc}
\usepackage[lining]{ebgaramond}
\usepackage{newtxtext,newtxmath}
\usepackage{graphicx,tikz,xcolor,hyperref,pifont}
\usepackage[margin=0pt,headheight=0pt,headsep=0pt,footskip=0pt]{geometry}
\hypersetup{colorlinks=true,urlcolor=black,linkcolor=black,citecolor=black}
\begin{document}
\setcounter{page}{83}
\begin{tikzpicture}[remember picture,overlay]
\node[anchor=base west,inner sep=0pt,outer sep=0pt] at ([xshift=127.580pt,yshift=-81.360pt]current page.north west) {{\fontsize{9.960pt}{10.558pt}\selectfont  \makebox[342.77pt][s]{International Journal on Cybernetics \& Informatics (IJCI) Vol.15, No.4, August 2026}}};
\node[anchor=base west,inner sep=0pt,outer sep=0pt] at ([xshift=82.104pt,yshift=-767.856pt]current page.north west) {{\fontsize{9.960pt}{10.558pt}\selectfont  DOI:\href{https://doi.org/10.5121/ijci.2026.15406}{10.5121/ijci.2026.15406}}};
\node[anchor=base west,inner sep=0pt,outer sep=0pt] at ([xshift=116.180pt,yshift=-130.220pt]current page.north west) {{\fontsize{20.040pt}{21.242pt}\selectfont \ebgaramond\bfseries\scshape S}};
\node[anchor=base west,inner sep=0pt,outer sep=0pt] at ([xshift=126.380pt,yshift=-130.220pt]current page.north west) {{\fontsize{15.960pt}{16.918pt}\selectfont \ebgaramond\bfseries\scshape ECURING}};
\node[anchor=base west,inner sep=0pt,outer sep=0pt] at ([xshift=207.290pt,yshift=-130.220pt]current page.north west) {{\fontsize{20.040pt}{21.242pt}\selectfont \ebgaramond\bfseries\scshape A}};
\node[anchor=base west,inner sep=0pt,outer sep=0pt] at ([xshift=220.490pt,yshift=-130.220pt]current page.north west) {{\fontsize{15.960pt}{16.918pt}\selectfont \ebgaramond\bfseries\scshape GENTIC}};
\node[anchor=base west,inner sep=0pt,outer sep=0pt] at ([xshift=289.130pt,yshift=-130.220pt]current page.north west) {{\fontsize{20.040pt}{21.242pt}\selectfont \ebgaramond\bfseries\scshape C}};
\node[anchor=base west,inner sep=0pt,outer sep=0pt] at ([xshift=302.810pt,yshift=-130.220pt]current page.north west) {{\fontsize{15.960pt}{16.918pt}\selectfont \ebgaramond\bfseries\scshape OMMERCE THROUGH}};
\node[anchor=base west,inner sep=0pt,outer sep=0pt] at ([xshift=114.500pt,yshift=-152.780pt]current page.north west) {{\fontsize{20.040pt}{21.242pt}\selectfont \ebgaramond\bfseries\scshape B}};
\node[anchor=base west,inner sep=0pt,outer sep=0pt] at ([xshift=128.060pt,yshift=-152.780pt]current page.north west) {{\fontsize{15.960pt}{16.918pt}\selectfont \ebgaramond\bfseries\scshape EHAVIOR}};
\node[anchor=base west,inner sep=0pt,outer sep=0pt] at ([xshift=203.570pt,yshift=-152.780pt]current page.north west) {{\fontsize{20.040pt}{21.242pt}\selectfont \ebgaramond\bfseries\scshape -A}};
\node[anchor=base west,inner sep=0pt,outer sep=0pt] at ([xshift=223.250pt,yshift=-152.780pt]current page.north west) {{\fontsize{15.960pt}{16.918pt}\selectfont \ebgaramond\bfseries\scshape WARE\ }};
\node[anchor=base west,inner sep=0pt,outer sep=0pt] at ([xshift=276.250pt,yshift=-152.780pt]current page.north west) {{\fontsize{20.040pt}{21.242pt}\selectfont \ebgaramond\bfseries\scshape F}};
\node[anchor=base west,inner sep=0pt,outer sep=0pt] at ([xshift=286.610pt,yshift=-152.780pt]current page.north west) {{\fontsize{15.960pt}{16.918pt}\selectfont \ebgaramond\bfseries\scshape RAUD\ }};
\node[anchor=base west,inner sep=0pt,outer sep=0pt] at ([xshift=340.790pt,yshift=-152.780pt]current page.north west) {{\fontsize{20.040pt}{21.242pt}\selectfont \ebgaramond\bfseries\scshape D}};
\node[anchor=base west,inner sep=0pt,outer sep=0pt] at ([xshift=356.510pt,yshift=-152.780pt]current page.north west) {{\fontsize{15.960pt}{16.918pt}\selectfont \ebgaramond\bfseries\scshape ETECTION AND}};
\node[anchor=base west,inner sep=0pt,outer sep=0pt] at ([xshift=121.340pt,yshift=-175.220pt]current page.north west) {{\fontsize{20.040pt}{21.242pt}\selectfont \ebgaramond\bfseries\scshape C}};
\node[anchor=base west,inner sep=0pt,outer sep=0pt] at ([xshift=134.900pt,yshift=-175.220pt]current page.north west) {{\fontsize{15.960pt}{16.918pt}\selectfont \ebgaramond\bfseries\scshape \makebox[166.81pt][s]{YBERSECURITY FOR}}};
\node[anchor=base west,inner sep=0pt,outer sep=0pt] at ([xshift=307.730pt,yshift=-175.220pt]current page.north west) {{\fontsize{20.040pt}{21.242pt}\selectfont \ebgaramond\bfseries\scshape T}};
\node[anchor=base west,inner sep=0pt,outer sep=0pt] at ([xshift=319.650pt,yshift=-175.220pt]current page.north west) {{\fontsize{15.960pt}{16.918pt}\selectfont \ebgaramond\bfseries\scshape RUSTED\ }};
\node[anchor=base west,inner sep=0pt,outer sep=0pt] at ([xshift=389.630pt,yshift=-175.220pt]current page.north west) {{\fontsize{20.040pt}{21.242pt}\selectfont \ebgaramond\bfseries\scshape AI-A}};
\node[anchor=base west,inner sep=0pt,outer sep=0pt] at ([xshift=430.670pt,yshift=-175.220pt]current page.north west) {{\fontsize{15.960pt}{16.918pt}\selectfont \ebgaramond\bfseries\scshape GENT}};
\node[anchor=base west,inner sep=0pt,outer sep=0pt] at ([xshift=182.300pt,yshift=-197.780pt]current page.north west) {{\fontsize{20.040pt}{21.242pt}\selectfont \ebgaramond\bfseries\scshape P}};
\node[anchor=base west,inner sep=0pt,outer sep=0pt] at ([xshift=193.250pt,yshift=-197.780pt]current page.north west) {{\fontsize{15.960pt}{16.918pt}\selectfont \ebgaramond\bfseries\scshape AYMENT}};
\node[anchor=base west,inner sep=0pt,outer sep=0pt] at ([xshift=268.370pt,yshift=-197.780pt]current page.north west) {{\fontsize{20.040pt}{21.242pt}\selectfont \ebgaramond\bfseries\scshape A}};
\node[anchor=base west,inner sep=0pt,outer sep=0pt] at ([xshift=281.210pt,yshift=-197.780pt]current page.north west) {{\fontsize{15.960pt}{16.918pt}\selectfont \ebgaramond\bfseries\scshape UTHORIZATION}};
\node[anchor=base west,inner sep=0pt,outer sep=0pt] at ([xshift=244.010pt,yshift=-231.260pt]current page.north west) {{\fontsize{12.960pt}{13.738pt}\selectfont  \makebox[110.54pt][s]{Kayode L. Ogunsusi}}};
\node[anchor=base west,inner sep=0pt,outer sep=0pt] at ([xshift=137.420pt,yshift=-258.980pt]current page.north west) {{\fontsize{12.000pt}{12.720pt}\selectfont  \makebox[323.60pt][s]{Credit and Fraud Risk, American Express, Phoenix, Arizona, USA}}};
\node[anchor=base west,inner sep=0pt,outer sep=0pt] at ([xshift=82.104pt,yshift=-286.490pt]current page.north west) {{\fontsize{12.960pt}{13.738pt}\selectfont \bfseries A}};
\node[anchor=base west,inner sep=0pt,outer sep=0pt] at ([xshift=91.464pt,yshift=-286.490pt]current page.north west) {{\fontsize{10.560pt}{11.194pt}\selectfont \bfseries BSTRACT}};
\node[anchor=base west,inner sep=0pt,outer sep=0pt] at ([xshift=82.104pt,yshift=-311.330pt]current page.north west) {{\fontsize{9.960pt}{10.558pt}\selectfont \itshape \makebox[433.23pt][s]{The emergence of autonomous artificial intelligence (AI) agents in digital commerce introduces}}};
\node[anchor=base west,inner sep=0pt,outer sep=0pt] at ([xshift=82.104pt,yshift=-322.730pt]current page.north west) {{\fontsize{9.960pt}{10.558pt}\selectfont \itshape \makebox[433.24pt][s]{cybersecurity risks involving delegated authority, agent impersonation, compromised agents, malicious}}};
\node[anchor=base west,inner sep=0pt,outer sep=0pt] at ([xshift=82.104pt,yshift=-334.250pt]current page.north west) {{\fontsize{9.960pt}{10.558pt}\selectfont \itshape \makebox[433.17pt][s]{delegation, and unauthorized payment execution. This study proposes the Autonomous Commerce Agent}}};
\node[anchor=base west,inner sep=0pt,outer sep=0pt] at ([xshift=82.104pt,yshift=-345.770pt]current page.north west) {{\fontsize{9.960pt}{10.558pt}\selectfont \itshape \makebox[433.45pt][s]{Guard System (ACAGS), a behavior-aware cybersecurity framework integrating supervised machine}}};
\node[anchor=base west,inner sep=0pt,outer sep=0pt] at ([xshift=82.104pt,yshift=-357.290pt]current page.north west) {{\fontsize{9.960pt}{10.558pt}\selectfont \itshape \makebox[433.41pt][s]{learning, historical behavioral analysis, agent-specific authorization controls, and explainable AI for}}};
\node[anchor=base west,inner sep=0pt,outer sep=0pt] at ([xshift=82.104pt,yshift=-368.810pt]current page.north west) {{\fontsize{9.960pt}{10.558pt}\selectfont \itshape \makebox[433.51pt][s]{payment authorization. ACAGS was evaluated using a large-scale synthetic financial transaction dataset}}};
\node[anchor=base west,inner sep=0pt,outer sep=0pt] at ([xshift=82.104pt,yshift=-380.330pt]current page.north west) {{\fontsize{9.960pt}{10.558pt}\selectfont \itshape \makebox[433.36pt][s]{extended with a simulated agentic-commerce layer representing agent identity, delegated spending limits,}}};
\node[anchor=base west,inner sep=0pt,outer sep=0pt] at ([xshift=82.104pt,yshift=-391.730pt]current page.north west) {{\fontsize{9.960pt}{10.558pt}\selectfont \itshape \makebox[433.25pt][s]{authorization scope, trust characteristics, and five adversarial scenarios. XGBoost was selected as the}}};
\node[anchor=base west,inner sep=0pt,outer sep=0pt] at ([xshift=82.104pt,yshift=-403.250pt]current page.north west) {{\fontsize{9.960pt}{10.558pt}\selectfont \itshape \makebox[433.21pt][s]{primary classifier following comparative evaluation of four supervised learning models. Across 10}}};
\node[anchor=base west,inner sep=0pt,outer sep=0pt] at ([xshift=82.104pt,yshift=-414.770pt]current page.north west) {{\fontsize{9.960pt}{10.558pt}\selectfont \itshape \makebox[433.66pt][s]{experimental runs, behavioral features increased mean F1-score from 52.12\% to 67.49\%, while the}}};
\node[anchor=base west,inner sep=0pt,outer sep=0pt] at ([xshift=82.104pt,yshift=-426.290pt]current page.north west) {{\fontsize{9.960pt}{10.558pt}\selectfont \itshape \makebox[433.52pt][s]{addition of agent-specific features further increased it to 69.46\%. ACAGS detected 99.77\% of abnormal}}};
\node[anchor=base west,inner sep=0pt,outer sep=0pt] at ([xshift=82.104pt,yshift=-437.810pt]current page.north west) {{\fontsize{9.960pt}{10.558pt}\selectfont \itshape \makebox[433.63pt][s]{agent behavior, 99.73\% of agent impersonation, 100\% of malicious delegation, and 100\% of scope}}};
\node[anchor=base west,inner sep=0pt,outer sep=0pt] at ([xshift=82.104pt,yshift=-449.210pt]current page.north west) {{\fontsize{9.960pt}{10.558pt}\selectfont \itshape \makebox[257.39pt][s]{violations, while compromised-agent detection reached 72.27\%}}};
\node[anchor=base west,inner sep=0pt,outer sep=0pt] at ([xshift=339.670pt,yshift=-449.210pt]current page.north west) {{\fontsize{9.960pt}{10.558pt}\selectfont \bfseries .}};
\node[anchor=base west,inner sep=0pt,outer sep=0pt] at ([xshift=342.190pt,yshift=-449.210pt]current page.north west) {{\fontsize{9.960pt}{10.558pt}\selectfont \itshape \makebox[173.42pt][s]{The authorization policy translated fraud}}};
\node[anchor=base west,inner sep=0pt,outer sep=0pt] at ([xshift=82.104pt,yshift=-460.730pt]current page.north west) {{\fontsize{9.960pt}{10.558pt}\selectfont \itshape \makebox[433.19pt][s]{and agent-policy risk into APPROVE, REVIEW, and REJECT outcomes, while SHAP provided}}};
\node[anchor=base west,inner sep=0pt,outer sep=0pt] at ([xshift=82.104pt,yshift=-472.270pt]current page.north west) {{\fontsize{9.960pt}{10.558pt}\selectfont \itshape \makebox[433.30pt][s]{interpretable fraud-risk explanations. These findings provide controlled experimental evidence that}}};
\node[anchor=base west,inner sep=0pt,outer sep=0pt] at ([xshift=82.104pt,yshift=-483.790pt]current page.north west) {{\fontsize{9.960pt}{10.558pt}\selectfont \itshape \makebox[433.45pt][s]{combining behavioral fraud detection with delegated-agent authorization controls can strengthen security}}};
\node[anchor=base west,inner sep=0pt,outer sep=0pt] at ([xshift=82.104pt,yshift=-495.310pt]current page.north west) {{\fontsize{9.960pt}{10.558pt}\selectfont \itshape \makebox[433.37pt][s]{assessment in simulated agentic-commerce payment environments, while validation using operational}}};
\node[anchor=base west,inner sep=0pt,outer sep=0pt] at ([xshift=82.104pt,yshift=-506.830pt]current page.north west) {{\fontsize{9.960pt}{10.558pt}\selectfont \itshape \makebox[191.64pt][s]{agent-initiated transactions remains necessary.}}};
\node[anchor=base west,inner sep=0pt,outer sep=0pt] at ([xshift=82.104pt,yshift=-534.910pt]current page.north west) {{\fontsize{12.960pt}{13.738pt}\selectfont \bfseries K}};
\node[anchor=base west,inner sep=0pt,outer sep=0pt] at ([xshift=92.184pt,yshift=-534.910pt]current page.north west) {{\fontsize{10.560pt}{11.194pt}\selectfont \bfseries EYWORDS}};
\node[anchor=base west,inner sep=0pt,outer sep=0pt] at ([xshift=82.104pt,yshift=-559.630pt]current page.north west) {{\fontsize{9.960pt}{10.558pt}\selectfont \itshape \makebox[433.15pt][s]{Agentic Commerce, Cybersecurity, Fraud Detection, Payment Authorization, Autonomous AI Agents,}}};
\node[anchor=base west,inner sep=0pt,outer sep=0pt] at ([xshift=82.104pt,yshift=-571.150pt]current page.north west) {{\fontsize{9.960pt}{10.558pt}\selectfont \itshape \makebox[233.28pt][s]{Explainable AI, Behavioral Analytics, Machine Learning.}}};
\node[anchor=base west,inner sep=0pt,outer sep=0pt] at ([xshift=82.104pt,yshift=-600.190pt]current page.north west) {{\fontsize{14.040pt}{14.882pt}\selectfont \bfseries 1.}};
\node[anchor=base west,inner sep=0pt,outer sep=0pt] at ([xshift=100.100pt,yshift=-600.190pt]current page.north west) {{\fontsize{14.040pt}{14.882pt}\selectfont \bfseries I}};
\node[anchor=base west,inner sep=0pt,outer sep=0pt] at ([xshift=105.620pt,yshift=-600.190pt]current page.north west) {{\fontsize{11.040pt}{11.702pt}\selectfont \bfseries NTRODUCTIONS}};
\node[anchor=base west,inner sep=0pt,outer sep=0pt] at ([xshift=82.104pt,yshift=-626.110pt]current page.north west) {{\fontsize{11.040pt}{11.702pt}\selectfont  \makebox[431.43pt][s]{The rapid advancement of artificial intelligence is transforming digital commerce from human-}}};
\node[anchor=base west,inner sep=0pt,outer sep=0pt] at ([xshift=82.104pt,yshift=-638.710pt]current page.north west) {{\fontsize{11.040pt}{11.702pt}\selectfont  \makebox[433.83pt][s]{assisted online transactions to agentic commerce, where autonomous software agents can}}};
\node[anchor=base west,inner sep=0pt,outer sep=0pt] at ([xshift=82.104pt,yshift=-651.430pt]current page.north west) {{\fontsize{11.040pt}{11.702pt}\selectfont  \makebox[433.88pt][s]{discover products, negotiate prices, make purchasing decisions, andexecute payment transactions}}};
\node[anchor=base west,inner sep=0pt,outer sep=0pt] at ([xshift=82.104pt,yshift=-664.060pt]current page.north west) {{\fontsize{11.040pt}{11.702pt}\selectfont  \makebox[235.83pt][s]{on behalf of users with minimal human intervention}}};
\node[anchor=base west,inner sep=0pt,outer sep=0pt] at ([xshift=318.790pt,yshift=-664.060pt]current page.north west) {{\fontsize{11.040pt}{11.702pt}\selectfont \bfseries [}};
\node[anchor=base west,inner sep=0pt,outer sep=0pt] at ([xshift=322.390pt,yshift=-664.060pt]current page.north west) {{\fontsize{11.040pt}{11.702pt}\selectfont  \makebox[193.55pt][s]{1]--[3]. This paradigm promises significant}}};
\node[anchor=base west,inner sep=0pt,outer sep=0pt] at ([xshift=82.104pt,yshift=-676.660pt]current page.north west) {{\fontsize{11.040pt}{11.702pt}\selectfont  \makebox[433.64pt][s]{improvements in operational efficiency, personalized customer experiences, and transaction}}};
\node[anchor=base west,inner sep=0pt,outer sep=0pt] at ([xshift=82.104pt,yshift=-689.380pt]current page.north west) {{\fontsize{11.040pt}{11.702pt}\selectfont  \makebox[433.60pt][s]{automation across digital marketplaces. As AI agents assume increasing responsibility for}}};
\node[anchor=base west,inner sep=0pt,outer sep=0pt] at ([xshift=82.104pt,yshift=-701.980pt]current page.north west) {{\fontsize{11.040pt}{11.702pt}\selectfont  \makebox[433.68pt][s]{financial decision-making, however, payment authorization systems must evolve to verify not}}};
\node[anchor=base west,inner sep=0pt,outer sep=0pt] at ([xshift=82.104pt,yshift=-714.700pt]current page.north west) {{\fontsize{11.040pt}{11.702pt}\selectfont  \makebox[433.62pt][s]{only transaction legitimacy but also the trustworthiness and behavioral consistency of}}};
\node[anchor=base west,inner sep=0pt,outer sep=0pt] at ([xshift=82.104pt,yshift=-727.300pt]current page.north west) {{\fontsize{11.040pt}{11.702pt}\selectfont  \makebox[178.71pt][s]{autonomous purchasing entities [3], [4].}}};
\node[anchor=base west,inner sep=0pt,outer sep=0pt] at ([xshift=503.38pt,yshift=-770.01pt]current page.north west) {\fontsize{9.96pt}{10pt}\selectfont 83};
\end{tikzpicture}
\null
\newpage
\begin{tikzpicture}[remember picture,overlay]
\node[anchor=base west,inner sep=0pt,outer sep=0pt] at ([xshift=127.580pt,yshift=-81.360pt]current page.north west) {{\fontsize{9.960pt}{10.558pt}\selectfont  \makebox[342.77pt][s]{International Journal on Cybernetics \& Informatics (IJCI) Vol.15, No.4, August 2026}}};
\node[anchor=base west,inner sep=0pt,outer sep=0pt] at ([xshift=503.380pt,yshift=-767.856pt]current page.north west) {{\fontsize{9.960pt}{10.558pt}\selectfont  84}};
\node[anchor=base west,inner sep=0pt,outer sep=0pt] at ([xshift=82.104pt,yshift=-100.220pt]current page.north west) {{\fontsize{11.040pt}{11.702pt}\selectfont  \makebox[433.62pt][s]{The growing adoption of autonomous purchasing agents introduces a new class of cybersecurity}}};
\node[anchor=base west,inner sep=0pt,outer sep=0pt] at ([xshift=82.104pt,yshift=-112.820pt]current page.north west) {{\fontsize{11.040pt}{11.702pt}\selectfont  \makebox[433.99pt][s]{challenges that extends beyond traditional payment fraud. Unlike conventional e-commerce}}};
\node[anchor=base west,inner sep=0pt,outer sep=0pt] at ([xshift=82.104pt,yshift=-125.540pt]current page.north west) {{\fontsize{11.040pt}{11.702pt}\selectfont  \makebox[433.64pt][s]{environments, agentic commerce relies on delegated financial authority, creating opportunities}}};
\node[anchor=base west,inner sep=0pt,outer sep=0pt] at ([xshift=82.104pt,yshift=-138.140pt]current page.north west) {{\fontsize{11.040pt}{11.702pt}\selectfont  \makebox[433.68pt][s]{for AI-agent impersonation, authorization abuse, malicious prompt manipulation, credential}}};
\node[anchor=base west,inner sep=0pt,outer sep=0pt] at ([xshift=82.104pt,yshift=-150.860pt]current page.north west) {{\fontsize{11.040pt}{11.702pt}\selectfont  \makebox[433.88pt][s]{compromise, and unauthorized transaction execution [2], [11], [12]. These emerging threats}}};
\node[anchor=base west,inner sep=0pt,outer sep=0pt] at ([xshift=82.104pt,yshift=-163.460pt]current page.north west) {{\fontsize{11.040pt}{11.702pt}\selectfont  \makebox[433.56pt][s]{require security mechanisms capable of evaluating transaction context and behavioral patterns}}};
\node[anchor=base west,inner sep=0pt,outer sep=0pt] at ([xshift=82.104pt,yshift=-176.060pt]current page.north west) {{\fontsize{11.040pt}{11.702pt}\selectfont  \makebox[431.43pt][s]{rather than relying solely on static rules or predefined fraud signatures. Consequently, behavior-}}};
\node[anchor=base west,inner sep=0pt,outer sep=0pt] at ([xshift=82.104pt,yshift=-188.780pt]current page.north west) {{\fontsize{11.040pt}{11.702pt}\selectfont  \makebox[433.94pt][s]{aware cybersecurity is becoming an essential component of next-generation payment}}};
\node[anchor=base west,inner sep=0pt,outer sep=0pt] at ([xshift=82.104pt,yshift=-201.380pt]current page.north west) {{\fontsize{11.040pt}{11.702pt}\selectfont  \makebox[101.16pt][s]{authorization systems.}}};
\node[anchor=base west,inner sep=0pt,outer sep=0pt] at ([xshift=82.104pt,yshift=-226.700pt]current page.north west) {{\fontsize{11.040pt}{11.702pt}\selectfont  \makebox[433.63pt][s]{Financial institutions have made substantial progress in applying machine learning to fraud}}};
\node[anchor=base west,inner sep=0pt,outer sep=0pt] at ([xshift=82.104pt,yshift=-239.300pt]current page.north west) {{\fontsize{11.040pt}{11.702pt}\selectfont  \makebox[433.58pt][s]{detection, with ensemble learning techniques such as Random Forest, XGBoost, and LightGBM}}};
\node[anchor=base west,inner sep=0pt,outer sep=0pt] at ([xshift=82.104pt,yshift=-252.020pt]current page.north west) {{\fontsize{11.040pt}{11.702pt}\selectfont  \makebox[431.24pt][s]{demonstrating strong performance in classification and financial fraud-detection applications [5]--}}};
\node[anchor=base west,inner sep=0pt,outer sep=0pt] at ([xshift=82.104pt,yshift=-264.620pt]current page.north west) {{\fontsize{11.040pt}{11.702pt}\selectfont  \makebox[433.65pt][s]{[7], [9], [10]. Recent studies have also shown that behavioral analytics and explainable AI}}};
\node[anchor=base west,inner sep=0pt,outer sep=0pt] at ([xshift=82.104pt,yshift=-277.340pt]current page.north west) {{\fontsize{11.040pt}{11.702pt}\selectfont  \makebox[433.67pt][s]{improve fraud detection by incorporating customer spending patterns and increasing model}}};
\node[anchor=base west,inner sep=0pt,outer sep=0pt] at ([xshift=82.104pt,yshift=-289.970pt]current page.north west) {{\fontsize{11.040pt}{11.702pt}\selectfont  \makebox[433.70pt][s]{transparency [8]--[10]. Nevertheless, most existing solutions are designed for conventional digital}}};
\node[anchor=base west,inner sep=0pt,outer sep=0pt] at ([xshift=82.104pt,yshift=-302.690pt]current page.north west) {{\fontsize{11.040pt}{11.702pt}\selectfont  \makebox[433.61pt][s]{payment environments, where transactions are initiated directly by human users. As autonomous}}};
\node[anchor=base west,inner sep=0pt,outer sep=0pt] at ([xshift=82.104pt,yshift=-315.290pt]current page.north west) {{\fontsize{11.040pt}{11.702pt}\selectfont  \makebox[433.62pt][s]{AI agents become active participants in commercial transactions, the assumptions underlying}}};
\node[anchor=base west,inner sep=0pt,outer sep=0pt] at ([xshift=82.104pt,yshift=-327.890pt]current page.north west) {{\fontsize{11.040pt}{11.702pt}\selectfont  \makebox[433.62pt][s]{many existing fraud detection systems may no longer be sufficient to address the evolving}}};
\node[anchor=base west,inner sep=0pt,outer sep=0pt] at ([xshift=82.104pt,yshift=-340.610pt]current page.north west) {{\fontsize{11.040pt}{11.702pt}\selectfont  \makebox[219.99pt][s]{cybersecurity requirements of agentic commerce.}}};
\node[anchor=base west,inner sep=0pt,outer sep=0pt] at ([xshift=82.104pt,yshift=-365.930pt]current page.north west) {{\fontsize{11.040pt}{11.702pt}\selectfont  \makebox[433.61pt][s]{Although prior research has investigated financial fraud detection, behavioral analytics,}}};
\node[anchor=base west,inner sep=0pt,outer sep=0pt] at ([xshift=82.104pt,yshift=-378.530pt]current page.north west) {{\fontsize{11.040pt}{11.702pt}\selectfont  \makebox[433.67pt][s]{explainable AI, and the security of autonomous commerce independently, comparatively little}}};
\node[anchor=base west,inner sep=0pt,outer sep=0pt] at ([xshift=82.104pt,yshift=-391.130pt]current page.north west) {{\fontsize{11.040pt}{11.702pt}\selectfont  \makebox[433.58pt][s]{attention has been given to integrating these capabilities into a unified cybersecurity framework}}};
\node[anchor=base west,inner sep=0pt,outer sep=0pt] at ([xshift=82.104pt,yshift=-403.850pt]current page.north west) {{\fontsize{11.040pt}{11.702pt}\selectfont  \makebox[433.98pt][s]{for securing payment authorization systems in emerging agentic commerce environments [3],}}};
\node[anchor=base west,inner sep=0pt,outer sep=0pt] at ([xshift=82.104pt,yshift=-416.450pt]current page.north west) {{\fontsize{11.040pt}{11.702pt}\selectfont  \makebox[434.00pt][s]{[11]--[13]. This research gap highlights the need for frameworks that combine behavior-aware}}};
\node[anchor=base west,inner sep=0pt,outer sep=0pt] at ([xshift=82.104pt,yshift=-429.170pt]current page.north west) {{\fontsize{11.040pt}{11.702pt}\selectfont  \makebox[433.91pt][s]{transaction analysis, intelligent fraud detection, and explainable decision-making while remaining}}};
\node[anchor=base west,inner sep=0pt,outer sep=0pt] at ([xshift=82.104pt,yshift=-441.770pt]current page.north west) {{\fontsize{11.040pt}{11.702pt}\selectfont  \makebox[389.24pt][s]{adaptable to the evolving threat landscape associated with autonomous financial agents.}}};
\node[anchor=base west,inner sep=0pt,outer sep=0pt] at ([xshift=82.104pt,yshift=-467.090pt]current page.north west) {{\fontsize{11.040pt}{11.702pt}\selectfont  \makebox[207.08pt][s]{To address this gap, this study proposes the}}};
\node[anchor=base west,inner sep=0pt,outer sep=0pt] at ([xshift=291.290pt,yshift=-467.090pt]current page.north west) {{\fontsize{11.040pt}{11.702pt}\selectfont \bfseries \makebox[224.64pt][s]{Autonomous Commerce Agent Guard System}}};
\node[anchor=base west,inner sep=0pt,outer sep=0pt] at ([xshift=82.104pt,yshift=-479.710pt]current page.north west) {{\fontsize{11.040pt}{11.702pt}\selectfont \bfseries (ACAGS)}};
\node[anchor=base west,inner sep=0pt,outer sep=0pt] at ([xshift=128.060pt,yshift=-479.710pt]current page.north west) {{\fontsize{11.040pt}{11.702pt}\selectfont  \makebox[387.89pt][s]{, a cybersecurity framework for behavior-aware fraud detection and trusted payment}}};
\node[anchor=base west,inner sep=0pt,outer sep=0pt] at ([xshift=82.104pt,yshift=-492.430pt]current page.north west) {{\fontsize{11.040pt}{11.702pt}\selectfont  \makebox[433.65pt][s]{authorization in agentic commerce. ACAGS integrates behavioral feature engineering, supervised}}};
\node[anchor=base west,inner sep=0pt,outer sep=0pt] at ([xshift=82.104pt,yshift=-505.030pt]current page.north west) {{\fontsize{11.040pt}{11.702pt}\selectfont  \makebox[433.89pt][s]{machine learning, agent-specific authorization controls, and SHAP-based explainable AI to}}};
\node[anchor=base west,inner sep=0pt,outer sep=0pt] at ([xshift=82.104pt,yshift=-517.630pt]current page.north west) {{\fontsize{11.040pt}{11.702pt}\selectfont  \makebox[433.63pt][s]{assess transaction risk and support interpretable authorization decisions. The study evaluates}}};
\node[anchor=base west,inner sep=0pt,outer sep=0pt] at ([xshift=82.104pt,yshift=-530.350pt]current page.north west) {{\fontsize{11.040pt}{11.702pt}\selectfont  \makebox[433.96pt][s]{Logistic Regression, Random Forest, XGBoost, and LightGBM on a large-scale financial}}};
\node[anchor=base west,inner sep=0pt,outer sep=0pt] at ([xshift=82.104pt,yshift=-542.950pt]current page.north west) {{\fontsize{11.040pt}{11.702pt}\selectfont  \makebox[433.90pt][s]{transaction dataset; extends the evaluation through simulated agent-specific features and}}};
\node[anchor=base west,inner sep=0pt,outer sep=0pt] at ([xshift=82.104pt,yshift=-555.670pt]current page.north west) {{\fontsize{11.040pt}{11.702pt}\selectfont  \makebox[431.43pt][s]{adversarial scenarios; and performs ablation analysis, statistical testing, and authorization-}}};
\node[anchor=base west,inner sep=0pt,outer sep=0pt] at ([xshift=82.104pt,yshift=-568.270pt]current page.north west) {{\fontsize{11.040pt}{11.702pt}\selectfont  \makebox[433.90pt][s]{outcome assessment. Together, these contributions provide a behavior-aware and explainable}}};
\node[anchor=base west,inner sep=0pt,outer sep=0pt] at ([xshift=82.104pt,yshift=-580.870pt]current page.north west) {{\fontsize{11.040pt}{11.702pt}\selectfont  \makebox[433.95pt][s]{approach for evaluating fraud and authorization risks in simulated agentic-commerce payment}}};
\node[anchor=base west,inner sep=0pt,outer sep=0pt] at ([xshift=82.104pt,yshift=-593.590pt]current page.north west) {{\fontsize{11.040pt}{11.702pt}\selectfont  environments.}};
\node[anchor=base west,inner sep=0pt,outer sep=0pt] at ([xshift=82.104pt,yshift=-621.790pt]current page.north west) {{\fontsize{14.040pt}{14.882pt}\selectfont \bfseries 2.}};
\node[anchor=base west,inner sep=0pt,outer sep=0pt] at ([xshift=100.100pt,yshift=-621.790pt]current page.north west) {{\fontsize{14.040pt}{14.882pt}\selectfont \bfseries L}};
\node[anchor=base west,inner sep=0pt,outer sep=0pt] at ([xshift=109.460pt,yshift=-621.790pt]current page.north west) {{\fontsize{11.040pt}{11.702pt}\selectfont \bfseries ITERATURE}};
\node[anchor=base west,inner sep=0pt,outer sep=0pt] at ([xshift=176.660pt,yshift=-621.790pt]current page.north west) {{\fontsize{14.040pt}{14.882pt}\selectfont \bfseries R}};
\node[anchor=base west,inner sep=0pt,outer sep=0pt] at ([xshift=186.860pt,yshift=-621.790pt]current page.north west) {{\fontsize{11.040pt}{11.702pt}\selectfont \bfseries EVIEW}};
\node[anchor=base west,inner sep=0pt,outer sep=0pt] at ([xshift=82.104pt,yshift=-647.590pt]current page.north west) {{\fontsize{11.040pt}{11.702pt}\selectfont  \makebox[433.99pt][s]{Existing research on agentic commerce spans payment authorization, cybersecurity, AI-based}}};
\node[anchor=base west,inner sep=0pt,outer sep=0pt] at ([xshift=82.104pt,yshift=-660.340pt]current page.north west) {{\fontsize{11.040pt}{11.702pt}\selectfont  \makebox[433.67pt][s]{fraud detection, behavioral analytics, and explainable AI. While prior studies have advanced}}};
\node[anchor=base west,inner sep=0pt,outer sep=0pt] at ([xshift=82.104pt,yshift=-672.940pt]current page.north west) {{\fontsize{11.040pt}{11.702pt}\selectfont  \makebox[433.61pt][s]{these areas individually, their integration within autonomous payment environments remains}}};
\node[anchor=base west,inner sep=0pt,outer sep=0pt] at ([xshift=82.104pt,yshift=-685.660pt]current page.north west) {{\fontsize{11.040pt}{11.702pt}\selectfont  \makebox[433.63pt][s]{limited. This review examines these complementary areas and identifies the gap motivating the}}};
\node[anchor=base west,inner sep=0pt,outer sep=0pt] at ([xshift=82.104pt,yshift=-698.260pt]current page.north west) {{\fontsize{11.040pt}{11.702pt}\selectfont  \makebox[136.23pt][s]{proposed ACAGS framework.}}};
\end{tikzpicture}
\null
\newpage
\begin{tikzpicture}[remember picture,overlay]
\node[anchor=base west,inner sep=0pt,outer sep=0pt] at ([xshift=127.580pt,yshift=-81.360pt]current page.north west) {{\fontsize{9.960pt}{10.558pt}\selectfont  \makebox[342.77pt][s]{International Journal on Cybernetics \& Informatics (IJCI) Vol.15, No.4, August 2026}}};
\node[anchor=base west,inner sep=0pt,outer sep=0pt] at ([xshift=503.380pt,yshift=-767.856pt]current page.north west) {{\fontsize{9.960pt}{10.558pt}\selectfont  85}};
\node[anchor=base west,inner sep=0pt,outer sep=0pt] at ([xshift=82.104pt,yshift=-101.180pt]current page.north west) {{\fontsize{12.000pt}{12.720pt}\selectfont \bfseries 2.1.}};
\node[anchor=base west,inner sep=0pt,outer sep=0pt] at ([xshift=104.660pt,yshift=-101.180pt]current page.north west) {{\fontsize{12.000pt}{12.720pt}\selectfont \bfseries \makebox[246.17pt][s]{Agentic Commerce and Payment Authorization}}};
\node[anchor=base west,inner sep=0pt,outer sep=0pt] at ([xshift=82.104pt,yshift=-126.620pt]current page.north west) {{\fontsize{11.040pt}{11.702pt}\selectfont  \makebox[433.61pt][s]{The rapid advancement of artificial intelligence is reshaping digital commerce from conventional}}};
\node[anchor=base west,inner sep=0pt,outer sep=0pt] at ([xshift=82.104pt,yshift=-139.340pt]current page.north west) {{\fontsize{11.040pt}{11.702pt}\selectfont  \makebox[126.90pt][s]{user-driven transactions to}}};
\node[anchor=base west,inner sep=0pt,outer sep=0pt] at ([xshift=212.810pt,yshift=-139.340pt]current page.north west) {{\fontsize{11.040pt}{11.702pt}\selectfont \bfseries \makebox[88.38pt][s]{agentic commerce}}};
\node[anchor=base west,inner sep=0pt,outer sep=0pt] at ([xshift=301.250pt,yshift=-139.340pt]current page.north west) {{\fontsize{11.040pt}{11.702pt}\selectfont  \makebox[214.68pt][s]{, in which autonomous software agents can}}};
\node[anchor=base west,inner sep=0pt,outer sep=0pt] at ([xshift=82.104pt,yshift=-151.940pt]current page.north west) {{\fontsize{11.040pt}{11.702pt}\selectfont  \makebox[433.65pt][s]{independently discover products, compare alternatives, negotiate prices, and execute purchasing}}};
\node[anchor=base west,inner sep=0pt,outer sep=0pt] at ([xshift=82.104pt,yshift=-164.660pt]current page.north west) {{\fontsize{11.040pt}{11.702pt}\selectfont  \makebox[433.87pt][s]{decisions on behalf of users. Unlike traditional e-commerce, where payment authorization is}}};
\node[anchor=base west,inner sep=0pt,outer sep=0pt] at ([xshift=82.104pt,yshift=-177.260pt]current page.north west) {{\fontsize{11.040pt}{11.702pt}\selectfont  \makebox[433.71pt][s]{explicitly initiated and verified by human participants, agentic commerce delegates portions of}}};
\node[anchor=base west,inner sep=0pt,outer sep=0pt] at ([xshift=82.104pt,yshift=-189.860pt]current page.north west) {{\fontsize{11.040pt}{11.702pt}\selectfont  \makebox[433.65pt][s]{the purchasing and payment lifecycle to intelligent agents operating within predefined user}}};
\node[anchor=base west,inner sep=0pt,outer sep=0pt] at ([xshift=82.104pt,yshift=-202.580pt]current page.north west) {{\fontsize{11.040pt}{11.702pt}\selectfont  \makebox[433.63pt][s]{preferences and authorization policies. This transformation has the potential to improve}}};
\node[anchor=base west,inner sep=0pt,outer sep=0pt] at ([xshift=82.104pt,yshift=-215.180pt]current page.north west) {{\fontsize{11.040pt}{11.702pt}\selectfont  \makebox[433.67pt][s]{transaction efficiency, personalization, and scalability while fundamentally changing how}}};
\node[anchor=base west,inner sep=0pt,outer sep=0pt] at ([xshift=82.104pt,yshift=-227.900pt]current page.north west) {{\fontsize{11.040pt}{11.702pt}\selectfont  \makebox[256.01pt][s]{financial transactions are initiated and authorized [1]--[4].}}};
\node[anchor=base west,inner sep=0pt,outer sep=0pt] at ([xshift=82.104pt,yshift=-253.100pt]current page.north west) {{\fontsize{11.040pt}{11.702pt}\selectfont  \makebox[433.60pt][s]{As autonomous agents increasingly participate in digital payments, authorization must extend}}};
\node[anchor=base west,inner sep=0pt,outer sep=0pt] at ([xshift=82.104pt,yshift=-265.820pt]current page.north west) {{\fontsize{11.040pt}{11.702pt}\selectfont  \makebox[433.71pt][s]{beyond identity and credential verification to continuous behavioral assessment. Their}}};
\node[anchor=base west,inner sep=0pt,outer sep=0pt] at ([xshift=82.104pt,yshift=-278.420pt]current page.north west) {{\fontsize{11.040pt}{11.702pt}\selectfont  \makebox[433.65pt][s]{interactions across merchants and financial institutions introduce risks involving delegated}}};
\node[anchor=base west,inner sep=0pt,outer sep=0pt] at ([xshift=82.104pt,yshift=-291.170pt]current page.north west) {{\fontsize{11.040pt}{11.702pt}\selectfont  \makebox[433.63pt][s]{authority, compromised agents, and fraudulent transactions. Existing research largely emphasizes}}};
\node[anchor=base west,inner sep=0pt,outer sep=0pt] at ([xshift=82.104pt,yshift=-303.770pt]current page.north west) {{\fontsize{11.040pt}{11.702pt}\selectfont  \makebox[433.89pt][s]{authorization and trust mechanisms, leaving a need for behavior-aware fraud detection that}}};
\node[anchor=base west,inner sep=0pt,outer sep=0pt] at ([xshift=82.104pt,yshift=-316.490pt]current page.north west) {{\fontsize{11.040pt}{11.702pt}\selectfont  \makebox[433.63pt][s]{continuously evaluates transaction legitimacy in agentic commerce environments [3], [4], [11],}}};
\node[anchor=base west,inner sep=0pt,outer sep=0pt] at ([xshift=82.104pt,yshift=-329.090pt]current page.north west) {{\fontsize{11.040pt}{11.702pt}\selectfont  [12].}};
\node[anchor=base west,inner sep=0pt,outer sep=0pt] at ([xshift=82.104pt,yshift=-355.370pt]current page.north west) {{\fontsize{12.000pt}{12.720pt}\selectfont \bfseries 2.2.}};
\node[anchor=base west,inner sep=0pt,outer sep=0pt] at ([xshift=104.660pt,yshift=-355.370pt]current page.north west) {{\fontsize{12.000pt}{12.720pt}\selectfont \bfseries \makebox[244.99pt][s]{Cybersecurity Challenges in Agentic Commerce}}};
\node[anchor=base west,inner sep=0pt,outer sep=0pt] at ([xshift=82.104pt,yshift=-380.810pt]current page.north west) {{\fontsize{11.040pt}{11.702pt}\selectfont  \makebox[433.81pt][s]{The increasing autonomy of AI-driven purchasing agents fundamentally expands the}}};
\node[anchor=base west,inner sep=0pt,outer sep=0pt] at ([xshift=82.104pt,yshift=-393.530pt]current page.north west) {{\fontsize{11.040pt}{11.702pt}\selectfont  \makebox[433.71pt][s]{cybersecurity attack surface of digital payment ecosystems. Unlike conventional payment}}};
\node[anchor=base west,inner sep=0pt,outer sep=0pt] at ([xshift=82.104pt,yshift=-406.130pt]current page.north west) {{\fontsize{11.040pt}{11.702pt}\selectfont  \makebox[433.60pt][s]{systems, where transaction authorization is directly initiated and monitored by human users,}}};
\node[anchor=base west,inner sep=0pt,outer sep=0pt] at ([xshift=82.104pt,yshift=-418.730pt]current page.north west) {{\fontsize{11.040pt}{11.702pt}\selectfont  \makebox[433.82pt][s]{agentic commerce introduces delegated decision-making, allowing autonomous agents to execute}}};
\node[anchor=base west,inner sep=0pt,outer sep=0pt] at ([xshift=82.104pt,yshift=-431.450pt]current page.north west) {{\fontsize{11.040pt}{11.702pt}\selectfont  \makebox[433.70pt][s]{purchases based on predefined objectives and contextual information. While this autonomy}}};
\node[anchor=base west,inner sep=0pt,outer sep=0pt] at ([xshift=82.104pt,yshift=-444.050pt]current page.north west) {{\fontsize{11.040pt}{11.702pt}\selectfont  \makebox[433.58pt][s]{improves efficiency and scalability, it also creates new opportunities for adversaries to exploit}}};
\node[anchor=base west,inner sep=0pt,outer sep=0pt] at ([xshift=82.104pt,yshift=-456.770pt]current page.north west) {{\fontsize{11.040pt}{11.702pt}\selectfont  \makebox[433.66pt][s]{authorization workflows through compromised agents, manipulated decision policies, prompt}}};
\node[anchor=base west,inner sep=0pt,outer sep=0pt] at ([xshift=82.104pt,yshift=-469.370pt]current page.north west) {{\fontsize{11.040pt}{11.702pt}\selectfont  \makebox[383.14pt][s]{injection, identity spoofing, credential theft, and unauthorized transaction execution}}};
\node[anchor=base west,inner sep=0pt,outer sep=0pt] at ([xshift=467.020pt,yshift=-469.370pt]current page.north west) {{\fontsize{11.040pt}{11.702pt}\selectfont \bfseries [}};
\node[anchor=base west,inner sep=0pt,outer sep=0pt] at ([xshift=470.740pt,yshift=-469.370pt]current page.north west) {{\fontsize{11.040pt}{11.702pt}\selectfont  11], [12],}};
\node[anchor=base west,inner sep=0pt,outer sep=0pt] at ([xshift=82.104pt,yshift=-481.990pt]current page.north west) {{\fontsize{11.040pt}{11.702pt}\selectfont  [15].}};
\node[anchor=base west,inner sep=0pt,outer sep=0pt] at ([xshift=82.104pt,yshift=-507.310pt]current page.north west) {{\fontsize{11.040pt}{11.702pt}\selectfont  \makebox[433.57pt][s]{Existing security research has primarily focused on strengthening authentication, authorization}}};
\node[anchor=base west,inner sep=0pt,outer sep=0pt] at ([xshift=82.104pt,yshift=-520.030pt]current page.north west) {{\fontsize{11.040pt}{11.702pt}\selectfont  \makebox[433.63pt][s]{protocols, trusted agent identities, and secure communication mechanisms for autonomous}}};
\node[anchor=base west,inner sep=0pt,outer sep=0pt] at ([xshift=82.104pt,yshift=-532.630pt]current page.north west) {{\fontsize{11.040pt}{11.702pt}\selectfont  \makebox[433.60pt][s]{systems. Although these approaches improve the integrity of agent interactions, they generally}}};
\node[anchor=base west,inner sep=0pt,outer sep=0pt] at ([xshift=82.104pt,yshift=-545.230pt]current page.north west) {{\fontsize{11.040pt}{11.702pt}\selectfont  \makebox[433.58pt][s]{emphasize preventive security controls rather than continuously assessing transaction behavior}}};
\node[anchor=base west,inner sep=0pt,outer sep=0pt] at ([xshift=82.104pt,yshift=-557.950pt]current page.north west) {{\fontsize{11.040pt}{11.702pt}\selectfont  \makebox[433.67pt][s]{after authorization decisions have been initiated. Consequently, malicious or compromised agents}}};
\node[anchor=base west,inner sep=0pt,outer sep=0pt] at ([xshift=82.104pt,yshift=-570.550pt]current page.north west) {{\fontsize{11.040pt}{11.702pt}\selectfont  \makebox[433.61pt][s]{may still generate transactions that satisfy authorization policies while exhibiting behavioral}}};
\node[anchor=base west,inner sep=0pt,outer sep=0pt] at ([xshift=82.104pt,yshift=-583.270pt]current page.north west) {{\fontsize{11.040pt}{11.702pt}\selectfont  \makebox[433.64pt][s]{characteristics inconsistent with legitimate purchasing activities. Addressing this limitation}}};
\node[anchor=base west,inner sep=0pt,outer sep=0pt] at ([xshift=82.104pt,yshift=-595.870pt]current page.north west) {{\fontsize{11.040pt}{11.702pt}\selectfont  \makebox[433.66pt][s]{requires cybersecurity frameworks capable of combining intelligent fraud detection with}}};
\node[anchor=base west,inner sep=0pt,outer sep=0pt] at ([xshift=82.104pt,yshift=-608.470pt]current page.north west) {{\fontsize{11.040pt}{11.702pt}\selectfont  \makebox[433.90pt][s]{continuous behavioral analysis and explainable decision-making throughout the payment}}};
\node[anchor=base west,inner sep=0pt,outer sep=0pt] at ([xshift=82.104pt,yshift=-621.190pt]current page.north west) {{\fontsize{11.040pt}{11.702pt}\selectfont  \makebox[165.39pt][s]{authorization process[11], [12], [14].}}};
\node[anchor=base west,inner sep=0pt,outer sep=0pt] at ([xshift=82.104pt,yshift=-647.470pt]current page.north west) {{\fontsize{12.000pt}{12.720pt}\selectfont \bfseries 2.3.}};
\node[anchor=base west,inner sep=0pt,outer sep=0pt] at ([xshift=104.660pt,yshift=-647.470pt]current page.north west) {{\fontsize{12.000pt}{12.720pt}\selectfont \bfseries \makebox[245.39pt][s]{AI-Based Fraud Detection for Payment Systems}}};
\node[anchor=base west,inner sep=0pt,outer sep=0pt] at ([xshift=82.104pt,yshift=-672.940pt]current page.north west) {{\fontsize{11.040pt}{11.702pt}\selectfont  \makebox[433.97pt][s]{The increasing sophistication of financial fraud has driven widespread adoption of AI-based fraud}}};
\node[anchor=base west,inner sep=0pt,outer sep=0pt] at ([xshift=82.104pt,yshift=-685.660pt]current page.north west) {{\fontsize{11.040pt}{11.702pt}\selectfont  \makebox[433.62pt][s]{detection techniques capable of identifying complex and evolving transaction patterns. Recent}}};
\node[anchor=base west,inner sep=0pt,outer sep=0pt] at ([xshift=82.104pt,yshift=-698.260pt]current page.north west) {{\fontsize{11.040pt}{11.702pt}\selectfont  \makebox[433.65pt][s]{studies consistently demonstrate that ensemble learning methods outperform traditional statistical}}};
\node[anchor=base west,inner sep=0pt,outer sep=0pt] at ([xshift=82.104pt,yshift=-710.860pt]current page.north west) {{\fontsize{11.040pt}{11.702pt}\selectfont  \makebox[433.62pt][s]{approaches in highly imbalanced payment datasets, achieving substantial improvements in fraud}}};
\node[anchor=base west,inner sep=0pt,outer sep=0pt] at ([xshift=82.104pt,yshift=-723.580pt]current page.north west) {{\fontsize{11.040pt}{11.702pt}\selectfont  \makebox[433.66pt][s]{detection accuracy, precision, and recall. In particular, XGBoost, Random Forest, and LightGBM}}};
\node[anchor=base west,inner sep=0pt,outer sep=0pt] at ([xshift=82.104pt,yshift=-736.180pt]current page.north west) {{\fontsize{11.040pt}{11.702pt}\selectfont  \makebox[433.75pt][s]{have been widely adopted because of their ability to capture nonlinear relationships among}}};
\end{tikzpicture}
\null
\newpage
\begin{tikzpicture}[remember picture,overlay]
\node[anchor=base west,inner sep=0pt,outer sep=0pt] at ([xshift=127.580pt,yshift=-81.360pt]current page.north west) {{\fontsize{9.960pt}{10.558pt}\selectfont  \makebox[342.77pt][s]{International Journal on Cybernetics \& Informatics (IJCI) Vol.15, No.4, August 2026}}};
\node[anchor=base west,inner sep=0pt,outer sep=0pt] at ([xshift=503.380pt,yshift=-767.856pt]current page.north west) {{\fontsize{9.960pt}{10.558pt}\selectfont  86}};
\node[anchor=base west,inner sep=0pt,outer sep=0pt] at ([xshift=82.104pt,yshift=-100.220pt]current page.north west) {{\fontsize{11.040pt}{11.702pt}\selectfont  \makebox[433.61pt][s]{transactional attributes while maintaining strong predictive performance across diverse financial}}};
\node[anchor=base west,inner sep=0pt,outer sep=0pt] at ([xshift=82.104pt,yshift=-112.820pt]current page.north west) {{\fontsize{11.040pt}{11.702pt}\selectfont  \makebox[104.76pt][s]{environments [9], [10].}}};
\node[anchor=base west,inner sep=0pt,outer sep=0pt] at ([xshift=82.104pt,yshift=-138.140pt]current page.north west) {{\fontsize{11.040pt}{11.702pt}\selectfont  \makebox[433.72pt][s]{Despite these advances, most existing fraud detection models are developed for conventional}}};
\node[anchor=base west,inner sep=0pt,outer sep=0pt] at ([xshift=82.104pt,yshift=-150.860pt]current page.north west) {{\fontsize{11.040pt}{11.702pt}\selectfont  \makebox[433.64pt][s]{payment ecosystems in which transactions are initiated directly by human users. Their primary}}};
\node[anchor=base west,inner sep=0pt,outer sep=0pt] at ([xshift=82.104pt,yshift=-163.460pt]current page.north west) {{\fontsize{11.040pt}{11.702pt}\selectfont  \makebox[433.61pt][s]{emphasis remains on improving predictive performance through feature optimization, class}}};
\node[anchor=base west,inner sep=0pt,outer sep=0pt] at ([xshift=82.104pt,yshift=-176.060pt]current page.north west) {{\fontsize{11.040pt}{11.702pt}\selectfont  \makebox[433.64pt][s]{imbalance handling, or model tuning, with comparatively limited consideration of the}}};
\node[anchor=base west,inner sep=0pt,outer sep=0pt] at ([xshift=82.104pt,yshift=-188.780pt]current page.north west) {{\fontsize{11.040pt}{11.702pt}\selectfont  \makebox[433.74pt][s]{cybersecurity implications introduced by autonomous purchasing agents. As payment}}};
\node[anchor=base west,inner sep=0pt,outer sep=0pt] at ([xshift=82.104pt,yshift=-201.380pt]current page.north west) {{\fontsize{11.040pt}{11.702pt}\selectfont  \makebox[433.86pt][s]{authorization increasingly involves AI-driven decision-making, fraud detection must evolve}}};
\node[anchor=base west,inner sep=0pt,outer sep=0pt] at ([xshift=82.104pt,yshift=-214.100pt]current page.north west) {{\fontsize{11.040pt}{11.702pt}\selectfont  \makebox[433.67pt][s]{beyond prediction accuracy to incorporate behavioral context, continuous transaction assessment,}}};
\node[anchor=base west,inner sep=0pt,outer sep=0pt] at ([xshift=82.104pt,yshift=-226.700pt]current page.north west) {{\fontsize{11.040pt}{11.702pt}\selectfont  \makebox[433.64pt][s]{and explainable security mechanisms capable of supporting trustworthy authorization decisions}}};
\node[anchor=base west,inner sep=0pt,outer sep=0pt] at ([xshift=82.104pt,yshift=-239.300pt]current page.north west) {{\fontsize{11.040pt}{11.702pt}\selectfont  \makebox[136.61pt][s]{in agentic commerce [12], [13]}}};
\node[anchor=base west,inner sep=0pt,outer sep=0pt] at ([xshift=218.810pt,yshift=-239.300pt]current page.north west) {{\fontsize{11.040pt}{11.702pt}\selectfont \bfseries .}};
\node[anchor=base west,inner sep=0pt,outer sep=0pt] at ([xshift=82.104pt,yshift=-265.580pt]current page.north west) {{\fontsize{12.000pt}{12.720pt}\selectfont \bfseries 2.4.}};
\node[anchor=base west,inner sep=0pt,outer sep=0pt] at ([xshift=104.660pt,yshift=-265.580pt]current page.north west) {{\fontsize{12.000pt}{12.720pt}\selectfont \bfseries \makebox[208.70pt][s]{Behavioral Analytics and Explainable AI}}};
\node[anchor=base west,inner sep=0pt,outer sep=0pt] at ([xshift=82.104pt,yshift=-291.170pt]current page.north west) {{\fontsize{11.040pt}{11.702pt}\selectfont  \makebox[433.58pt][s]{Behavioral analytics has become an increasingly important component of modern financial fraud}}};
\node[anchor=base west,inner sep=0pt,outer sep=0pt] at ([xshift=82.104pt,yshift=-303.770pt]current page.north west) {{\fontsize{11.040pt}{11.702pt}\selectfont  \makebox[433.61pt][s]{detection by enabling security systems to evaluate transaction context rather than individual}}};
\node[anchor=base west,inner sep=0pt,outer sep=0pt] at ([xshift=82.104pt,yshift=-316.490pt]current page.north west) {{\fontsize{11.040pt}{11.702pt}\selectfont  \makebox[433.58pt][s]{transactions in isolation. Recent studies show that features derived from historical spending}}};
\node[anchor=base west,inner sep=0pt,outer sep=0pt] at ([xshift=82.104pt,yshift=-329.090pt]current page.north west) {{\fontsize{11.040pt}{11.702pt}\selectfont  \makebox[433.67pt][s]{behavior, transaction timing, merchant interactions, and customer purchasing patterns}}};
\node[anchor=base west,inner sep=0pt,outer sep=0pt] at ([xshift=82.104pt,yshift=-341.690pt]current page.north west) {{\fontsize{11.040pt}{11.702pt}\selectfont  \makebox[433.63pt][s]{significantly improve the identification of anomalous financial activities compared with models}}};
\node[anchor=base west,inner sep=0pt,outer sep=0pt] at ([xshift=82.104pt,yshift=-354.410pt]current page.north west) {{\fontsize{11.040pt}{11.702pt}\selectfont  \makebox[433.61pt][s]{relying solely on static transactional attributes. These behavioral representations provide}}};
\node[anchor=base west,inner sep=0pt,outer sep=0pt] at ([xshift=82.104pt,yshift=-367.010pt]current page.north west) {{\fontsize{11.040pt}{11.702pt}\selectfont  \makebox[433.63pt][s]{additional contextual intelligence that enhances the ability to distinguish legitimate customer}}};
\node[anchor=base west,inner sep=0pt,outer sep=0pt] at ([xshift=82.104pt,yshift=-379.730pt]current page.north west) {{\fontsize{11.040pt}{11.702pt}\selectfont  \makebox[244.13pt][s]{behavior from fraudulent transaction patterns [9], [10].}}};
\node[anchor=base west,inner sep=0pt,outer sep=0pt] at ([xshift=82.104pt,yshift=-404.930pt]current page.north west) {{\fontsize{11.040pt}{11.702pt}\selectfont  \makebox[433.99pt][s]{In parallel, explainable AI has emerged as an important requirement for deploying AI-based}}};
\node[anchor=base west,inner sep=0pt,outer sep=0pt] at ([xshift=82.104pt,yshift=-417.650pt]current page.north west) {{\fontsize{11.040pt}{11.702pt}\selectfont  \makebox[433.60pt][s]{fraud detection systems within regulated financial environments. Techniques such as SHAP}}};
\node[anchor=base west,inner sep=0pt,outer sep=0pt] at ([xshift=82.104pt,yshift=-430.250pt]current page.north west) {{\fontsize{11.040pt}{11.702pt}\selectfont  \makebox[433.62pt][s]{improve model transparency by quantifying the contribution of individual features to prediction}}};
\node[anchor=base west,inner sep=0pt,outer sep=0pt] at ([xshift=82.104pt,yshift=-442.970pt]current page.north west) {{\fontsize{11.040pt}{11.702pt}\selectfont  \makebox[433.60pt][s]{outcomes, enabling analysts to interpret fraud decisions [8]. Explainability also supports broader}}};
\node[anchor=base west,inner sep=0pt,outer sep=0pt] at ([xshift=82.104pt,yshift=-455.570pt]current page.north west) {{\fontsize{11.040pt}{11.702pt}\selectfont  \makebox[433.98pt][s]{requirements for trustworthy and accountable AI deployment in high-impact applications [14].}}};
\node[anchor=base west,inner sep=0pt,outer sep=0pt] at ([xshift=82.104pt,yshift=-468.170pt]current page.north west) {{\fontsize{11.040pt}{11.702pt}\selectfont  \makebox[433.62pt][s]{While behavioral analytics and explainability have each demonstrated value independently, their}}};
\node[anchor=base west,inner sep=0pt,outer sep=0pt] at ([xshift=82.104pt,yshift=-480.910pt]current page.north west) {{\fontsize{11.040pt}{11.702pt}\selectfont  \makebox[433.67pt][s]{integration within cybersecurity frameworks specifically designed for agentic commerce payment}}};
\node[anchor=base west,inner sep=0pt,outer sep=0pt] at ([xshift=82.104pt,yshift=-493.510pt]current page.north west) {{\fontsize{11.040pt}{11.702pt}\selectfont  \makebox[275.93pt][s]{authorization remains an emerging area of research [11]--[13].}}};
\node[anchor=base west,inner sep=0pt,outer sep=0pt] at ([xshift=82.104pt,yshift=-519.790pt]current page.north west) {{\fontsize{12.000pt}{12.720pt}\selectfont \bfseries 2.5.}};
\node[anchor=base west,inner sep=0pt,outer sep=0pt] at ([xshift=104.660pt,yshift=-519.790pt]current page.north west) {{\fontsize{12.000pt}{12.720pt}\selectfont \bfseries Research Gap}};
\node[anchor=base west,inner sep=0pt,outer sep=0pt] at ([xshift=82.104pt,yshift=-545.230pt]current page.north west) {{\fontsize{11.040pt}{11.702pt}\selectfont  \makebox[433.58pt][s]{The literature reviewed in the preceding subsections demonstrates significant progress in four}}};
\node[anchor=base west,inner sep=0pt,outer sep=0pt] at ([xshift=82.104pt,yshift=-557.950pt]current page.north west) {{\fontsize{11.040pt}{11.702pt}\selectfont  \makebox[433.63pt][s]{complementary research areas: agentic commerce, cybersecurity for autonomous payment}}};
\node[anchor=base west,inner sep=0pt,outer sep=0pt] at ([xshift=82.104pt,yshift=-570.550pt]current page.north west) {{\fontsize{11.040pt}{11.702pt}\selectfont  \makebox[433.63pt][s]{systems, AI-based financial fraud detection, and explainable behavioral analytics. Recent studies}}};
\node[anchor=base west,inner sep=0pt,outer sep=0pt] at ([xshift=82.104pt,yshift=-583.270pt]current page.north west) {{\fontsize{11.040pt}{11.702pt}\selectfont  \makebox[433.65pt][s]{have strengthened payment authorization through improved authentication mechanisms, trust}}};
\node[anchor=base west,inner sep=0pt,outer sep=0pt] at ([xshift=82.104pt,yshift=-595.870pt]current page.north west) {{\fontsize{11.040pt}{11.702pt}\selectfont  \makebox[433.78pt][s]{models, and machine learning-based fraud detection, while explainable AI has enhanced the}}};
\node[anchor=base west,inner sep=0pt,outer sep=0pt] at ([xshift=82.104pt,yshift=-608.470pt]current page.north west) {{\fontsize{11.040pt}{11.702pt}\selectfont  \makebox[433.80pt][s]{transparency of predictive decision-making.Collectively, these advances have contributed to}}};
\node[anchor=base west,inner sep=0pt,outer sep=0pt] at ([xshift=82.104pt,yshift=-621.190pt]current page.north west) {{\fontsize{11.040pt}{11.702pt}\selectfont  \makebox[433.93pt][s]{improving the security, reliability, and transparency of digital payment and AI-enabled financial}}};
\node[anchor=base west,inner sep=0pt,outer sep=0pt] at ([xshift=82.104pt,yshift=-633.790pt]current page.north west) {{\fontsize{11.040pt}{11.702pt}\selectfont  \makebox[116.43pt][s]{systems [3], [8], [9], [14].}}};
\node[anchor=base west,inner sep=0pt,outer sep=0pt] at ([xshift=82.104pt,yshift=-659.110pt]current page.north west) {{\fontsize{11.040pt}{11.702pt}\selectfont  \makebox[433.70pt][s]{Despite these developments, several limitations remain. Existing fraud detection studies primarily}}};
\node[anchor=base west,inner sep=0pt,outer sep=0pt] at ([xshift=82.104pt,yshift=-671.860pt]current page.north west) {{\fontsize{11.040pt}{11.702pt}\selectfont  \makebox[433.65pt][s]{target conventional payment environments, where transactions are initiated directly by human}}};
\node[anchor=base west,inner sep=0pt,outer sep=0pt] at ([xshift=82.104pt,yshift=-684.460pt]current page.north west) {{\fontsize{11.040pt}{11.702pt}\selectfont  \makebox[433.68pt][s]{users, while research on agentic commerce has largely focused on authorization protocols,}}};
\node[anchor=base west,inner sep=0pt,outer sep=0pt] at ([xshift=82.104pt,yshift=-697.060pt]current page.north west) {{\fontsize{11.040pt}{11.702pt}\selectfont  \makebox[433.61pt][s]{identity management, and trust establishment for autonomous agents. Comparatively limited}}};
\node[anchor=base west,inner sep=0pt,outer sep=0pt] at ([xshift=82.104pt,yshift=-709.780pt]current page.north west) {{\fontsize{11.040pt}{11.702pt}\selectfont  \makebox[433.83pt][s]{attention has been given to integrating behavior-aware fraud detection, behavioral feature}}};
\node[anchor=base west,inner sep=0pt,outer sep=0pt] at ([xshift=82.104pt,yshift=-722.380pt]current page.north west) {{\fontsize{11.040pt}{11.702pt}\selectfont  \makebox[433.80pt][s]{engineering, and explainable AI into a unified cybersecurity framework capable of continuously}}};
\node[anchor=base west,inner sep=0pt,outer sep=0pt] at ([xshift=82.104pt,yshift=-735.100pt]current page.north west) {{\fontsize{11.040pt}{11.702pt}\selectfont  assessing}};
\node[anchor=base west,inner sep=0pt,outer sep=0pt] at ([xshift=141.113pt,yshift=-735.100pt]current page.north west) {{\fontsize{11.040pt}{11.702pt}\selectfont  payment}};
\node[anchor=base west,inner sep=0pt,outer sep=0pt] at ([xshift=196.887pt,yshift=-735.100pt]current page.north west) {{\fontsize{11.040pt}{11.702pt}\selectfont  authorization}};
\node[anchor=base west,inner sep=0pt,outer sep=0pt] at ([xshift=272.820pt,yshift=-735.100pt]current page.north west) {{\fontsize{11.040pt}{11.702pt}\selectfont  decisions}};
\node[anchor=base west,inner sep=0pt,outer sep=0pt] at ([xshift=331.696pt,yshift=-735.100pt]current page.north west) {{\fontsize{11.040pt}{11.702pt}\selectfont  in}};
\node[anchor=base west,inner sep=0pt,outer sep=0pt] at ([xshift=358.203pt,yshift=-735.100pt]current page.north west) {{\fontsize{11.040pt}{11.702pt}\selectfont  emerging}};
\node[anchor=base west,inner sep=0pt,outer sep=0pt] at ([xshift=417.687pt,yshift=-735.100pt]current page.north west) {{\fontsize{11.040pt}{11.702pt}\selectfont  agentic}};
\node[anchor=base west,inner sep=0pt,outer sep=0pt] at ([xshift=467.466pt,yshift=-735.100pt]current page.north west) {{\fontsize{11.040pt}{11.702pt}\selectfont  commerce}};
\node[anchor=base west,inner sep=0pt,outer sep=0pt] at ([xshift=82.104pt,yshift=-747.696pt]current page.north west) {{\fontsize{11.040pt}{11.702pt}\selectfont  \makebox[433.63pt][s]{environments.Addressing this gap is important for strengthening the security, transparency, and}}};
\end{tikzpicture}
\null
\newpage
\begin{tikzpicture}[remember picture,overlay]
\fill[black!15] ([xshift=91.944pt,yshift=-152.660pt]current page.north west) rectangle ([xshift=160.464pt,yshift=-187.100pt]current page.north west);
\fill[black!15] ([xshift=97.104pt,yshift=-152.660pt]current page.north west) rectangle ([xshift=155.304pt,yshift=-164.060pt]current page.north west);
\fill[black!15] ([xshift=160.940pt,yshift=-152.660pt]current page.north west) rectangle ([xshift=213.884pt,yshift=-187.100pt]current page.north west);
\fill[black!15] ([xshift=166.100pt,yshift=-152.660pt]current page.north west) rectangle ([xshift=208.724pt,yshift=-164.060pt]current page.north west);
\fill[black!15] ([xshift=166.100pt,yshift=-164.060pt]current page.north west) rectangle ([xshift=208.724pt,yshift=-175.580pt]current page.north west);
\fill[black!15] ([xshift=166.100pt,yshift=-175.580pt]current page.north west) rectangle ([xshift=208.724pt,yshift=-187.100pt]current page.north west);
\fill[black!15] ([xshift=214.370pt,yshift=-152.660pt]current page.north west) rectangle ([xshift=272.570pt,yshift=-187.100pt]current page.north west);
\fill[black!15] ([xshift=219.530pt,yshift=-152.660pt]current page.north west) rectangle ([xshift=267.410pt,yshift=-164.060pt]current page.north west);
\fill[black!15] ([xshift=219.530pt,yshift=-164.060pt]current page.north west) rectangle ([xshift=267.410pt,yshift=-175.580pt]current page.north west);
\fill[black!15] ([xshift=273.050pt,yshift=-152.660pt]current page.north west) rectangle ([xshift=334.874pt,yshift=-187.100pt]current page.north west);
\fill[black!15] ([xshift=278.210pt,yshift=-152.660pt]current page.north west) rectangle ([xshift=329.714pt,yshift=-164.060pt]current page.north west);
\fill[black!15] ([xshift=278.210pt,yshift=-164.060pt]current page.north west) rectangle ([xshift=329.714pt,yshift=-175.580pt]current page.north west);
\fill[black!15] ([xshift=335.350pt,yshift=-152.660pt]current page.north west) rectangle ([xshift=393.190pt,yshift=-187.100pt]current page.north west);
\fill[black!15] ([xshift=340.510pt,yshift=-152.660pt]current page.north west) rectangle ([xshift=388.030pt,yshift=-164.060pt]current page.north west);
\fill[black!15] ([xshift=340.510pt,yshift=-164.060pt]current page.north west) rectangle ([xshift=388.030pt,yshift=-175.580pt]current page.north west);
\fill[black!15] ([xshift=393.670pt,yshift=-152.660pt]current page.north west) rectangle ([xshift=503.490pt,yshift=-187.100pt]current page.north west);
\fill[black!15] ([xshift=398.830pt,yshift=-152.660pt]current page.north west) rectangle ([xshift=498.334pt,yshift=-164.060pt]current page.north west);
\fill[black!15] ([xshift=398.830pt,yshift=-164.060pt]current page.north west) rectangle ([xshift=498.334pt,yshift=-175.580pt]current page.north west);
\fill[black] ([xshift=91.464pt,yshift=-152.060pt]current page.north west) rectangle ([xshift=91.944pt,yshift=-152.540pt]current page.north west);
\fill[black] ([xshift=91.464pt,yshift=-152.060pt]current page.north west) rectangle ([xshift=91.944pt,yshift=-152.540pt]current page.north west);
\fill[black] ([xshift=91.944pt,yshift=-152.060pt]current page.north west) rectangle ([xshift=160.464pt,yshift=-152.540pt]current page.north west);
\fill[black] ([xshift=160.460pt,yshift=-152.060pt]current page.north west) rectangle ([xshift=160.940pt,yshift=-152.540pt]current page.north west);
\fill[black] ([xshift=160.940pt,yshift=-152.060pt]current page.north west) rectangle ([xshift=213.884pt,yshift=-152.540pt]current page.north west);
\fill[black] ([xshift=213.890pt,yshift=-152.060pt]current page.north west) rectangle ([xshift=214.370pt,yshift=-152.540pt]current page.north west);
\fill[black] ([xshift=214.370pt,yshift=-152.060pt]current page.north west) rectangle ([xshift=272.570pt,yshift=-152.540pt]current page.north west);
\fill[black] ([xshift=272.570pt,yshift=-152.060pt]current page.north west) rectangle ([xshift=273.050pt,yshift=-152.540pt]current page.north west);
\fill[black] ([xshift=273.050pt,yshift=-152.060pt]current page.north west) rectangle ([xshift=334.874pt,yshift=-152.540pt]current page.north west);
\fill[black] ([xshift=334.870pt,yshift=-152.060pt]current page.north west) rectangle ([xshift=335.350pt,yshift=-152.540pt]current page.north west);
\fill[black] ([xshift=335.350pt,yshift=-152.060pt]current page.north west) rectangle ([xshift=393.190pt,yshift=-152.540pt]current page.north west);
\fill[black] ([xshift=393.190pt,yshift=-152.060pt]current page.north west) rectangle ([xshift=393.670pt,yshift=-152.540pt]current page.north west);
\fill[black] ([xshift=393.670pt,yshift=-152.060pt]current page.north west) rectangle ([xshift=503.490pt,yshift=-152.540pt]current page.north west);
\fill[black] ([xshift=503.500pt,yshift=-152.060pt]current page.north west) rectangle ([xshift=503.980pt,yshift=-152.540pt]current page.north west);
\fill[black] ([xshift=503.500pt,yshift=-152.060pt]current page.north west) rectangle ([xshift=503.980pt,yshift=-152.540pt]current page.north west);
\fill[black] ([xshift=91.464pt,yshift=-152.540pt]current page.north west) rectangle ([xshift=91.944pt,yshift=-187.100pt]current page.north west);
\fill[black] ([xshift=160.460pt,yshift=-152.540pt]current page.north west) rectangle ([xshift=160.940pt,yshift=-187.100pt]current page.north west);
\fill[black] ([xshift=213.890pt,yshift=-152.540pt]current page.north west) rectangle ([xshift=214.370pt,yshift=-187.100pt]current page.north west);
\fill[black] ([xshift=272.570pt,yshift=-152.540pt]current page.north west) rectangle ([xshift=273.050pt,yshift=-187.100pt]current page.north west);
\fill[black] ([xshift=334.870pt,yshift=-152.540pt]current page.north west) rectangle ([xshift=335.350pt,yshift=-187.100pt]current page.north west);
\fill[black] ([xshift=393.190pt,yshift=-152.540pt]current page.north west) rectangle ([xshift=393.670pt,yshift=-187.100pt]current page.north west);
\fill[black] ([xshift=503.500pt,yshift=-152.540pt]current page.north west) rectangle ([xshift=503.980pt,yshift=-187.100pt]current page.north west);
\fill[black] ([xshift=91.464pt,yshift=-187.100pt]current page.north west) rectangle ([xshift=91.944pt,yshift=-187.580pt]current page.north west);
\fill[black] ([xshift=91.944pt,yshift=-187.100pt]current page.north west) rectangle ([xshift=160.464pt,yshift=-187.580pt]current page.north west);
\fill[black] ([xshift=160.460pt,yshift=-187.100pt]current page.north west) rectangle ([xshift=160.940pt,yshift=-187.580pt]current page.north west);
\fill[black] ([xshift=160.940pt,yshift=-187.100pt]current page.north west) rectangle ([xshift=213.884pt,yshift=-187.580pt]current page.north west);
\fill[black] ([xshift=213.890pt,yshift=-187.100pt]current page.north west) rectangle ([xshift=214.370pt,yshift=-187.580pt]current page.north west);
\fill[black] ([xshift=214.370pt,yshift=-187.100pt]current page.north west) rectangle ([xshift=272.570pt,yshift=-187.580pt]current page.north west);
\fill[black] ([xshift=272.570pt,yshift=-187.100pt]current page.north west) rectangle ([xshift=273.050pt,yshift=-187.580pt]current page.north west);
\fill[black] ([xshift=273.050pt,yshift=-187.100pt]current page.north west) rectangle ([xshift=334.874pt,yshift=-187.580pt]current page.north west);
\fill[black] ([xshift=334.870pt,yshift=-187.100pt]current page.north west) rectangle ([xshift=335.350pt,yshift=-187.580pt]current page.north west);
\fill[black] ([xshift=335.350pt,yshift=-187.100pt]current page.north west) rectangle ([xshift=393.190pt,yshift=-187.580pt]current page.north west);
\fill[black] ([xshift=393.190pt,yshift=-187.100pt]current page.north west) rectangle ([xshift=393.670pt,yshift=-187.580pt]current page.north west);
\fill[black] ([xshift=393.670pt,yshift=-187.100pt]current page.north west) rectangle ([xshift=503.490pt,yshift=-187.580pt]current page.north west);
\fill[black] ([xshift=503.500pt,yshift=-187.100pt]current page.north west) rectangle ([xshift=503.980pt,yshift=-187.580pt]current page.north west);
\fill[black] ([xshift=91.464pt,yshift=-187.580pt]current page.north west) rectangle ([xshift=91.944pt,yshift=-245.060pt]current page.north west);
\fill[black] ([xshift=160.460pt,yshift=-187.580pt]current page.north west) rectangle ([xshift=160.940pt,yshift=-245.060pt]current page.north west);
\fill[black] ([xshift=213.890pt,yshift=-187.580pt]current page.north west) rectangle ([xshift=214.370pt,yshift=-245.060pt]current page.north west);
\fill[black] ([xshift=272.570pt,yshift=-187.580pt]current page.north west) rectangle ([xshift=273.050pt,yshift=-245.060pt]current page.north west);
\fill[black] ([xshift=334.870pt,yshift=-187.580pt]current page.north west) rectangle ([xshift=335.350pt,yshift=-245.060pt]current page.north west);
\fill[black] ([xshift=393.190pt,yshift=-187.580pt]current page.north west) rectangle ([xshift=393.670pt,yshift=-245.060pt]current page.north west);
\fill[black] ([xshift=503.500pt,yshift=-187.580pt]current page.north west) rectangle ([xshift=503.980pt,yshift=-245.060pt]current page.north west);
\fill[black] ([xshift=91.464pt,yshift=-245.060pt]current page.north west) rectangle ([xshift=91.944pt,yshift=-245.540pt]current page.north west);
\fill[black] ([xshift=91.944pt,yshift=-245.060pt]current page.north west) rectangle ([xshift=160.464pt,yshift=-245.540pt]current page.north west);
\fill[black] ([xshift=160.460pt,yshift=-245.060pt]current page.north west) rectangle ([xshift=160.940pt,yshift=-245.540pt]current page.north west);
\fill[black] ([xshift=160.940pt,yshift=-245.060pt]current page.north west) rectangle ([xshift=213.884pt,yshift=-245.540pt]current page.north west);
\fill[black] ([xshift=213.890pt,yshift=-245.060pt]current page.north west) rectangle ([xshift=214.370pt,yshift=-245.540pt]current page.north west);
\fill[black] ([xshift=214.370pt,yshift=-245.060pt]current page.north west) rectangle ([xshift=272.570pt,yshift=-245.540pt]current page.north west);
\fill[black] ([xshift=272.570pt,yshift=-245.060pt]current page.north west) rectangle ([xshift=273.050pt,yshift=-245.540pt]current page.north west);
\fill[black] ([xshift=273.050pt,yshift=-245.060pt]current page.north west) rectangle ([xshift=334.874pt,yshift=-245.540pt]current page.north west);
\fill[black] ([xshift=334.870pt,yshift=-245.060pt]current page.north west) rectangle ([xshift=335.350pt,yshift=-245.540pt]current page.north west);
\fill[black] ([xshift=335.350pt,yshift=-245.060pt]current page.north west) rectangle ([xshift=393.190pt,yshift=-245.540pt]current page.north west);
\fill[black] ([xshift=393.190pt,yshift=-245.060pt]current page.north west) rectangle ([xshift=393.670pt,yshift=-245.540pt]current page.north west);
\fill[black] ([xshift=393.670pt,yshift=-245.060pt]current page.north west) rectangle ([xshift=503.490pt,yshift=-245.540pt]current page.north west);
\fill[black] ([xshift=503.500pt,yshift=-245.060pt]current page.north west) rectangle ([xshift=503.980pt,yshift=-245.540pt]current page.north west);
\fill[black] ([xshift=91.464pt,yshift=-245.540pt]current page.north west) rectangle ([xshift=91.944pt,yshift=-268.580pt]current page.north west);
\fill[black] ([xshift=160.460pt,yshift=-245.540pt]current page.north west) rectangle ([xshift=160.940pt,yshift=-268.580pt]current page.north west);
\fill[black] ([xshift=213.890pt,yshift=-245.540pt]current page.north west) rectangle ([xshift=214.370pt,yshift=-268.580pt]current page.north west);
\fill[black] ([xshift=272.570pt,yshift=-245.540pt]current page.north west) rectangle ([xshift=273.050pt,yshift=-268.580pt]current page.north west);
\fill[black] ([xshift=334.870pt,yshift=-245.540pt]current page.north west) rectangle ([xshift=335.350pt,yshift=-268.580pt]current page.north west);
\fill[black] ([xshift=393.190pt,yshift=-245.540pt]current page.north west) rectangle ([xshift=393.670pt,yshift=-268.580pt]current page.north west);
\fill[black] ([xshift=503.500pt,yshift=-245.540pt]current page.north west) rectangle ([xshift=503.980pt,yshift=-268.580pt]current page.north west);
\fill[black] ([xshift=91.464pt,yshift=-268.580pt]current page.north west) rectangle ([xshift=91.944pt,yshift=-269.060pt]current page.north west);
\fill[black] ([xshift=91.944pt,yshift=-268.580pt]current page.north west) rectangle ([xshift=160.464pt,yshift=-269.060pt]current page.north west);
\fill[black] ([xshift=160.460pt,yshift=-268.580pt]current page.north west) rectangle ([xshift=160.940pt,yshift=-269.060pt]current page.north west);
\fill[black] ([xshift=160.940pt,yshift=-268.580pt]current page.north west) rectangle ([xshift=213.884pt,yshift=-269.060pt]current page.north west);
\fill[black] ([xshift=213.890pt,yshift=-268.580pt]current page.north west) rectangle ([xshift=214.370pt,yshift=-269.060pt]current page.north west);
\fill[black] ([xshift=214.370pt,yshift=-268.580pt]current page.north west) rectangle ([xshift=272.570pt,yshift=-269.060pt]current page.north west);
\fill[black] ([xshift=272.570pt,yshift=-268.580pt]current page.north west) rectangle ([xshift=273.050pt,yshift=-269.060pt]current page.north west);
\fill[black] ([xshift=273.050pt,yshift=-268.580pt]current page.north west) rectangle ([xshift=334.874pt,yshift=-269.060pt]current page.north west);
\fill[black] ([xshift=334.870pt,yshift=-268.580pt]current page.north west) rectangle ([xshift=335.350pt,yshift=-269.060pt]current page.north west);
\fill[black] ([xshift=335.350pt,yshift=-268.580pt]current page.north west) rectangle ([xshift=393.190pt,yshift=-269.060pt]current page.north west);
\fill[black] ([xshift=393.190pt,yshift=-268.580pt]current page.north west) rectangle ([xshift=393.670pt,yshift=-269.060pt]current page.north west);
\fill[black] ([xshift=393.670pt,yshift=-268.580pt]current page.north west) rectangle ([xshift=503.490pt,yshift=-269.060pt]current page.north west);
\fill[black] ([xshift=503.500pt,yshift=-268.580pt]current page.north west) rectangle ([xshift=503.980pt,yshift=-269.060pt]current page.north west);
\fill[black] ([xshift=91.464pt,yshift=-269.066pt]current page.north west) rectangle ([xshift=91.944pt,yshift=-326.690pt]current page.north west);
\fill[black] ([xshift=160.460pt,yshift=-269.066pt]current page.north west) rectangle ([xshift=160.940pt,yshift=-326.690pt]current page.north west);
\fill[black] ([xshift=213.890pt,yshift=-269.066pt]current page.north west) rectangle ([xshift=214.370pt,yshift=-326.690pt]current page.north west);
\fill[black] ([xshift=272.570pt,yshift=-269.066pt]current page.north west) rectangle ([xshift=273.050pt,yshift=-326.690pt]current page.north west);
\fill[black] ([xshift=334.870pt,yshift=-269.066pt]current page.north west) rectangle ([xshift=335.350pt,yshift=-326.690pt]current page.north west);
\fill[black] ([xshift=393.190pt,yshift=-269.066pt]current page.north west) rectangle ([xshift=393.670pt,yshift=-326.690pt]current page.north west);
\fill[black] ([xshift=503.500pt,yshift=-269.066pt]current page.north west) rectangle ([xshift=503.980pt,yshift=-326.690pt]current page.north west);
\fill[black] ([xshift=91.464pt,yshift=-326.690pt]current page.north west) rectangle ([xshift=91.944pt,yshift=-327.170pt]current page.north west);
\fill[black] ([xshift=91.944pt,yshift=-326.690pt]current page.north west) rectangle ([xshift=160.464pt,yshift=-327.170pt]current page.north west);
\fill[black] ([xshift=160.460pt,yshift=-326.690pt]current page.north west) rectangle ([xshift=160.940pt,yshift=-327.170pt]current page.north west);
\fill[black] ([xshift=160.940pt,yshift=-326.690pt]current page.north west) rectangle ([xshift=213.884pt,yshift=-327.170pt]current page.north west);
\fill[black] ([xshift=213.890pt,yshift=-326.690pt]current page.north west) rectangle ([xshift=214.370pt,yshift=-327.170pt]current page.north west);
\fill[black] ([xshift=214.370pt,yshift=-326.690pt]current page.north west) rectangle ([xshift=272.570pt,yshift=-327.170pt]current page.north west);
\fill[black] ([xshift=272.570pt,yshift=-326.690pt]current page.north west) rectangle ([xshift=273.050pt,yshift=-327.170pt]current page.north west);
\fill[black] ([xshift=273.050pt,yshift=-326.690pt]current page.north west) rectangle ([xshift=334.874pt,yshift=-327.170pt]current page.north west);
\fill[black] ([xshift=334.870pt,yshift=-326.690pt]current page.north west) rectangle ([xshift=335.350pt,yshift=-327.170pt]current page.north west);
\fill[black] ([xshift=335.350pt,yshift=-326.690pt]current page.north west) rectangle ([xshift=393.190pt,yshift=-327.170pt]current page.north west);
\fill[black] ([xshift=393.190pt,yshift=-326.690pt]current page.north west) rectangle ([xshift=393.670pt,yshift=-327.170pt]current page.north west);
\fill[black] ([xshift=393.670pt,yshift=-326.690pt]current page.north west) rectangle ([xshift=503.490pt,yshift=-327.170pt]current page.north west);
\fill[black] ([xshift=503.500pt,yshift=-326.690pt]current page.north west) rectangle ([xshift=503.980pt,yshift=-327.170pt]current page.north west);
\fill[black] ([xshift=91.464pt,yshift=-327.170pt]current page.north west) rectangle ([xshift=91.944pt,yshift=-384.650pt]current page.north west);
\fill[black] ([xshift=160.460pt,yshift=-327.170pt]current page.north west) rectangle ([xshift=160.940pt,yshift=-384.650pt]current page.north west);
\fill[black] ([xshift=213.890pt,yshift=-327.170pt]current page.north west) rectangle ([xshift=214.370pt,yshift=-384.650pt]current page.north west);
\fill[black] ([xshift=272.570pt,yshift=-327.170pt]current page.north west) rectangle ([xshift=273.050pt,yshift=-384.650pt]current page.north west);
\fill[black] ([xshift=334.870pt,yshift=-327.170pt]current page.north west) rectangle ([xshift=335.350pt,yshift=-384.650pt]current page.north west);
\fill[black] ([xshift=393.190pt,yshift=-327.170pt]current page.north west) rectangle ([xshift=393.670pt,yshift=-384.650pt]current page.north west);
\fill[black] ([xshift=503.500pt,yshift=-327.170pt]current page.north west) rectangle ([xshift=503.980pt,yshift=-384.650pt]current page.north west);
\fill[black] ([xshift=91.464pt,yshift=-384.650pt]current page.north west) rectangle ([xshift=91.944pt,yshift=-385.130pt]current page.north west);
\fill[black] ([xshift=91.944pt,yshift=-384.650pt]current page.north west) rectangle ([xshift=160.464pt,yshift=-385.130pt]current page.north west);
\fill[black] ([xshift=160.460pt,yshift=-384.650pt]current page.north west) rectangle ([xshift=160.940pt,yshift=-385.130pt]current page.north west);
\fill[black] ([xshift=160.940pt,yshift=-384.650pt]current page.north west) rectangle ([xshift=213.884pt,yshift=-385.130pt]current page.north west);
\fill[black] ([xshift=213.890pt,yshift=-384.650pt]current page.north west) rectangle ([xshift=214.370pt,yshift=-385.130pt]current page.north west);
\fill[black] ([xshift=214.370pt,yshift=-384.650pt]current page.north west) rectangle ([xshift=272.570pt,yshift=-385.130pt]current page.north west);
\fill[black] ([xshift=272.570pt,yshift=-384.650pt]current page.north west) rectangle ([xshift=273.050pt,yshift=-385.130pt]current page.north west);
\fill[black] ([xshift=273.050pt,yshift=-384.650pt]current page.north west) rectangle ([xshift=334.874pt,yshift=-385.130pt]current page.north west);
\fill[black] ([xshift=334.870pt,yshift=-384.650pt]current page.north west) rectangle ([xshift=335.350pt,yshift=-385.130pt]current page.north west);
\fill[black] ([xshift=335.350pt,yshift=-384.650pt]current page.north west) rectangle ([xshift=393.190pt,yshift=-385.130pt]current page.north west);
\fill[black] ([xshift=393.190pt,yshift=-384.650pt]current page.north west) rectangle ([xshift=393.670pt,yshift=-385.130pt]current page.north west);
\fill[black] ([xshift=393.670pt,yshift=-384.650pt]current page.north west) rectangle ([xshift=503.490pt,yshift=-385.130pt]current page.north west);
\fill[black] ([xshift=503.500pt,yshift=-384.650pt]current page.north west) rectangle ([xshift=503.980pt,yshift=-385.130pt]current page.north west);
\fill[black] ([xshift=91.464pt,yshift=-385.130pt]current page.north west) rectangle ([xshift=91.944pt,yshift=-408.170pt]current page.north west);
\fill[black] ([xshift=91.464pt,yshift=-408.170pt]current page.north west) rectangle ([xshift=91.944pt,yshift=-408.650pt]current page.north west);
\fill[black] ([xshift=91.464pt,yshift=-408.170pt]current page.north west) rectangle ([xshift=91.944pt,yshift=-408.650pt]current page.north west);
\fill[black] ([xshift=91.944pt,yshift=-408.170pt]current page.north west) rectangle ([xshift=160.464pt,yshift=-408.650pt]current page.north west);
\fill[black] ([xshift=160.460pt,yshift=-385.130pt]current page.north west) rectangle ([xshift=160.940pt,yshift=-408.170pt]current page.north west);
\fill[black] ([xshift=160.460pt,yshift=-408.170pt]current page.north west) rectangle ([xshift=160.940pt,yshift=-408.650pt]current page.north west);
\fill[black] ([xshift=160.940pt,yshift=-408.170pt]current page.north west) rectangle ([xshift=213.884pt,yshift=-408.650pt]current page.north west);
\fill[black] ([xshift=213.890pt,yshift=-385.130pt]current page.north west) rectangle ([xshift=214.370pt,yshift=-408.170pt]current page.north west);
\fill[black] ([xshift=213.890pt,yshift=-408.170pt]current page.north west) rectangle ([xshift=214.370pt,yshift=-408.650pt]current page.north west);
\fill[black] ([xshift=214.370pt,yshift=-408.170pt]current page.north west) rectangle ([xshift=272.570pt,yshift=-408.650pt]current page.north west);
\fill[black] ([xshift=272.570pt,yshift=-385.130pt]current page.north west) rectangle ([xshift=273.050pt,yshift=-408.170pt]current page.north west);
\fill[black] ([xshift=272.570pt,yshift=-408.170pt]current page.north west) rectangle ([xshift=273.050pt,yshift=-408.650pt]current page.north west);
\fill[black] ([xshift=273.050pt,yshift=-408.170pt]current page.north west) rectangle ([xshift=334.874pt,yshift=-408.650pt]current page.north west);
\fill[black] ([xshift=334.870pt,yshift=-385.130pt]current page.north west) rectangle ([xshift=335.350pt,yshift=-408.170pt]current page.north west);
\fill[black] ([xshift=334.870pt,yshift=-408.170pt]current page.north west) rectangle ([xshift=335.350pt,yshift=-408.650pt]current page.north west);
\fill[black] ([xshift=335.350pt,yshift=-408.170pt]current page.north west) rectangle ([xshift=393.190pt,yshift=-408.650pt]current page.north west);
\fill[black] ([xshift=393.190pt,yshift=-385.130pt]current page.north west) rectangle ([xshift=393.670pt,yshift=-408.170pt]current page.north west);
\fill[black] ([xshift=393.190pt,yshift=-408.170pt]current page.north west) rectangle ([xshift=393.670pt,yshift=-408.650pt]current page.north west);
\fill[black] ([xshift=393.670pt,yshift=-408.170pt]current page.north west) rectangle ([xshift=503.490pt,yshift=-408.650pt]current page.north west);
\fill[black] ([xshift=503.500pt,yshift=-385.130pt]current page.north west) rectangle ([xshift=503.980pt,yshift=-408.170pt]current page.north west);
\fill[black] ([xshift=503.500pt,yshift=-408.170pt]current page.north west) rectangle ([xshift=503.980pt,yshift=-408.650pt]current page.north west);
\fill[black] ([xshift=503.500pt,yshift=-408.170pt]current page.north west) rectangle ([xshift=503.980pt,yshift=-408.650pt]current page.north west);
\node[anchor=base west,inner sep=0pt,outer sep=0pt] at ([xshift=127.580pt,yshift=-81.360pt]current page.north west) {{\fontsize{9.960pt}{10.558pt}\selectfont  \makebox[342.77pt][s]{International Journal on Cybernetics \& Informatics (IJCI) Vol.15, No.4, August 2026}}};
\node[anchor=base west,inner sep=0pt,outer sep=0pt] at ([xshift=503.380pt,yshift=-767.856pt]current page.north west) {{\fontsize{9.960pt}{10.558pt}\selectfont  87}};
\node[anchor=base west,inner sep=0pt,outer sep=0pt] at ([xshift=82.104pt,yshift=-100.220pt]current page.north west) {{\fontsize{11.040pt}{11.702pt}\selectfont  \makebox[433.75pt][s]{trustworthiness of next-generation digital payment systems as autonomous agents assume greater}}};
\node[anchor=base west,inner sep=0pt,outer sep=0pt] at ([xshift=82.104pt,yshift=-112.820pt]current page.north west) {{\fontsize{11.040pt}{11.702pt}\selectfont  \makebox[229.83pt][s]{roles in commercial transactions [3], [4], [11]--[13].}}};
\node[anchor=base west,inner sep=0pt,outer sep=0pt] at ([xshift=125.540pt,yshift=-137.300pt]current page.north west) {{\fontsize{9.960pt}{10.558pt}\selectfont  \makebox[346.73pt][s]{Table 1. Comparison of Representative Studies and the Proposed ACAGS Framework}}};
\node[anchor=base west,inner sep=0pt,outer sep=0pt] at ([xshift=97.104pt,yshift=-161.900pt]current page.north west) {{\fontsize{9.960pt}{10.558pt}\selectfont \bfseries Study}};
\node[anchor=base west,inner sep=0pt,outer sep=0pt] at ([xshift=166.100pt,yshift=-161.900pt]current page.north west) {{\fontsize{9.960pt}{10.558pt}\selectfont \bfseries AI-Based}};
\node[anchor=base west,inner sep=0pt,outer sep=0pt] at ([xshift=166.100pt,yshift=-173.420pt]current page.north west) {{\fontsize{9.960pt}{10.558pt}\selectfont \bfseries Fraud}};
\node[anchor=base west,inner sep=0pt,outer sep=0pt] at ([xshift=166.100pt,yshift=-184.940pt]current page.north west) {{\fontsize{9.960pt}{10.558pt}\selectfont \bfseries Detection}};
\node[anchor=base west,inner sep=0pt,outer sep=0pt] at ([xshift=219.530pt,yshift=-161.900pt]current page.north west) {{\fontsize{9.960pt}{10.558pt}\selectfont \bfseries Behavioral}};
\node[anchor=base west,inner sep=0pt,outer sep=0pt] at ([xshift=219.530pt,yshift=-173.420pt]current page.north west) {{\fontsize{9.960pt}{10.558pt}\selectfont \bfseries Analytics}};
\node[anchor=base west,inner sep=0pt,outer sep=0pt] at ([xshift=278.210pt,yshift=-161.900pt]current page.north west) {{\fontsize{9.960pt}{10.558pt}\selectfont \bfseries Explainable}};
\node[anchor=base west,inner sep=0pt,outer sep=0pt] at ([xshift=278.210pt,yshift=-173.420pt]current page.north west) {{\fontsize{9.960pt}{10.558pt}\selectfont \bfseries AI}};
\node[anchor=base west,inner sep=0pt,outer sep=0pt] at ([xshift=340.510pt,yshift=-161.900pt]current page.north west) {{\fontsize{9.960pt}{10.558pt}\selectfont \bfseries Agentic}};
\node[anchor=base west,inner sep=0pt,outer sep=0pt] at ([xshift=340.510pt,yshift=-173.420pt]current page.north west) {{\fontsize{9.960pt}{10.558pt}\selectfont \bfseries Commerce}};
\node[anchor=base west,inner sep=0pt,outer sep=0pt] at ([xshift=398.830pt,yshift=-161.900pt]current page.north west) {{\fontsize{9.960pt}{10.558pt}\selectfont \bfseries Security/Authorization}};
\node[anchor=base west,inner sep=0pt,outer sep=0pt] at ([xshift=398.830pt,yshift=-173.420pt]current page.north west) {{\fontsize{9.960pt}{10.558pt}\selectfont \bfseries Controls}};
\node[anchor=base west,inner sep=0pt,outer sep=0pt] at ([xshift=97.104pt,yshift=-196.940pt]current page.north west) {{\fontsize{9.960pt}{10.558pt}\selectfont  Established}};
\node[anchor=base west,inner sep=0pt,outer sep=0pt] at ([xshift=97.104pt,yshift=-208.460pt]current page.north west) {{\fontsize{9.960pt}{10.558pt}\selectfont  fraud-}};
\node[anchor=base west,inner sep=0pt,outer sep=0pt] at ([xshift=97.104pt,yshift=-219.980pt]current page.north west) {{\fontsize{9.960pt}{10.558pt}\selectfont  detection}};
\node[anchor=base west,inner sep=0pt,outer sep=0pt] at ([xshift=97.104pt,yshift=-231.500pt]current page.north west) {{\fontsize{9.960pt}{10.558pt}\selectfont  studies}};
\node[anchor=base west,inner sep=0pt,outer sep=0pt] at ([xshift=140.998pt,yshift=-231.500pt]current page.north west) {{\fontsize{9.960pt}{10.558pt}\selectfont  [9],}};
\node[anchor=base west,inner sep=0pt,outer sep=0pt] at ([xshift=97.104pt,yshift=-242.900pt]current page.north west) {{\fontsize{9.960pt}{10.558pt}\selectfont  [10]}};
\node[anchor=base west,inner sep=0pt,outer sep=0pt] at ([xshift=166.100pt,yshift=-198.380pt]current page.north west) {{\fontsize{9.960pt}{10.558pt}\selectfont  $\checkmark$}};
\node[anchor=base west,inner sep=0pt,outer sep=0pt] at ([xshift=219.530pt,yshift=-196.940pt]current page.north west) {{\fontsize{9.960pt}{10.558pt}\selectfont  Partial}};
\node[anchor=base west,inner sep=0pt,outer sep=0pt] at ([xshift=278.210pt,yshift=-198.380pt]current page.north west) {{\fontsize{9.960pt}{10.558pt}\selectfont  \textsf{X}}};
\node[anchor=base west,inner sep=0pt,outer sep=0pt] at ([xshift=340.510pt,yshift=-198.380pt]current page.north west) {{\fontsize{9.960pt}{10.558pt}\selectfont  \textsf{X}}};
\node[anchor=base west,inner sep=0pt,outer sep=0pt] at ([xshift=398.830pt,yshift=-196.940pt]current page.north west) {{\fontsize{9.960pt}{10.558pt}\selectfont  Partial}};
\node[anchor=base west,inner sep=0pt,outer sep=0pt] at ([xshift=97.104pt,yshift=-254.900pt]current page.north west) {{\fontsize{9.960pt}{10.558pt}\selectfont  Explainable}};
\node[anchor=base west,inner sep=0pt,outer sep=0pt] at ([xshift=97.104pt,yshift=-266.420pt]current page.north west) {{\fontsize{9.960pt}{10.558pt}\selectfont  AI method [8]}};
\node[anchor=base west,inner sep=0pt,outer sep=0pt] at ([xshift=166.100pt,yshift=-256.340pt]current page.north west) {{\fontsize{9.960pt}{10.558pt}\selectfont  \textsf{X}}};
\node[anchor=base west,inner sep=0pt,outer sep=0pt] at ([xshift=219.530pt,yshift=-256.340pt]current page.north west) {{\fontsize{9.960pt}{10.558pt}\selectfont  \textsf{X}}};
\node[anchor=base west,inner sep=0pt,outer sep=0pt] at ([xshift=278.210pt,yshift=-256.340pt]current page.north west) {{\fontsize{9.960pt}{10.558pt}\selectfont  $\checkmark$}};
\node[anchor=base west,inner sep=0pt,outer sep=0pt] at ([xshift=340.510pt,yshift=-256.340pt]current page.north west) {{\fontsize{9.960pt}{10.558pt}\selectfont  \textsf{X}}};
\node[anchor=base west,inner sep=0pt,outer sep=0pt] at ([xshift=398.830pt,yshift=-256.340pt]current page.north west) {{\fontsize{9.960pt}{10.558pt}\selectfont  \textsf{X}}};
\node[anchor=base west,inner sep=0pt,outer sep=0pt] at ([xshift=97.104pt,yshift=-278.420pt]current page.north west) {{\fontsize{9.960pt}{10.558pt}\selectfont  Agentic}};
\node[anchor=base west,inner sep=0pt,outer sep=0pt] at ([xshift=97.104pt,yshift=-289.970pt]current page.north west) {{\fontsize{9.960pt}{10.558pt}\selectfont  commerce}};
\node[anchor=base west,inner sep=0pt,outer sep=0pt] at ([xshift=97.104pt,yshift=-301.490pt]current page.north west) {{\fontsize{9.960pt}{10.558pt}\selectfont  infrastructure}};
\node[anchor=base west,inner sep=0pt,outer sep=0pt] at ([xshift=97.104pt,yshift=-313.010pt]current page.north west) {{\fontsize{9.960pt}{10.558pt}\selectfont  studies [11]--}};
\node[anchor=base west,inner sep=0pt,outer sep=0pt] at ([xshift=97.104pt,yshift=-324.530pt]current page.north west) {{\fontsize{9.960pt}{10.558pt}\selectfont  [13]}};
\node[anchor=base west,inner sep=0pt,outer sep=0pt] at ([xshift=166.100pt,yshift=-279.860pt]current page.north west) {{\fontsize{9.960pt}{10.558pt}\selectfont  \textsf{X}}};
\node[anchor=base west,inner sep=0pt,outer sep=0pt] at ([xshift=219.530pt,yshift=-278.420pt]current page.north west) {{\fontsize{9.960pt}{10.558pt}\selectfont  Partial}};
\node[anchor=base west,inner sep=0pt,outer sep=0pt] at ([xshift=278.210pt,yshift=-279.860pt]current page.north west) {{\fontsize{9.960pt}{10.558pt}\selectfont  \textsf{X}}};
\node[anchor=base west,inner sep=0pt,outer sep=0pt] at ([xshift=340.510pt,yshift=-279.860pt]current page.north west) {{\fontsize{9.960pt}{10.558pt}\selectfont  $\checkmark$}};
\node[anchor=base west,inner sep=0pt,outer sep=0pt] at ([xshift=398.830pt,yshift=-279.860pt]current page.north west) {{\fontsize{9.960pt}{10.558pt}\selectfont  $\checkmark$}};
\node[anchor=base west,inner sep=0pt,outer sep=0pt] at ([xshift=97.104pt,yshift=-336.410pt]current page.north west) {{\fontsize{9.960pt}{10.558pt}\selectfont  Agentic}};
\node[anchor=base west,inner sep=0pt,outer sep=0pt] at ([xshift=97.104pt,yshift=-347.930pt]current page.north west) {{\fontsize{9.960pt}{10.558pt}\selectfont  payment and}};
\node[anchor=base west,inner sep=0pt,outer sep=0pt] at ([xshift=97.104pt,yshift=-359.450pt]current page.north west) {{\fontsize{9.960pt}{10.558pt}\selectfont  authorization}};
\node[anchor=base west,inner sep=0pt,outer sep=0pt] at ([xshift=97.104pt,yshift=-370.970pt]current page.north west) {{\fontsize{9.960pt}{10.558pt}\selectfont  studies}};
\node[anchor=base west,inner sep=0pt,outer sep=0pt] at ([xshift=140.998pt,yshift=-370.970pt]current page.north west) {{\fontsize{9.960pt}{10.558pt}\selectfont  [3],}};
\node[anchor=base west,inner sep=0pt,outer sep=0pt] at ([xshift=97.104pt,yshift=-382.490pt]current page.north west) {{\fontsize{9.960pt}{10.558pt}\selectfont  [4]}};
\node[anchor=base west,inner sep=0pt,outer sep=0pt] at ([xshift=166.100pt,yshift=-337.970pt]current page.north west) {{\fontsize{9.960pt}{10.558pt}\selectfont  \textsf{X}}};
\node[anchor=base west,inner sep=0pt,outer sep=0pt] at ([xshift=219.530pt,yshift=-337.970pt]current page.north west) {{\fontsize{9.960pt}{10.558pt}\selectfont  \textsf{X}}};
\node[anchor=base west,inner sep=0pt,outer sep=0pt] at ([xshift=278.210pt,yshift=-337.970pt]current page.north west) {{\fontsize{9.960pt}{10.558pt}\selectfont  \textsf{X}}};
\node[anchor=base west,inner sep=0pt,outer sep=0pt] at ([xshift=340.510pt,yshift=-337.970pt]current page.north west) {{\fontsize{9.960pt}{10.558pt}\selectfont  $\checkmark$}};
\node[anchor=base west,inner sep=0pt,outer sep=0pt] at ([xshift=398.830pt,yshift=-337.970pt]current page.north west) {{\fontsize{9.960pt}{10.558pt}\selectfont  $\checkmark$}};
\node[anchor=base west,inner sep=0pt,outer sep=0pt] at ([xshift=97.104pt,yshift=-394.490pt]current page.north west) {{\fontsize{9.960pt}{10.558pt}\selectfont  Proposed}};
\node[anchor=base west,inner sep=0pt,outer sep=0pt] at ([xshift=97.104pt,yshift=-406.010pt]current page.north west) {{\fontsize{9.960pt}{10.558pt}\selectfont  ACAGS}};
\node[anchor=base west,inner sep=0pt,outer sep=0pt] at ([xshift=166.100pt,yshift=-395.930pt]current page.north west) {{\fontsize{9.960pt}{10.558pt}\selectfont  $\checkmark$}};
\node[anchor=base west,inner sep=0pt,outer sep=0pt] at ([xshift=219.530pt,yshift=-395.930pt]current page.north west) {{\fontsize{9.960pt}{10.558pt}\selectfont  $\checkmark$}};
\node[anchor=base west,inner sep=0pt,outer sep=0pt] at ([xshift=278.210pt,yshift=-395.930pt]current page.north west) {{\fontsize{9.960pt}{10.558pt}\selectfont  $\checkmark$}};
\node[anchor=base west,inner sep=0pt,outer sep=0pt] at ([xshift=340.510pt,yshift=-395.930pt]current page.north west) {{\fontsize{9.960pt}{10.558pt}\selectfont  $\checkmark$}};
\node[anchor=base west,inner sep=0pt,outer sep=0pt] at ([xshift=398.830pt,yshift=-395.930pt]current page.north west) {{\fontsize{9.960pt}{10.558pt}\selectfont  $\checkmark$}};
\node[anchor=base west,inner sep=0pt,outer sep=0pt] at ([xshift=82.104pt,yshift=-431.450pt]current page.north west) {{\fontsize{11.040pt}{11.702pt}\selectfont  \makebox[433.61pt][s]{The comparative analysis indicates that existing research generally addresses fraud detection,}}};
\node[anchor=base west,inner sep=0pt,outer sep=0pt] at ([xshift=82.104pt,yshift=-444.170pt]current page.north west) {{\fontsize{11.040pt}{11.702pt}\selectfont  \makebox[433.61pt][s]{explainability, agentic payment infrastructure, or security and authorization controls as distinct}}};
\node[anchor=base west,inner sep=0pt,outer sep=0pt] at ([xshift=82.104pt,yshift=-456.770pt]current page.north west) {{\fontsize{11.040pt}{11.702pt}\selectfont  \makebox[433.65pt][s]{areas. In contrast, the proposed Autonomous Commerce Agent Guard System (ACAGS)}}};
\node[anchor=base west,inner sep=0pt,outer sep=0pt] at ([xshift=82.104pt,yshift=-469.370pt]current page.north west) {{\fontsize{11.040pt}{11.702pt}\selectfont  \makebox[433.67pt][s]{integrates these complementary capabilities into a unified cybersecurity framework designed to}}};
\node[anchor=base west,inner sep=0pt,outer sep=0pt] at ([xshift=82.104pt,yshift=-482.110pt]current page.north west) {{\fontsize{11.040pt}{11.702pt}\selectfont  \makebox[434.00pt][s]{strengthen behavior-aware fraud detection for agentic commerce payment authorization systems.}}};
\node[anchor=base west,inner sep=0pt,outer sep=0pt] at ([xshift=82.104pt,yshift=-494.710pt]current page.north west) {{\fontsize{11.040pt}{11.702pt}\selectfont  \makebox[427.16pt][s]{The following section presents the architecture and key components of the proposed framework.}}};
\node[anchor=base west,inner sep=0pt,outer sep=0pt] at ([xshift=82.104pt,yshift=-522.910pt]current page.north west) {{\fontsize{14.040pt}{14.882pt}\selectfont \bfseries 3.}};
\node[anchor=base west,inner sep=0pt,outer sep=0pt] at ([xshift=100.100pt,yshift=-522.910pt]current page.north west) {{\fontsize{14.040pt}{14.882pt}\selectfont \bfseries P}};
\node[anchor=base west,inner sep=0pt,outer sep=0pt] at ([xshift=108.740pt,yshift=-522.910pt]current page.north west) {{\fontsize{11.040pt}{11.702pt}\selectfont \bfseries ROPOSED}};
\node[anchor=base west,inner sep=0pt,outer sep=0pt] at ([xshift=164.060pt,yshift=-522.910pt]current page.north west) {{\fontsize{14.040pt}{14.882pt}\selectfont \bfseries A}};
\node[anchor=base west,inner sep=0pt,outer sep=0pt] at ([xshift=174.260pt,yshift=-522.910pt]current page.north west) {{\fontsize{11.040pt}{11.702pt}\selectfont \bfseries CAGS}};
\node[anchor=base west,inner sep=0pt,outer sep=0pt] at ([xshift=207.410pt,yshift=-522.910pt]current page.north west) {{\fontsize{14.040pt}{14.882pt}\selectfont \bfseries F}};
\node[anchor=base west,inner sep=0pt,outer sep=0pt] at ([xshift=215.930pt,yshift=-522.910pt]current page.north west) {{\fontsize{11.040pt}{11.702pt}\selectfont \bfseries RAMEWORK}};
\node[anchor=base west,inner sep=0pt,outer sep=0pt] at ([xshift=82.104pt,yshift=-548.830pt]current page.north west) {{\fontsize{11.040pt}{11.702pt}\selectfont  \makebox[433.63pt][s]{The increasing adoption of autonomous AI agents in digital commerce requires cybersecurity}}};
\node[anchor=base west,inner sep=0pt,outer sep=0pt] at ([xshift=82.104pt,yshift=-561.430pt]current page.north west) {{\fontsize{11.040pt}{11.702pt}\selectfont  \makebox[433.74pt][s]{solutions that extend beyond conventional payment fraud detection. Existing payment}}};
\node[anchor=base west,inner sep=0pt,outer sep=0pt] at ([xshift=82.104pt,yshift=-574.030pt]current page.north west) {{\fontsize{11.040pt}{11.702pt}\selectfont  \makebox[433.64pt][s]{authorization mechanisms primarily evaluate transaction legitimacy using predefined rules or}}};
\node[anchor=base west,inner sep=0pt,outer sep=0pt] at ([xshift=82.104pt,yshift=-586.750pt]current page.north west) {{\fontsize{11.040pt}{11.702pt}\selectfont  \makebox[433.70pt][s]{predictive models, but they provide limited capability to continuously assess behavioral}}};
\node[anchor=base west,inner sep=0pt,outer sep=0pt] at ([xshift=82.104pt,yshift=-599.350pt]current page.north west) {{\fontsize{11.040pt}{11.702pt}\selectfont  \makebox[433.68pt][s]{consistency within autonomous purchasing environments. To address this challenge, this paper}}};
\node[anchor=base west,inner sep=0pt,outer sep=0pt] at ([xshift=82.104pt,yshift=-612.070pt]current page.north west) {{\fontsize{11.040pt}{11.702pt}\selectfont  proposes the}};
\node[anchor=base west,inner sep=0pt,outer sep=0pt] at ([xshift=148.460pt,yshift=-612.070pt]current page.north west) {{\fontsize{11.040pt}{11.702pt}\selectfont \bfseries \makebox[284.11pt][s]{Autonomous Commerce Agent Guard System (ACAGS)}}};
\node[anchor=base west,inner sep=0pt,outer sep=0pt] at ([xshift=432.790pt,yshift=-612.070pt]current page.north west) {{\fontsize{11.040pt}{11.702pt}\selectfont  \makebox[83.22pt][s]{, a cybersecurity}}};
\node[anchor=base west,inner sep=0pt,outer sep=0pt] at ([xshift=82.104pt,yshift=-624.670pt]current page.north west) {{\fontsize{11.040pt}{11.702pt}\selectfont  \makebox[433.80pt][s]{framework designed to strengthen behavior-aware fraud detection for payment authorization}}};
\node[anchor=base west,inner sep=0pt,outer sep=0pt] at ([xshift=82.104pt,yshift=-637.270pt]current page.north west) {{\fontsize{11.040pt}{11.702pt}\selectfont  \makebox[431.31pt][s]{systems operating in agentic commerce.ACAGS integrates behavioral and agent intelligence, AI-}}};
\node[anchor=base west,inner sep=0pt,outer sep=0pt] at ([xshift=82.104pt,yshift=-649.990pt]current page.north west) {{\fontsize{11.040pt}{11.702pt}\selectfont  \makebox[433.76pt][s]{based fraud detection, agent-specific authorization controls, and explainable decision support into}}};
\node[anchor=base west,inner sep=0pt,outer sep=0pt] at ([xshift=82.104pt,yshift=-662.620pt]current page.north west) {{\fontsize{11.040pt}{11.702pt}\selectfont  \makebox[320.09pt][s]{a unified framework for secure and interpretable payment authorization.}}};
\node[anchor=base west,inner sep=0pt,outer sep=0pt] at ([xshift=82.104pt,yshift=-687.940pt]current page.north west) {{\fontsize{11.040pt}{11.702pt}\selectfont  \makebox[433.60pt][s]{The framework is designed as a modular architecture that analyzes transaction behavior before}}};
\node[anchor=base west,inner sep=0pt,outer sep=0pt] at ([xshift=82.104pt,yshift=-700.540pt]current page.north west) {{\fontsize{11.040pt}{11.702pt}\selectfont  \makebox[433.64pt][s]{supporting payment authorization decisions. Rather than relying solely on static transaction}}};
\node[anchor=base west,inner sep=0pt,outer sep=0pt] at ([xshift=82.104pt,yshift=-713.260pt]current page.north west) {{\fontsize{11.040pt}{11.702pt}\selectfont  \makebox[433.86pt][s]{attributes, ACAGS evaluates contextual behavioral characteristics---including spending patterns,}}};
\node[anchor=base west,inner sep=0pt,outer sep=0pt] at ([xshift=82.104pt,yshift=-725.860pt]current page.north west) {{\fontsize{11.040pt}{11.702pt}\selectfont  \makebox[433.92pt][s]{temporal activity, merchant interactions, and customer transaction history---to identify anomalous}}};
\node[anchor=base west,inner sep=0pt,outer sep=0pt] at ([xshift=82.104pt,yshift=-738.580pt]current page.north west) {{\fontsize{11.040pt}{11.702pt}\selectfont  \makebox[433.86pt][s]{payment behavior that may indicate fraudulent activity.By combining behavioral intelligence,}}};
\end{tikzpicture}
\null
\newpage
\begin{tikzpicture}[remember picture,overlay]
\node[anchor=base west,inner sep=0pt,outer sep=0pt] at ([xshift=127.580pt,yshift=-81.360pt]current page.north west) {{\fontsize{9.960pt}{10.558pt}\selectfont  \makebox[342.77pt][s]{International Journal on Cybernetics \& Informatics (IJCI) Vol.15, No.4, August 2026}}};
\node[anchor=base west,inner sep=0pt,outer sep=0pt] at ([xshift=503.380pt,yshift=-767.856pt]current page.north west) {{\fontsize{9.960pt}{10.558pt}\selectfont  88}};
\node[anchor=base west,inner sep=0pt,outer sep=0pt] at ([xshift=82.104pt,yshift=-100.220pt]current page.north west) {{\fontsize{11.040pt}{11.702pt}\selectfont  \makebox[433.98pt][s]{agent-specific authorization controls, and explainable AI, the framework supports fraud-risk}}};
\node[anchor=base west,inner sep=0pt,outer sep=0pt] at ([xshift=82.104pt,yshift=-112.820pt]current page.north west) {{\fontsize{11.040pt}{11.702pt}\selectfont  \makebox[382.88pt][s]{assessment and interpretable authorization within autonomous payment environments.}}};
\node[anchor=base west,inner sep=0pt,outer sep=0pt] at ([xshift=82.104pt,yshift=-139.100pt]current page.north west) {{\fontsize{12.000pt}{12.720pt}\selectfont \bfseries 3.1.}};
\node[anchor=base west,inner sep=0pt,outer sep=0pt] at ([xshift=104.660pt,yshift=-139.100pt]current page.north west) {{\fontsize{12.000pt}{12.720pt}\selectfont \bfseries \makebox[91.52pt][s]{Design Objectives}}};
\node[anchor=base west,inner sep=0pt,outer sep=0pt] at ([xshift=82.104pt,yshift=-164.660pt]current page.north west) {{\fontsize{11.040pt}{11.702pt}\selectfont  \makebox[433.65pt][s]{The proposed Autonomous Commerce Agent Guard System (ACAGS) was designed to}}};
\node[anchor=base west,inner sep=0pt,outer sep=0pt] at ([xshift=82.104pt,yshift=-177.260pt]current page.north west) {{\fontsize{11.040pt}{11.702pt}\selectfont  \makebox[433.72pt][s]{strengthen cybersecurity in agentic commerce payment authorization systems by integrating}}};
\node[anchor=base west,inner sep=0pt,outer sep=0pt] at ([xshift=82.104pt,yshift=-189.860pt]current page.north west) {{\fontsize{11.040pt}{11.702pt}\selectfont  \makebox[433.78pt][s]{behavior-aware fraud detection, agent-specific authorization controls, and explainable AI. The}}};
\node[anchor=base west,inner sep=0pt,outer sep=0pt] at ([xshift=82.104pt,yshift=-202.580pt]current page.north west) {{\fontsize{11.040pt}{11.702pt}\selectfont  \makebox[433.60pt][s]{framework aims to (i) enhance fraud detection through contextual behavioral analysis, (ii)}}};
\node[anchor=base west,inner sep=0pt,outer sep=0pt] at ([xshift=82.104pt,yshift=-215.180pt]current page.north west) {{\fontsize{11.040pt}{11.702pt}\selectfont  \makebox[433.89pt][s]{incorporate agent identity, delegated authority, and authorization-scope constraints into}}};
\node[anchor=base west,inner sep=0pt,outer sep=0pt] at ([xshift=82.104pt,yshift=-227.900pt]current page.north west) {{\fontsize{11.040pt}{11.702pt}\selectfont  \makebox[433.90pt][s]{transaction assessment, (iii) provide transparent and interpretable fraud-risk explanations, (iv)}}};
\node[anchor=base west,inner sep=0pt,outer sep=0pt] at ([xshift=82.104pt,yshift=-240.500pt]current page.north west) {{\fontsize{11.040pt}{11.702pt}\selectfont  \makebox[433.74pt][s]{support secure payment authorization through APPROVE, REVIEW, and REJECT outcomes,}}};
\node[anchor=base west,inner sep=0pt,outer sep=0pt] at ([xshift=82.104pt,yshift=-253.100pt]current page.north west) {{\fontsize{11.040pt}{11.702pt}\selectfont  \makebox[433.61pt][s]{and (v) maintain a modular architecture adaptable to existing payment authorization}}};
\node[anchor=base west,inner sep=0pt,outer sep=0pt] at ([xshift=82.104pt,yshift=-265.820pt]current page.north west) {{\fontsize{11.040pt}{11.702pt}\selectfont  \makebox[433.61pt][s]{infrastructures. These objectives establish the design foundation of ACAGS and guide the}}};
\node[anchor=base west,inner sep=0pt,outer sep=0pt] at ([xshift=82.104pt,yshift=-278.420pt]current page.north west) {{\fontsize{11.040pt}{11.702pt}\selectfont  \makebox[204.39pt][s]{framework architecture presented in Figure 1.}}};
\node[anchor=base west,inner sep=0pt,outer sep=0pt] at ([xshift=106.940pt,yshift=-488.350pt]current page.north west) {{\fontsize{9.960pt}{10.558pt}\selectfont  \makebox[383.86pt][s]{Figure 1. Architecture of ACAGS for behavior-aware fraud detection and agent-aware payment}}};
\node[anchor=base west,inner sep=0pt,outer sep=0pt] at ([xshift=271.250pt,yshift=-499.870pt]current page.north west) {{\fontsize{9.960pt}{10.558pt}\selectfont  authorization}};
\node[anchor=base west,inner sep=0pt,outer sep=0pt] at ([xshift=82.104pt,yshift=-525.910pt]current page.north west) {{\fontsize{12.000pt}{12.720pt}\selectfont \bfseries 3.2.}};
\node[anchor=base west,inner sep=0pt,outer sep=0pt] at ([xshift=104.660pt,yshift=-525.910pt]current page.north west) {{\fontsize{12.000pt}{12.720pt}\selectfont \bfseries \makebox[130.83pt][s]{Framework Architecture}}};
\node[anchor=base west,inner sep=0pt,outer sep=0pt] at ([xshift=82.104pt,yshift=-551.350pt]current page.north west) {{\fontsize{11.040pt}{11.702pt}\selectfont  \makebox[433.66pt][s]{Figure 1 illustrates the architecture of the proposed Autonomous Commerce Agent Guard System}}};
\node[anchor=base west,inner sep=0pt,outer sep=0pt] at ([xshift=82.104pt,yshift=-563.950pt]current page.north west) {{\fontsize{11.040pt}{11.702pt}\selectfont  \makebox[433.67pt][s]{(ACAGS), a modular cybersecurity framework designed to strengthen payment authorization in}}};
\node[anchor=base west,inner sep=0pt,outer sep=0pt] at ([xshift=82.104pt,yshift=-576.670pt]current page.north west) {{\fontsize{11.040pt}{11.702pt}\selectfont  \makebox[433.98pt][s]{agentic commerce environments through behavior-aware fraud detection and agent-specific}}};
\node[anchor=base west,inner sep=0pt,outer sep=0pt] at ([xshift=82.104pt,yshift=-589.270pt]current page.north west) {{\fontsize{11.040pt}{11.702pt}\selectfont  \makebox[433.62pt][s]{authorization controls. The framework processes each payment authorization request through five}}};
\node[anchor=base west,inner sep=0pt,outer sep=0pt] at ([xshift=82.104pt,yshift=-601.990pt]current page.north west) {{\fontsize{11.040pt}{11.702pt}\selectfont  \makebox[431.31pt][s]{interconnected functional modules: the Behavioral and Agent Intelligence Module, the Behavior-}}};
\node[anchor=base west,inner sep=0pt,outer sep=0pt] at ([xshift=82.104pt,yshift=-614.590pt]current page.north west) {{\fontsize{11.040pt}{11.702pt}\selectfont  \makebox[431.43pt][s]{Aware Fraud Detection Module, the Explainability and Risk Interpretation Module, the Agent-}}};
\node[anchor=base west,inner sep=0pt,outer sep=0pt] at ([xshift=82.104pt,yshift=-627.190pt]current page.north west) {{\fontsize{11.040pt}{11.702pt}\selectfont  \makebox[433.63pt][s]{Specific Authorization Controls Module, and the Payment Authorization Decision Module.}}};
\node[anchor=base west,inner sep=0pt,outer sep=0pt] at ([xshift=82.104pt,yshift=-639.910pt]current page.north west) {{\fontsize{11.040pt}{11.702pt}\selectfont  \makebox[433.89pt][s]{Together, these modules integrate behavioral risk assessment, agent-specific authorization}}};
\node[anchor=base west,inner sep=0pt,outer sep=0pt] at ([xshift=82.104pt,yshift=-652.510pt]current page.north west) {{\fontsize{11.040pt}{11.702pt}\selectfont  \makebox[384.92pt][s]{controls, and explainable decision support within the payment authorization workflow.}}};
\node[anchor=base west,inner sep=0pt,outer sep=0pt] at ([xshift=82.104pt,yshift=-677.860pt]current page.north west) {{\fontsize{11.040pt}{11.702pt}\selectfont  \makebox[433.81pt][s]{The Behavioral and Agent Intelligence Module transforms raw payment transactions into}}};
\node[anchor=base west,inner sep=0pt,outer sep=0pt] at ([xshift=82.104pt,yshift=-690.460pt]current page.north west) {{\fontsize{11.040pt}{11.702pt}\selectfont  \makebox[433.74pt][s]{behavior-aware and agent-aware representations. It derives contextual characteristics including}}};
\node[anchor=base west,inner sep=0pt,outer sep=0pt] at ([xshift=82.104pt,yshift=-703.180pt]current page.north west) {{\fontsize{11.040pt}{11.702pt}\selectfont  \makebox[433.62pt][s]{customer spending patterns, temporal behavior, merchant interactions, historical transaction}}};
\node[anchor=base west,inner sep=0pt,outer sep=0pt] at ([xshift=82.104pt,yshift=-715.780pt]current page.north west) {{\fontsize{11.040pt}{11.702pt}\selectfont  \makebox[433.61pt][s]{activity, agent identity and trust, and delegated authorization information. These features provide}}};
\node[anchor=base west,inner sep=0pt,outer sep=0pt] at ([xshift=82.104pt,yshift=-728.500pt]current page.north west) {{\fontsize{11.040pt}{11.702pt}\selectfont  \makebox[433.95pt][s]{the contextual information required for fraud-risk assessment and agent-aware payment}}};
\node[anchor=base west,inner sep=0pt,outer sep=0pt] at ([xshift=82.104pt,yshift=-741.100pt]current page.north west) {{\fontsize{11.040pt}{11.702pt}\selectfont  authorization.}};
\node[anchor=north west,inner sep=0pt,outer sep=0pt] at ([xshift=158.000pt,yshift=-292.000pt]current page.north west) {\includegraphics[width=282.000pt,height=175.000pt]{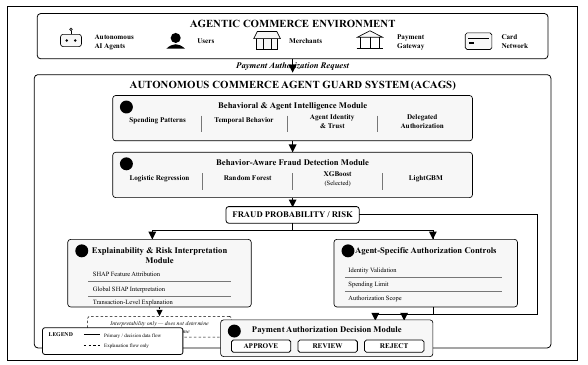}};
\end{tikzpicture}
\null
\newpage
\begin{tikzpicture}[remember picture,overlay]
\node[anchor=base west,inner sep=0pt,outer sep=0pt] at ([xshift=127.580pt,yshift=-81.360pt]current page.north west) {{\fontsize{9.960pt}{10.558pt}\selectfont  \makebox[342.77pt][s]{International Journal on Cybernetics \& Informatics (IJCI) Vol.15, No.4, August 2026}}};
\node[anchor=base west,inner sep=0pt,outer sep=0pt] at ([xshift=503.380pt,yshift=-767.856pt]current page.north west) {{\fontsize{9.960pt}{10.558pt}\selectfont  89}};
\node[anchor=base west,inner sep=0pt,outer sep=0pt] at ([xshift=82.104pt,yshift=-100.220pt]current page.north west) {{\fontsize{11.040pt}{11.702pt}\selectfont  \makebox[433.89pt][s]{The extracted features are subsequently evaluated by the Behavior-Aware Fraud Detection}}};
\node[anchor=base west,inner sep=0pt,outer sep=0pt] at ([xshift=82.104pt,yshift=-112.820pt]current page.north west) {{\fontsize{11.040pt}{11.702pt}\selectfont  \makebox[433.68pt][s]{Module, which supports multiple supervised learning classifiers. Logistic Regression, Random}}};
\node[anchor=base west,inner sep=0pt,outer sep=0pt] at ([xshift=82.104pt,yshift=-125.540pt]current page.north west) {{\fontsize{11.040pt}{11.702pt}\selectfont  \makebox[433.73pt][s]{Forest, XGBoost, and LightGBM were comparatively evaluated under identical experimental}}};
\node[anchor=base west,inner sep=0pt,outer sep=0pt] at ([xshift=82.104pt,yshift=-138.140pt]current page.north west) {{\fontsize{11.040pt}{11.702pt}\selectfont  \makebox[433.63pt][s]{conditions. Based on the baseline results presented in Section 5.1, XGBoost provided the}}};
\node[anchor=base west,inner sep=0pt,outer sep=0pt] at ([xshift=82.104pt,yshift=-150.860pt]current page.north west) {{\fontsize{11.040pt}{11.702pt}\selectfont  \makebox[433.56pt][s]{strongest overall balance across the evaluated performance metrics and was therefore selected as}}};
\node[anchor=base west,inner sep=0pt,outer sep=0pt] at ([xshift=82.104pt,yshift=-163.460pt]current page.north west) {{\fontsize{11.040pt}{11.702pt}\selectfont  \makebox[295.37pt][s]{the primary classifier for the subsequent agent-aware experiments.}}};
\node[anchor=base west,inner sep=0pt,outer sep=0pt] at ([xshift=82.104pt,yshift=-188.780pt]current page.north west) {{\fontsize{11.040pt}{11.702pt}\selectfont  \makebox[433.62pt][s]{To improve decision transparency, fraud predictions are forwarded to the Explainability and Risk}}};
\node[anchor=base west,inner sep=0pt,outer sep=0pt] at ([xshift=82.104pt,yshift=-201.380pt]current page.north west) {{\fontsize{11.040pt}{11.702pt}\selectfont  \makebox[433.60pt][s]{Interpretation Module, where SHAP [8] quantifies the contribution of individual features to}}};
\node[anchor=base west,inner sep=0pt,outer sep=0pt] at ([xshift=82.104pt,yshift=-214.100pt]current page.north west) {{\fontsize{11.040pt}{11.702pt}\selectfont  \makebox[433.91pt][s]{model predictions. The resulting explanations support interpretation of fraud-risk assessments but}}};
\node[anchor=base west,inner sep=0pt,outer sep=0pt] at ([xshift=82.104pt,yshift=-226.700pt]current page.north west) {{\fontsize{11.040pt}{11.702pt}\selectfont  \makebox[433.88pt][s]{do not determine authorization outcomes. In parallel, the Agent-Specific Authorization Controls}}};
\node[anchor=base west,inner sep=0pt,outer sep=0pt] at ([xshift=82.104pt,yshift=-239.300pt]current page.north west) {{\fontsize{11.040pt}{11.702pt}\selectfont  \makebox[433.68pt][s]{Module evaluates agent identity, delegated spending limits, and authorization scope. The}}};
\node[anchor=base west,inner sep=0pt,outer sep=0pt] at ([xshift=82.104pt,yshift=-252.020pt]current page.north west) {{\fontsize{11.040pt}{11.702pt}\selectfont  \makebox[431.31pt][s]{Payment Authorization Decision Module then combines the predicted fraud risk with these agent-}}};
\node[anchor=base west,inner sep=0pt,outer sep=0pt] at ([xshift=82.104pt,yshift=-264.620pt]current page.north west) {{\fontsize{11.040pt}{11.702pt}\selectfont  \makebox[433.92pt][s]{specific controls to produce APPROVE, REVIEW, or REJECT outcomes. This modular}}};
\node[anchor=base west,inner sep=0pt,outer sep=0pt] at ([xshift=82.104pt,yshift=-277.340pt]current page.north west) {{\fontsize{11.040pt}{11.702pt}\selectfont  \makebox[433.62pt][s]{workflow separates model explainability from authorization logic while integrating behavioral}}};
\node[anchor=base west,inner sep=0pt,outer sep=0pt] at ([xshift=82.104pt,yshift=-289.970pt]current page.north west) {{\fontsize{11.040pt}{11.702pt}\selectfont  \makebox[331.25pt][s]{fraud detection and agent-specific controls within the ACAGS framework.}}};
\node[anchor=base west,inner sep=0pt,outer sep=0pt] at ([xshift=82.104pt,yshift=-316.250pt]current page.north west) {{\fontsize{12.000pt}{12.720pt}\selectfont \bfseries 3.3.}};
\node[anchor=base west,inner sep=0pt,outer sep=0pt] at ([xshift=104.660pt,yshift=-316.250pt]current page.north west) {{\fontsize{12.000pt}{12.720pt}\selectfont \bfseries \makebox[216.01pt][s]{Behavioral and Agent Intelligence Module}}};
\node[anchor=base west,inner sep=0pt,outer sep=0pt] at ([xshift=82.104pt,yshift=-341.690pt]current page.north west) {{\fontsize{11.040pt}{11.702pt}\selectfont  \makebox[433.66pt][s]{The Behavioral and Agent Intelligence Module transforms raw payment transaction data into}}};
\node[anchor=base west,inner sep=0pt,outer sep=0pt] at ([xshift=82.104pt,yshift=-354.410pt]current page.north west) {{\fontsize{11.040pt}{11.702pt}\selectfont  \makebox[433.90pt][s]{contextual representations that support fraud-risk assessment and agent-aware authorization. It}}};
\node[anchor=base west,inner sep=0pt,outer sep=0pt] at ([xshift=82.104pt,yshift=-367.010pt]current page.north west) {{\fontsize{11.040pt}{11.702pt}\selectfont  \makebox[433.64pt][s]{captures monetary behavior, temporal behavior, merchant interactions, historical customer}}};
\node[anchor=base west,inner sep=0pt,outer sep=0pt] at ([xshift=82.104pt,yshift=-379.730pt]current page.north west) {{\fontsize{11.040pt}{11.702pt}\selectfont  \makebox[433.72pt][s]{activity, and agent-specific information including identity, trust, delegated spending limits, and}}};
\node[anchor=base west,inner sep=0pt,outer sep=0pt] at ([xshift=82.104pt,yshift=-392.330pt]current page.north west) {{\fontsize{11.040pt}{11.702pt}\selectfont  \makebox[91.32pt][s]{authorization scope.}}};
\node[anchor=base west,inner sep=0pt,outer sep=0pt] at ([xshift=82.104pt,yshift=-417.650pt]current page.north west) {{\fontsize{11.040pt}{11.702pt}\selectfont  \makebox[433.72pt][s]{Monetary and temporal features characterize spending amounts, transaction timing, and}}};
\node[anchor=base west,inner sep=0pt,outer sep=0pt] at ([xshift=82.104pt,yshift=-430.250pt]current page.north west) {{\fontsize{11.040pt}{11.702pt}\selectfont  \makebox[433.59pt][s]{frequency, while merchant and historical features capture purchasing relationships and prior}}};
\node[anchor=base west,inner sep=0pt,outer sep=0pt] at ([xshift=82.104pt,yshift=-442.970pt]current page.north west) {{\fontsize{11.040pt}{11.702pt}\selectfont  \makebox[433.72pt][s]{customer activity. Agent-specific attributes provide additional context regarding the identity,}}};
\node[anchor=base west,inner sep=0pt,outer sep=0pt] at ([xshift=82.104pt,yshift=-455.570pt]current page.north west) {{\fontsize{11.040pt}{11.702pt}\selectfont  \makebox[402.56pt][s]{trustworthiness, and delegated authority of the autonomous agent initiating the transaction.}}};
\node[anchor=base west,inner sep=0pt,outer sep=0pt] at ([xshift=82.104pt,yshift=-480.910pt]current page.north west) {{\fontsize{11.040pt}{11.702pt}\selectfont  \makebox[433.89pt][s]{The extracted features are subsequently forwarded to the Behavior-Aware Fraud Detection}}};
\node[anchor=base west,inner sep=0pt,outer sep=0pt] at ([xshift=82.104pt,yshift=-493.510pt]current page.north west) {{\fontsize{11.040pt}{11.702pt}\selectfont  \makebox[433.58pt][s]{Module, where they are used to assess transaction risk and support secure payment authorization}}};
\node[anchor=base west,inner sep=0pt,outer sep=0pt] at ([xshift=82.104pt,yshift=-506.230pt]current page.north west) {{\fontsize{11.040pt}{11.702pt}\selectfont  decisions.}};
\node[anchor=base west,inner sep=0pt,outer sep=0pt] at ([xshift=82.104pt,yshift=-532.390pt]current page.north west) {{\fontsize{12.000pt}{12.720pt}\selectfont \bfseries 3.4.}};
\node[anchor=base west,inner sep=0pt,outer sep=0pt] at ([xshift=104.660pt,yshift=-532.390pt]current page.north west) {{\fontsize{12.000pt}{12.720pt}\selectfont \bfseries \makebox[211.98pt][s]{Behavior-Aware Fraud Detection Module}}};
\node[anchor=base west,inner sep=0pt,outer sep=0pt] at ([xshift=82.104pt,yshift=-557.950pt]current page.north west) {{\fontsize{11.040pt}{11.702pt}\selectfont  \makebox[433.62pt][s]{The Behavior-Aware Fraud Detection Module evaluates the features generated by the Behavioral}}};
\node[anchor=base west,inner sep=0pt,outer sep=0pt] at ([xshift=82.104pt,yshift=-570.550pt]current page.north west) {{\fontsize{11.040pt}{11.702pt}\selectfont  \makebox[433.63pt][s]{and Agent Intelligence Module to estimate the likelihood of fraudulent activity during payment}}};
\node[anchor=base west,inner sep=0pt,outer sep=0pt] at ([xshift=82.104pt,yshift=-583.270pt]current page.north west) {{\fontsize{11.040pt}{11.702pt}\selectfont  \makebox[433.67pt][s]{authorization.ACAGS supports multiple supervised machine learning classifiers, including}}};
\node[anchor=base west,inner sep=0pt,outer sep=0pt] at ([xshift=82.104pt,yshift=-595.870pt]current page.north west) {{\fontsize{11.040pt}{11.702pt}\selectfont  \makebox[433.71pt][s]{Logistic Regression, Random Forest, XGBoost, and LightGBM, enabling comparative evaluation}}};
\node[anchor=base west,inner sep=0pt,outer sep=0pt] at ([xshift=82.104pt,yshift=-608.470pt]current page.north west) {{\fontsize{11.040pt}{11.702pt}\selectfont  \makebox[186.03pt][s]{under consistent experimental conditions.}}};
\node[anchor=base west,inner sep=0pt,outer sep=0pt] at ([xshift=82.104pt,yshift=-633.790pt]current page.north west) {{\fontsize{11.040pt}{11.702pt}\selectfont  \makebox[433.78pt][s]{The predictions generated by the Behavior-Aware Fraud Detection Module are interpreted using}}};
\node[anchor=base west,inner sep=0pt,outer sep=0pt] at ([xshift=82.104pt,yshift=-646.510pt]current page.north west) {{\fontsize{11.040pt}{11.702pt}\selectfont  \makebox[433.65pt][s]{SHAP [8] within the Explainability and Risk Interpretation Module, providing transparent}}};
\node[anchor=base west,inner sep=0pt,outer sep=0pt] at ([xshift=82.104pt,yshift=-659.110pt]current page.north west) {{\fontsize{11.040pt}{11.702pt}\selectfont  \makebox[433.75pt][s]{feature-level explanations of fraud-risk predictions. The predicted fraud probability is separately}}};
\node[anchor=base west,inner sep=0pt,outer sep=0pt] at ([xshift=82.104pt,yshift=-671.860pt]current page.north west) {{\fontsize{11.040pt}{11.702pt}\selectfont  \makebox[431.31pt][s]{passed to the Payment Authorization Decision Module, where it is evaluated alongside agent-}}};
\node[anchor=base west,inner sep=0pt,outer sep=0pt] at ([xshift=82.104pt,yshift=-684.460pt]current page.north west) {{\fontsize{11.040pt}{11.702pt}\selectfont  \makebox[433.68pt][s]{specific authorization controls to produce APPROVE, REVIEW, or REJECT outcomes. SHAP}}};
\node[anchor=base west,inner sep=0pt,outer sep=0pt] at ([xshift=82.104pt,yshift=-697.060pt]current page.north west) {{\fontsize{11.040pt}{11.702pt}\selectfont  \makebox[433.79pt][s]{supports interpretation of the fraud-risk assessment but does not independently determine the}}};
\node[anchor=base west,inner sep=0pt,outer sep=0pt] at ([xshift=82.104pt,yshift=-709.780pt]current page.north west) {{\fontsize{11.040pt}{11.702pt}\selectfont  \makebox[102.96pt][s]{authorization decision.}}};
\end{tikzpicture}
\null
\newpage
\begin{tikzpicture}[remember picture,overlay]
\node[anchor=base west,inner sep=0pt,outer sep=0pt] at ([xshift=127.580pt,yshift=-81.360pt]current page.north west) {{\fontsize{9.960pt}{10.558pt}\selectfont  \makebox[342.77pt][s]{International Journal on Cybernetics \& Informatics (IJCI) Vol.15, No.4, August 2026}}};
\node[anchor=base west,inner sep=0pt,outer sep=0pt] at ([xshift=503.380pt,yshift=-767.856pt]current page.north west) {{\fontsize{9.960pt}{10.558pt}\selectfont  90}};
\node[anchor=base west,inner sep=0pt,outer sep=0pt] at ([xshift=82.104pt,yshift=-101.180pt]current page.north west) {{\fontsize{12.000pt}{12.720pt}\selectfont \bfseries 3.5.}};
\node[anchor=base west,inner sep=0pt,outer sep=0pt] at ([xshift=104.660pt,yshift=-101.180pt]current page.north west) {{\fontsize{12.000pt}{12.720pt}\selectfont \bfseries Threat Model}};
\node[anchor=base west,inner sep=0pt,outer sep=0pt] at ([xshift=82.104pt,yshift=-126.620pt]current page.north west) {{\fontsize{11.040pt}{11.702pt}\selectfont  \makebox[433.64pt][s]{ACAGS is designed to address cybersecurity threats that arise when autonomous AI agents}}};
\node[anchor=base west,inner sep=0pt,outer sep=0pt] at ([xshift=82.104pt,yshift=-139.340pt]current page.north west) {{\fontsize{11.040pt}{11.702pt}\selectfont  \makebox[433.74pt][s]{initiate payment transactions under delegated authority. Unlike conventional payment}}};
\node[anchor=base west,inner sep=0pt,outer sep=0pt] at ([xshift=82.104pt,yshift=-151.940pt]current page.north west) {{\fontsize{11.040pt}{11.702pt}\selectfont  \makebox[433.60pt][s]{environments in which transactions are initiated directly by users, agentic commerce introduces}}};
\node[anchor=base west,inner sep=0pt,outer sep=0pt] at ([xshift=82.104pt,yshift=-164.660pt]current page.north west) {{\fontsize{11.040pt}{11.702pt}\selectfont  \makebox[433.60pt][s]{software agents capable of selecting products, interacting with merchants, and executing}}};
\node[anchor=base west,inner sep=0pt,outer sep=0pt] at ([xshift=82.104pt,yshift=-177.260pt]current page.north west) {{\fontsize{11.040pt}{11.702pt}\selectfont  \makebox[433.71pt][s]{payments on behalf of customers.This autonomy expands the attack surface by introducing risks}}};
\node[anchor=base west,inner sep=0pt,outer sep=0pt] at ([xshift=82.104pt,yshift=-189.860pt]current page.north west) {{\fontsize{11.040pt}{11.702pt}\selectfont  \makebox[433.58pt][s]{associated with agent identity, delegation, authorization scope, and compromised agent behavior}}};
\node[anchor=base west,inner sep=0pt,outer sep=0pt] at ([xshift=82.104pt,yshift=-202.580pt]current page.north west) {{\fontsize{11.040pt}{11.702pt}\selectfont  \makebox[66.00pt][s]{[3], [11]--[13].}}};
\node[anchor=base west,inner sep=0pt,outer sep=0pt] at ([xshift=82.104pt,yshift=-227.900pt]current page.north west) {{\fontsize{11.040pt}{11.702pt}\selectfont  \makebox[433.65pt][s]{The threat model considers payment authorization systems, customer financial accounts,}}};
\node[anchor=base west,inner sep=0pt,outer sep=0pt] at ([xshift=82.104pt,yshift=-240.500pt]current page.north west) {{\fontsize{11.040pt}{11.702pt}\selectfont  \makebox[433.62pt][s]{transaction integrity, delegated authorization policies, and agent credentials as protected assets.}}};
\node[anchor=base west,inner sep=0pt,outer sep=0pt] at ([xshift=82.104pt,yshift=-253.100pt]current page.north west) {{\fontsize{11.040pt}{11.702pt}\selectfont  \makebox[433.66pt][s]{Potential adversaries include external cybercriminals, compromised or malicious autonomous}}};
\node[anchor=base west,inner sep=0pt,outer sep=0pt] at ([xshift=82.104pt,yshift=-265.820pt]current page.north west) {{\fontsize{11.040pt}{11.702pt}\selectfont  \makebox[433.68pt][s]{agents, account-takeover attackers, and other actors seeking to exploit delegated payment}}};
\node[anchor=base west,inner sep=0pt,outer sep=0pt] at ([xshift=82.104pt,yshift=-278.420pt]current page.north west) {{\fontsize{11.040pt}{11.702pt}\selectfont  \makebox[433.96pt][s]{authority. Five agent-specific attack scenarios are considered: compromised agent, in which a}}};
\node[anchor=base west,inner sep=0pt,outer sep=0pt] at ([xshift=82.104pt,yshift=-291.170pt]current page.north west) {{\fontsize{11.040pt}{11.702pt}\selectfont  \makebox[433.87pt][s]{legitimate agent is manipulated or taken over; agent impersonation, in which an attacker attempts}}};
\node[anchor=base west,inner sep=0pt,outer sep=0pt] at ([xshift=82.104pt,yshift=-303.770pt]current page.north west) {{\fontsize{11.040pt}{11.702pt}\selectfont  \makebox[135.17pt][s]{to operate as a trusted agent}}};
\node[anchor=base west,inner sep=0pt,outer sep=0pt] at ([xshift=217.490pt,yshift=-303.770pt]current page.north west) {{\fontsize{11.040pt}{11.702pt}\selectfont \bfseries ;}};
\node[anchor=base west,inner sep=0pt,outer sep=0pt] at ([xshift=226.370pt,yshift=-303.770pt]current page.north west) {{\fontsize{11.040pt}{11.702pt}\selectfont  \makebox[289.55pt][s]{malicious delegation, involving misuse of delegated spending}}};
\node[anchor=base west,inner sep=0pt,outer sep=0pt] at ([xshift=82.104pt,yshift=-316.490pt]current page.north west) {{\fontsize{11.040pt}{11.702pt}\selectfont  \makebox[433.68pt][s]{authority; scope violation, in which an agent attempts activity outside its permitted authorization}}};
\node[anchor=base west,inner sep=0pt,outer sep=0pt] at ([xshift=82.104pt,yshift=-329.090pt]current page.north west) {{\fontsize{11.040pt}{11.702pt}\selectfont  \makebox[433.77pt][s]{scope; and abnormal agent behavior, representing significant deviations from expected agent}}};
\node[anchor=base west,inner sep=0pt,outer sep=0pt] at ([xshift=82.104pt,yshift=-341.690pt]current page.north west) {{\fontsize{11.040pt}{11.702pt}\selectfont  \makebox[91.32pt][s]{transaction patterns.}}};
\node[anchor=base west,inner sep=0pt,outer sep=0pt] at ([xshift=82.104pt,yshift=-367.010pt]current page.north west) {{\fontsize{11.040pt}{11.702pt}\selectfont  \makebox[433.85pt][s]{ACAGS assumes that conventional authentication and access-control mechanisms remain part of}}};
\node[anchor=base west,inner sep=0pt,outer sep=0pt] at ([xshift=82.104pt,yshift=-379.730pt]current page.north west) {{\fontsize{11.040pt}{11.702pt}\selectfont  \makebox[433.58pt][s]{the underlying payment infrastructure. The framework therefore does not replace authentication}}};
\node[anchor=base west,inner sep=0pt,outer sep=0pt] at ([xshift=82.104pt,yshift=-392.330pt]current page.north west) {{\fontsize{11.040pt}{11.702pt}\selectfont  \makebox[433.62pt][s]{or payment-network security controls. Instead, it introduces an additional cybersecurity layer that}}};
\node[anchor=base west,inner sep=0pt,outer sep=0pt] at ([xshift=82.104pt,yshift=-404.930pt]current page.north west) {{\fontsize{11.040pt}{11.702pt}\selectfont  \makebox[433.63pt][s]{evaluates transaction-level behavioral risk together with agent identity, trust, delegated spending}}};
\node[anchor=base west,inner sep=0pt,outer sep=0pt] at ([xshift=82.104pt,yshift=-417.650pt]current page.north west) {{\fontsize{11.040pt}{11.702pt}\selectfont  \makebox[433.78pt][s]{limits, and authorization-scope constraints. This design enables transactions exhibiting elevated}}};
\node[anchor=base west,inner sep=0pt,outer sep=0pt] at ([xshift=82.104pt,yshift=-430.250pt]current page.north west) {{\fontsize{11.040pt}{11.702pt}\selectfont  \makebox[431.31pt][s]{fraud probability or agent-policy violations to be escalated for review or rejection while lower-}}};
\node[anchor=base west,inner sep=0pt,outer sep=0pt] at ([xshift=82.104pt,yshift=-442.970pt]current page.north west) {{\fontsize{11.040pt}{11.702pt}\selectfont  \makebox[276.53pt][s]{risk activity may proceed through the authorization workflow.}}};
\node[anchor=base west,inner sep=0pt,outer sep=0pt] at ([xshift=82.104pt,yshift=-468.170pt]current page.north west) {{\fontsize{11.040pt}{11.702pt}\selectfont  \makebox[431.43pt][s]{The threat model consequently combines behavioral anomaly detection with explicit agent-}}};
\node[anchor=base west,inner sep=0pt,outer sep=0pt] at ([xshift=82.104pt,yshift=-480.910pt]current page.north west) {{\fontsize{11.040pt}{11.702pt}\selectfont  \makebox[433.67pt][s]{authorization controls. This distinction is important because some attacks may occur through}}};
\node[anchor=base west,inner sep=0pt,outer sep=0pt] at ([xshift=82.104pt,yshift=-493.510pt]current page.north west) {{\fontsize{11.040pt}{11.702pt}\selectfont  \makebox[433.63pt][s]{apparently valid agent credentials and therefore may not be identifiable through identity}}};
\node[anchor=base west,inner sep=0pt,outer sep=0pt] at ([xshift=82.104pt,yshift=-506.230pt]current page.north west) {{\fontsize{11.040pt}{11.702pt}\selectfont  \makebox[433.64pt][s]{verification alone. The subsequent experimental evaluation operationalizes these five attack}}};
\node[anchor=base west,inner sep=0pt,outer sep=0pt] at ([xshift=82.104pt,yshift=-518.830pt]current page.north west) {{\fontsize{11.040pt}{11.702pt}\selectfont  \makebox[433.82pt][s]{scenarios in a controlled simulated agentic-commerce environment to assess their detectability}}};
\node[anchor=base west,inner sep=0pt,outer sep=0pt] at ([xshift=82.104pt,yshift=-531.430pt]current page.north west) {{\fontsize{11.040pt}{11.702pt}\selectfont  \makebox[168.39pt][s]{and resulting authorization outcomes.}}};
\node[anchor=base west,inner sep=0pt,outer sep=0pt] at ([xshift=82.104pt,yshift=-559.630pt]current page.north west) {{\fontsize{14.040pt}{14.882pt}\selectfont \bfseries 4.}};
\node[anchor=base west,inner sep=0pt,outer sep=0pt] at ([xshift=100.100pt,yshift=-559.630pt]current page.north west) {{\fontsize{14.040pt}{14.882pt}\selectfont \bfseries R}};
\node[anchor=base west,inner sep=0pt,outer sep=0pt] at ([xshift=110.300pt,yshift=-559.630pt]current page.north west) {{\fontsize{11.040pt}{11.702pt}\selectfont \bfseries ESEARCH}};
\node[anchor=base west,inner sep=0pt,outer sep=0pt] at ([xshift=166.220pt,yshift=-559.630pt]current page.north west) {{\fontsize{14.040pt}{14.882pt}\selectfont \bfseries M}};
\node[anchor=base west,inner sep=0pt,outer sep=0pt] at ([xshift=179.420pt,yshift=-559.630pt]current page.north west) {{\fontsize{11.040pt}{11.702pt}\selectfont \bfseries ETHODOLOGY}};
\node[anchor=base west,inner sep=0pt,outer sep=0pt] at ([xshift=82.104pt,yshift=-585.550pt]current page.north west) {{\fontsize{11.040pt}{11.702pt}\selectfont  \makebox[433.90pt][s]{The research methodology was designed to evaluate the behavior-aware fraud detection and}}};
\node[anchor=base west,inner sep=0pt,outer sep=0pt] at ([xshift=82.104pt,yshift=-598.150pt]current page.north west) {{\fontsize{11.040pt}{11.702pt}\selectfont  \makebox[433.64pt][s]{agent-aware authorization components of ACAGS. The evaluation combines behavioral feature}}};
\node[anchor=base west,inner sep=0pt,outer sep=0pt] at ([xshift=82.104pt,yshift=-610.870pt]current page.north west) {{\fontsize{11.040pt}{11.702pt}\selectfont  \makebox[433.90pt][s]{engineering, supervised machine learning, a controlled agentic-commerce simulation, explainable}}};
\node[anchor=base west,inner sep=0pt,outer sep=0pt] at ([xshift=82.104pt,yshift=-623.470pt]current page.north west) {{\fontsize{11.040pt}{11.702pt}\selectfont  \makebox[433.91pt][s]{AI, and agent-specific authorization controls. Four classifiers---Logistic Regression, Random}}};
\node[anchor=base west,inner sep=0pt,outer sep=0pt] at ([xshift=82.104pt,yshift=-636.070pt]current page.north west) {{\fontsize{11.040pt}{11.702pt}\selectfont  \makebox[433.82pt][s]{Forest, XGBoost, and LightGBM---were comparatively evaluated, followed by extended}}};
\node[anchor=base west,inner sep=0pt,outer sep=0pt] at ([xshift=82.104pt,yshift=-648.790pt]current page.north west) {{\fontsize{11.040pt}{11.702pt}\selectfont  \makebox[433.77pt][s]{experiments assessing agent-specific attack scenarios, feature contributions through ablation}}};
\node[anchor=base west,inner sep=0pt,outer sep=0pt] at ([xshift=82.104pt,yshift=-661.420pt]current page.north west) {{\fontsize{11.040pt}{11.702pt}\selectfont  \makebox[433.68pt][s]{analysis, statistical significance, and authorization outcomes. SHAP analysis was additionally}}};
\node[anchor=base west,inner sep=0pt,outer sep=0pt] at ([xshift=82.104pt,yshift=-674.140pt]current page.north west) {{\fontsize{11.040pt}{11.702pt}\selectfont  \makebox[288.77pt][s]{applied to interpret the factors influencing fraud-risk predictions.}}};
\node[anchor=base west,inner sep=0pt,outer sep=0pt] at ([xshift=82.104pt,yshift=-700.300pt]current page.north west) {{\fontsize{12.000pt}{12.720pt}\selectfont \bfseries 4.1.}};
\node[anchor=base west,inner sep=0pt,outer sep=0pt] at ([xshift=104.660pt,yshift=-700.300pt]current page.north west) {{\fontsize{12.000pt}{12.720pt}\selectfont \bfseries \makebox[110.67pt][s]{Experimental Design}}};
\node[anchor=base west,inner sep=0pt,outer sep=0pt] at ([xshift=82.104pt,yshift=-725.860pt]current page.north west) {{\fontsize{11.040pt}{11.702pt}\selectfont  \makebox[433.81pt][s]{Four supervised machine learning classifiers---Logistic Regression, Random Forest, XGBoost,}}};
\node[anchor=base west,inner sep=0pt,outer sep=0pt] at ([xshift=82.104pt,yshift=-738.460pt]current page.north west) {{\fontsize{11.040pt}{11.702pt}\selectfont  \makebox[433.71pt][s]{and LightGBM---were initially evaluated under identical experimental conditions to establish}}};
\end{tikzpicture}
\null
\newpage
\begin{tikzpicture}[remember picture,overlay]
\node[anchor=base west,inner sep=0pt,outer sep=0pt] at ([xshift=127.580pt,yshift=-81.360pt]current page.north west) {{\fontsize{9.960pt}{10.558pt}\selectfont  \makebox[342.77pt][s]{International Journal on Cybernetics \& Informatics (IJCI) Vol.15, No.4, August 2026}}};
\node[anchor=base west,inner sep=0pt,outer sep=0pt] at ([xshift=503.380pt,yshift=-767.856pt]current page.north west) {{\fontsize{9.960pt}{10.558pt}\selectfont  91}};
\node[anchor=base west,inner sep=0pt,outer sep=0pt] at ([xshift=82.104pt,yshift=-100.220pt]current page.north west) {{\fontsize{11.040pt}{11.702pt}\selectfont  \makebox[433.67pt][s]{baseline fraud-detection performance. XGBoost was subsequently used for the extended ACAGS}}};
\node[anchor=base west,inner sep=0pt,outer sep=0pt] at ([xshift=82.104pt,yshift=-112.820pt]current page.north west) {{\fontsize{11.040pt}{11.702pt}\selectfont  \makebox[433.74pt][s]{experiments involving behavioral and agent-specific features, simulated agent attacks, ablation}}};
\node[anchor=base west,inner sep=0pt,outer sep=0pt] at ([xshift=82.104pt,yshift=-125.540pt]current page.north west) {{\fontsize{11.040pt}{11.702pt}\selectfont  \makebox[375.44pt][s]{analysis, statistical validation, explainability, and authorization-outcome assessment.}}};
\node[anchor=base west,inner sep=0pt,outer sep=0pt] at ([xshift=82.104pt,yshift=-151.820pt]current page.north west) {{\fontsize{12.000pt}{12.720pt}\selectfont \bfseries 4.2.}};
\node[anchor=base west,inner sep=0pt,outer sep=0pt] at ([xshift=104.660pt,yshift=-151.820pt]current page.north west) {{\fontsize{12.000pt}{12.720pt}\selectfont \bfseries \makebox[224.45pt][s]{Dataset and Agentic-Commerce Simulation}}};
\node[anchor=base west,inner sep=0pt,outer sep=0pt] at ([xshift=82.104pt,yshift=-177.260pt]current page.north west) {{\fontsize{11.040pt}{11.702pt}\selectfont  \makebox[433.59pt][s]{The experimental evaluation utilized the Credit Card Transactions Fraud Detection Dataset [17],}}};
\node[anchor=base west,inner sep=0pt,outer sep=0pt] at ([xshift=82.104pt,yshift=-189.860pt]current page.north west) {{\fontsize{11.040pt}{11.702pt}\selectfont  \makebox[433.60pt][s]{a publicly available synthetic financial transaction dataset hosted on Kaggle and generated using}}};
\node[anchor=base west,inner sep=0pt,outer sep=0pt] at ([xshift=82.104pt,yshift=-202.580pt]current page.north west) {{\fontsize{11.040pt}{11.702pt}\selectfont  \makebox[433.98pt][s]{the Sparkov Data Generation tool. The analysis used the fraudTrain.csv dataset,}}};
\node[anchor=base west,inner sep=0pt,outer sep=0pt] at ([xshift=82.104pt,yshift=-215.180pt]current page.north west) {{\fontsize{11.040pt}{11.702pt}\selectfont  \makebox[145.98pt][s]{containing1,296,675 transactions}}};
\node[anchor=base west,inner sep=0pt,outer sep=0pt] at ([xshift=228.170pt,yshift=-215.180pt]current page.north west) {{\fontsize{11.040pt}{11.702pt}\selectfont \bfseries ,}};
\node[anchor=base west,inner sep=0pt,outer sep=0pt] at ([xshift=233.810pt,yshift=-215.180pt]current page.north west) {{\fontsize{11.040pt}{11.702pt}\selectfont  \makebox[282.28pt][s]{of which1,289,169 are legitimate and 7,506 are fraudulent. The}}};
\node[anchor=base west,inner sep=0pt,outer sep=0pt] at ([xshift=82.104pt,yshift=-227.900pt]current page.north west) {{\fontsize{11.040pt}{11.702pt}\selectfont  \makebox[433.57pt][s]{dataset includes transaction, customer, merchant, geographic, and temporal attributes relevant to}}};
\node[anchor=base west,inner sep=0pt,outer sep=0pt] at ([xshift=82.104pt,yshift=-240.500pt]current page.north west) {{\fontsize{11.040pt}{11.702pt}\selectfont  \makebox[433.71pt][s]{payment authorization, including transaction amount, merchant category, transaction time,}}};
\node[anchor=base west,inner sep=0pt,outer sep=0pt] at ([xshift=82.104pt,yshift=-253.100pt]current page.north west) {{\fontsize{11.040pt}{11.702pt}\selectfont  \makebox[190.11pt][s]{customer location, and merchant location.}}};
\node[anchor=base west,inner sep=0pt,outer sep=0pt] at ([xshift=82.104pt,yshift=-278.420pt]current page.north west) {{\fontsize{11.040pt}{11.702pt}\selectfont  \makebox[433.62pt][s]{Given the limited availability of publicly accessible datasets specifically representing}}};
\node[anchor=base west,inner sep=0pt,outer sep=0pt] at ([xshift=82.104pt,yshift=-291.170pt]current page.north west) {{\fontsize{11.040pt}{11.702pt}\selectfont  \makebox[433.64pt][s]{autonomous agent-initiated payment transactions, the transaction dataset was extended through a}}};
\node[anchor=base west,inner sep=0pt,outer sep=0pt] at ([xshift=82.104pt,yshift=-303.770pt]current page.north west) {{\fontsize{11.040pt}{11.702pt}\selectfont  \makebox[433.70pt][s]{controlled agentic-commerce simulation. Simulated autonomous AI agents were associated with}}};
\node[anchor=base west,inner sep=0pt,outer sep=0pt] at ([xshift=82.104pt,yshift=-316.490pt]current page.north west) {{\fontsize{11.040pt}{11.702pt}\selectfont  \makebox[433.90pt][s]{customer transactions and assigned agent-specific attributes representing agent identity, trust,}}};
\node[anchor=base west,inner sep=0pt,outer sep=0pt] at ([xshift=82.104pt,yshift=-329.090pt]current page.north west) {{\fontsize{11.040pt}{11.702pt}\selectfont  \makebox[232.67pt][s]{delegated spending limits, and authorization scope}}};
\node[anchor=base west,inner sep=0pt,outer sep=0pt] at ([xshift=314.930pt,yshift=-329.090pt]current page.north west) {{\fontsize{11.040pt}{11.702pt}\selectfont \bfseries .}};
\node[anchor=base west,inner sep=0pt,outer sep=0pt] at ([xshift=322.390pt,yshift=-329.090pt]current page.north west) {{\fontsize{11.040pt}{11.702pt}\selectfont  \makebox[193.60pt][s]{These attributes enabled the conventional}}};
\node[anchor=base west,inner sep=0pt,outer sep=0pt] at ([xshift=82.104pt,yshift=-341.690pt]current page.north west) {{\fontsize{11.040pt}{11.702pt}\selectfont  \makebox[433.64pt][s]{transaction environment to be augmented with delegated authorization characteristics required to}}};
\node[anchor=base west,inner sep=0pt,outer sep=0pt] at ([xshift=82.104pt,yshift=-354.410pt]current page.north west) {{\fontsize{11.040pt}{11.702pt}\selectfont  \makebox[176.91pt][s]{evaluate agent-aware payment security.}}};
\node[anchor=base west,inner sep=0pt,outer sep=0pt] at ([xshift=82.104pt,yshift=-379.730pt]current page.north west) {{\fontsize{11.040pt}{11.702pt}\selectfont  \makebox[433.78pt][s]{The simulation introduced five agent-specific attack scenarios consistent with the threat model}}};
\node[anchor=base west,inner sep=0pt,outer sep=0pt] at ([xshift=82.104pt,yshift=-392.330pt]current page.north west) {{\fontsize{11.040pt}{11.702pt}\selectfont  \makebox[433.79pt][s]{defined in Section 3.5: compromised agent, agent impersonation, malicious delegation, scope}}};
\node[anchor=base west,inner sep=0pt,outer sep=0pt] at ([xshift=82.104pt,yshift=-404.930pt]current page.north west) {{\fontsize{11.040pt}{11.702pt}\selectfont  \makebox[433.83pt][s]{violation, and abnormal agent behavior. A compromised-agent scenario represents misuse of an}}};
\node[anchor=base west,inner sep=0pt,outer sep=0pt] at ([xshift=82.104pt,yshift=-417.650pt]current page.north west) {{\fontsize{11.040pt}{11.702pt}\selectfont  \makebox[433.58pt][s]{otherwise legitimate agent; agent impersonation represents attempts to operate as a trusted agent;}}};
\node[anchor=base west,inner sep=0pt,outer sep=0pt] at ([xshift=82.104pt,yshift=-430.250pt]current page.north west) {{\fontsize{11.040pt}{11.702pt}\selectfont  \makebox[433.62pt][s]{malicious delegation represents abuse of delegated spending authority; scope violation represents}}};
\node[anchor=base west,inner sep=0pt,outer sep=0pt] at ([xshift=82.104pt,yshift=-442.970pt]current page.north west) {{\fontsize{11.040pt}{11.702pt}\selectfont  \makebox[433.68pt][s]{activity outside the agent's permitted authorization scope; and abnormal agent behavior represents}}};
\node[anchor=base west,inner sep=0pt,outer sep=0pt] at ([xshift=82.104pt,yshift=-455.570pt]current page.north west) {{\fontsize{11.040pt}{11.702pt}\selectfont  \makebox[433.63pt][s]{substantial deviations from expected transaction patterns. These scenarios were designed to}}};
\node[anchor=base west,inner sep=0pt,outer sep=0pt] at ([xshift=82.104pt,yshift=-468.170pt]current page.north west) {{\fontsize{11.040pt}{11.702pt}\selectfont  \makebox[433.58pt][s]{evaluate the ability of behavioral risk signals and explicit authorization controls to identify}}};
\node[anchor=base west,inner sep=0pt,outer sep=0pt] at ([xshift=82.104pt,yshift=-480.910pt]current page.north west) {{\fontsize{11.040pt}{11.702pt}\selectfont  \makebox[195.51pt][s]{malicious or policy-violating agent activity.}}};
\node[anchor=base west,inner sep=0pt,outer sep=0pt] at ([xshift=82.104pt,yshift=-506.230pt]current page.north west) {{\fontsize{11.040pt}{11.702pt}\selectfont  \makebox[431.43pt][s]{The resulting simulated environment contained 1,422 autonomous AI agents and 12,317 agent-}}};
\node[anchor=base west,inner sep=0pt,outer sep=0pt] at ([xshift=82.104pt,yshift=-518.830pt]current page.north west) {{\fontsize{11.040pt}{11.702pt}\selectfont  \makebox[433.90pt][s]{attack transactions distributed across the five attack scenarios. Agent-specific variables were}}};
\node[anchor=base west,inner sep=0pt,outer sep=0pt] at ([xshift=82.104pt,yshift=-531.430pt]current page.north west) {{\fontsize{11.040pt}{11.702pt}\selectfont  \makebox[433.98pt][s]{combined with transaction and behavioral features to support the extended fraud-risk and}}};
\node[anchor=base west,inner sep=0pt,outer sep=0pt] at ([xshift=82.104pt,yshift=-544.150pt]current page.north west) {{\fontsize{11.040pt}{11.702pt}\selectfont  \makebox[433.85pt][s]{authorization experiments. The simulation represents a controlled experimental extension of the}}};
\node[anchor=base west,inner sep=0pt,outer sep=0pt] at ([xshift=82.104pt,yshift=-556.750pt]current page.north west) {{\fontsize{11.040pt}{11.702pt}\selectfont  \makebox[433.98pt][s]{underlying transaction dataset rather than observations from a production agentic-commerce}}};
\node[anchor=base west,inner sep=0pt,outer sep=0pt] at ([xshift=82.104pt,yshift=-569.470pt]current page.north west) {{\fontsize{11.040pt}{11.702pt}\selectfont  payment system}};
\node[anchor=base west,inner sep=0pt,outer sep=0pt] at ([xshift=153.380pt,yshift=-569.470pt]current page.north west) {{\fontsize{11.040pt}{11.702pt}\selectfont \bfseries .}};
\node[anchor=base west,inner sep=0pt,outer sep=0pt] at ([xshift=82.104pt,yshift=-594.670pt]current page.north west) {{\fontsize{11.040pt}{11.702pt}\selectfont  \makebox[433.84pt][s]{Data preprocessing included the removal of non-modeling identifiers, categorical feature}}};
\node[anchor=base west,inner sep=0pt,outer sep=0pt] at ([xshift=82.104pt,yshift=-607.390pt]current page.north west) {{\fontsize{11.040pt}{11.702pt}\selectfont  \makebox[433.59pt][s]{encoding, replacement of infinite values, and median imputation of missing feature values. For}}};
\node[anchor=base west,inner sep=0pt,outer sep=0pt] at ([xshift=82.104pt,yshift=-619.990pt]current page.north west) {{\fontsize{11.040pt}{11.702pt}\selectfont  \makebox[433.99pt][s]{the baseline classifier comparison, the processed transaction dataset was partitioned into 80\%}}};
\node[anchor=base west,inner sep=0pt,outer sep=0pt] at ([xshift=82.104pt,yshift=-632.710pt]current page.north west) {{\fontsize{11.040pt}{11.702pt}\selectfont  \makebox[433.89pt][s]{training and 20\% testing subsets using stratified sampling to preserve the fraud-class distribution.}}};
\node[anchor=base west,inner sep=0pt,outer sep=0pt] at ([xshift=82.104pt,yshift=-645.310pt]current page.north west) {{\fontsize{11.040pt}{11.702pt}\selectfont  \makebox[433.58pt][s]{A fixed random state and identical data partition were used across the evaluated classifiers to}}};
\node[anchor=base west,inner sep=0pt,outer sep=0pt] at ([xshift=82.104pt,yshift=-658.030pt]current page.north west) {{\fontsize{11.040pt}{11.702pt}\selectfont  \makebox[197.31pt][s]{ensure a consistent comparative assessment.}}};
\node[anchor=base west,inner sep=0pt,outer sep=0pt] at ([xshift=82.104pt,yshift=-684.220pt]current page.north west) {{\fontsize{12.000pt}{12.720pt}\selectfont \bfseries 4.3.}};
\node[anchor=base west,inner sep=0pt,outer sep=0pt] at ([xshift=104.660pt,yshift=-684.220pt]current page.north west) {{\fontsize{12.000pt}{12.720pt}\selectfont \bfseries \makebox[167.43pt][s]{Behavioral Feature Engineering}}};
\node[anchor=base west,inner sep=0pt,outer sep=0pt] at ([xshift=82.104pt,yshift=-709.780pt]current page.north west) {{\fontsize{11.040pt}{11.702pt}\selectfont  \makebox[433.64pt][s]{Behavior-aware features were engineered to capture deviations between current transactions and}}};
\node[anchor=base west,inner sep=0pt,outer sep=0pt] at ([xshift=82.104pt,yshift=-722.380pt]current page.north west) {{\fontsize{11.040pt}{11.702pt}\selectfont  \makebox[433.62pt][s]{historical customer activity. These included transaction amount relative to prior spending,}}};
\node[anchor=base west,inner sep=0pt,outer sep=0pt] at ([xshift=82.104pt,yshift=-735.100pt]current page.north west) {{\fontsize{11.040pt}{11.702pt}\selectfont  \makebox[433.65pt][s]{spending deviation, transaction hour, nighttime activity, time since the previous transaction, prior}}};
\node[anchor=base west,inner sep=0pt,outer sep=0pt] at ([xshift=82.104pt,yshift=-747.696pt]current page.north west) {{\fontsize{11.040pt}{11.702pt}\selectfont  \makebox[433.89pt][s]{transaction activity, and customer--merchant and customer--category interactions. Historical}}};
\end{tikzpicture}
\null
\newpage
\begin{tikzpicture}[remember picture,overlay]
\node[anchor=base west,inner sep=0pt,outer sep=0pt] at ([xshift=127.580pt,yshift=-81.360pt]current page.north west) {{\fontsize{9.960pt}{10.558pt}\selectfont  \makebox[342.77pt][s]{International Journal on Cybernetics \& Informatics (IJCI) Vol.15, No.4, August 2026}}};
\node[anchor=base west,inner sep=0pt,outer sep=0pt] at ([xshift=503.380pt,yshift=-767.856pt]current page.north west) {{\fontsize{9.960pt}{10.558pt}\selectfont  92}};
\node[anchor=base west,inner sep=0pt,outer sep=0pt] at ([xshift=82.104pt,yshift=-100.220pt]current page.north west) {{\fontsize{11.040pt}{11.702pt}\selectfont  \makebox[434.00pt][s]{features were calculated using only transactions occurring before the transaction being evaluated,}}};
\node[anchor=base west,inner sep=0pt,outer sep=0pt] at ([xshift=82.104pt,yshift=-112.820pt]current page.north west) {{\fontsize{11.040pt}{11.702pt}\selectfont  \makebox[184.47pt][s]{thereby reducing the risk of data leakage.}}};
\node[anchor=base west,inner sep=0pt,outer sep=0pt] at ([xshift=82.104pt,yshift=-138.140pt]current page.north west) {{\fontsize{11.040pt}{11.702pt}\selectfont  \makebox[433.70pt][s]{As shown in Figure 2, fraudulent transactions generally exhibited higher amounts and greater}}};
\node[anchor=base west,inner sep=0pt,outer sep=0pt] at ([xshift=82.104pt,yshift=-150.860pt]current page.north west) {{\fontsize{11.040pt}{11.702pt}\selectfont  \makebox[429.20pt][s]{variability than legitimate transactions, supporting the inclusion of monetary behavioral features.}}};
\node[anchor=base west,inner sep=0pt,outer sep=0pt] at ([xshift=82.104pt,yshift=-176.060pt]current page.north west) {{\fontsize{11.040pt}{11.702pt}\selectfont  \makebox[433.91pt][s]{Temporal patterns also provided useful discriminatory information. Figure 3 shows variation in}}};
\node[anchor=base west,inner sep=0pt,outer sep=0pt] at ([xshift=82.104pt,yshift=-188.780pt]current page.north west) {{\fontsize{11.040pt}{11.702pt}\selectfont  \makebox[433.86pt][s]{fraud rates throughout the day, including elevated rates during late-night periods, supporting the}}};
\node[anchor=base west,inner sep=0pt,outer sep=0pt] at ([xshift=82.104pt,yshift=-201.380pt]current page.north west) {{\fontsize{11.040pt}{11.702pt}\selectfont  \makebox[329.69pt][s]{inclusion of transaction hour and nighttime activity as behavioral features.}}};
\node[anchor=base west,inner sep=0pt,outer sep=0pt] at ([xshift=183.500pt,yshift=-367.970pt]current page.north west) {{\fontsize{9.960pt}{10.558pt}\selectfont  Figure 2}};
\node[anchor=base west,inner sep=0pt,outer sep=0pt] at ([xshift=217.250pt,yshift=-367.970pt]current page.north west) {{\fontsize{9.960pt}{10.558pt}\selectfont \bfseries .}};
\node[anchor=base west,inner sep=0pt,outer sep=0pt] at ([xshift=222.290pt,yshift=-367.970pt]current page.north west) {{\fontsize{9.960pt}{10.558pt}\selectfont  \makebox[192.11pt][s]{Transaction amount distribution by fraud status}}};
\node[anchor=base west,inner sep=0pt,outer sep=0pt] at ([xshift=185.300pt,yshift=-523.510pt]current page.north west) {{\fontsize{9.960pt}{10.558pt}\selectfont  Figure 3}};
\node[anchor=base west,inner sep=0pt,outer sep=0pt] at ([xshift=219.050pt,yshift=-523.510pt]current page.north west) {{\fontsize{9.960pt}{10.558pt}\selectfont \bfseries .}};
\node[anchor=base west,inner sep=0pt,outer sep=0pt] at ([xshift=224.090pt,yshift=-523.510pt]current page.north west) {{\fontsize{9.960pt}{10.558pt}\selectfont  \makebox[188.39pt][s]{Hourly fraud rate across the evaluation dataset}}};
\node[anchor=base west,inner sep=0pt,outer sep=0pt] at ([xshift=82.104pt,yshift=-548.590pt]current page.north west) {{\fontsize{11.040pt}{11.702pt}\selectfont  \makebox[433.99pt][s]{As shown in Figure 3, fraud rates are not uniformly distributed throughout the day.Elevated fraud}}};
\node[anchor=base west,inner sep=0pt,outer sep=0pt] at ([xshift=82.104pt,yshift=-561.190pt]current page.north west) {{\fontsize{11.040pt}{11.702pt}\selectfont  \makebox[433.73pt][s]{rates are observed during late-night hours, suggesting that transaction timing may provide useful}}};
\node[anchor=base west,inner sep=0pt,outer sep=0pt] at ([xshift=82.104pt,yshift=-573.910pt]current page.north west) {{\fontsize{11.040pt}{11.702pt}\selectfont  \makebox[431.43pt][s]{discriminatory information for fraud detection. These findings further justify incorporating time-}}};
\node[anchor=base west,inner sep=0pt,outer sep=0pt] at ([xshift=82.104pt,yshift=-586.510pt]current page.north west) {{\fontsize{11.040pt}{11.702pt}\selectfont  \makebox[312.53pt][s]{dependent behavioral variables into the proposed ACAGS framework.}}};
\node[anchor=base west,inner sep=0pt,outer sep=0pt] at ([xshift=82.104pt,yshift=-612.790pt]current page.north west) {{\fontsize{12.000pt}{12.720pt}\selectfont \bfseries 4.4.}};
\node[anchor=base west,inner sep=0pt,outer sep=0pt] at ([xshift=104.660pt,yshift=-612.790pt]current page.north west) {{\fontsize{12.000pt}{12.720pt}\selectfont \bfseries \makebox[253.87pt][s]{Model Development and Experimental Validation}}};
\node[anchor=base west,inner sep=0pt,outer sep=0pt] at ([xshift=82.104pt,yshift=-638.230pt]current page.north west) {{\fontsize{11.040pt}{11.702pt}\selectfont  \makebox[433.78pt][s]{Four supervised learning classifiers---Logistic Regression, Random Forest [5], XGBoost [6], and}}};
\node[anchor=base west,inner sep=0pt,outer sep=0pt] at ([xshift=82.104pt,yshift=-650.950pt]current page.north west) {{\fontsize{11.040pt}{11.702pt}\selectfont  \makebox[433.94pt][s]{LightGBM [7]---were evaluated using the same training and testing partitions.These models were}}};
\node[anchor=base west,inner sep=0pt,outer sep=0pt] at ([xshift=82.104pt,yshift=-663.580pt]current page.north west) {{\fontsize{11.040pt}{11.702pt}\selectfont  \makebox[433.83pt][s]{selected to compare linear and ensemble-based approaches under consistent experimental}}};
\node[anchor=base west,inner sep=0pt,outer sep=0pt] at ([xshift=82.104pt,yshift=-676.180pt]current page.north west) {{\fontsize{11.040pt}{11.702pt}\selectfont  \makebox[433.56pt][s]{conditions. Based on the baseline evaluation, XGBoost was selected as the primary classifier for}}};
\node[anchor=base west,inner sep=0pt,outer sep=0pt] at ([xshift=82.104pt,yshift=-688.900pt]current page.north west) {{\fontsize{11.040pt}{11.702pt}\selectfont  \makebox[222.99pt][s]{the subsequent agent-aware ACAGS experiments.}}};
\node[anchor=base west,inner sep=0pt,outer sep=0pt] at ([xshift=82.104pt,yshift=-714.220pt]current page.north west) {{\fontsize{11.040pt}{11.702pt}\selectfont  \makebox[433.60pt][s]{To quantify the contribution of the proposed feature layers, an ablation study compared three}}};
\node[anchor=base west,inner sep=0pt,outer sep=0pt] at ([xshift=82.104pt,yshift=-726.820pt]current page.north west) {{\fontsize{11.040pt}{11.702pt}\selectfont  \makebox[433.90pt][s]{configurations: base transaction features, base plus behavioral features, and base plus behavioral}}};
\node[anchor=base west,inner sep=0pt,outer sep=0pt] at ([xshift=82.104pt,yshift=-739.420pt]current page.north west) {{\fontsize{11.040pt}{11.702pt}\selectfont  \makebox[433.84pt][s]{and agent-specific features. Robustness was assessed across 10 independent random seeds using}}};
\node[anchor=north west,inner sep=0pt,outer sep=0pt] at ([xshift=173.000pt,yshift=-207.000pt]current page.north west) {\includegraphics[width=252.000pt,height=144.000pt]{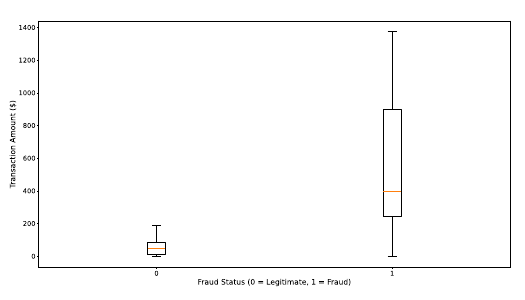}};
\node[anchor=north west,inner sep=0pt,outer sep=0pt] at ([xshift=193.000pt,yshift=-378.000pt]current page.north west) {\includegraphics[width=210.000pt,height=126.000pt]{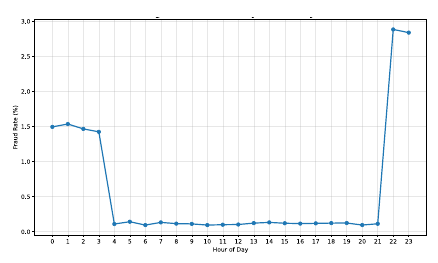}};
\end{tikzpicture}
\null
\newpage
\begin{tikzpicture}[remember picture,overlay]
\node[anchor=base west,inner sep=0pt,outer sep=0pt] at ([xshift=127.580pt,yshift=-81.360pt]current page.north west) {{\fontsize{9.960pt}{10.558pt}\selectfont  \makebox[342.77pt][s]{International Journal on Cybernetics \& Informatics (IJCI) Vol.15, No.4, August 2026}}};
\node[anchor=base west,inner sep=0pt,outer sep=0pt] at ([xshift=503.380pt,yshift=-767.856pt]current page.north west) {{\fontsize{9.960pt}{10.558pt}\selectfont  93}};
\node[anchor=base west,inner sep=0pt,outer sep=0pt] at ([xshift=82.104pt,yshift=-100.220pt]current page.north west) {{\fontsize{11.040pt}{11.702pt}\selectfont  \makebox[433.84pt][s]{identical seeds for each feature configuration, enabling paired comparisons. Statistical}}};
\node[anchor=base west,inner sep=0pt,outer sep=0pt] at ([xshift=82.104pt,yshift=-112.820pt]current page.north west) {{\fontsize{11.040pt}{11.702pt}\selectfont  \makebox[433.87pt][s]{significance was evaluated using paired t-tests and Wilcoxon signed-rank tests at a significance}}};
\node[anchor=base west,inner sep=0pt,outer sep=0pt] at ([xshift=82.104pt,yshift=-125.540pt]current page.north west) {{\fontsize{11.040pt}{11.702pt}\selectfont  level of}};
\node[anchor=base west,inner sep=0pt,outer sep=0pt] at ([xshift=118.220pt,yshift=-125.540pt]current page.north west) {{\fontsize{11.040pt}{11.702pt}\selectfont \bfseries \makebox[39.96pt][s]{$\alpha$ = 0.05.}}};
\node[anchor=base west,inner sep=0pt,outer sep=0pt] at ([xshift=82.104pt,yshift=-151.820pt]current page.north west) {{\fontsize{12.000pt}{12.720pt}\selectfont \bfseries 4.5.}};
\node[anchor=base west,inner sep=0pt,outer sep=0pt] at ([xshift=104.660pt,yshift=-151.820pt]current page.north west) {{\fontsize{12.000pt}{12.720pt}\selectfont \bfseries \makebox[233.09pt][s]{Evaluation Metrics and Statistical Validation}}};
\node[anchor=base west,inner sep=0pt,outer sep=0pt] at ([xshift=82.104pt,yshift=-177.260pt]current page.north west) {{\fontsize{11.040pt}{11.702pt}\selectfont  \makebox[433.95pt][s]{Model performance was evaluated using Accuracy, Precision, Recall, F1-score, ROC-AUC, and}}};
\node[anchor=base west,inner sep=0pt,outer sep=0pt] at ([xshift=82.104pt,yshift=-189.860pt]current page.north west) {{\fontsize{11.040pt}{11.702pt}\selectfont  \makebox[433.73pt][s]{Precision--Recall AUC (PR-AUC). Given the substantial class imbalance in the dataset, particular}}};
\node[anchor=base west,inner sep=0pt,outer sep=0pt] at ([xshift=82.104pt,yshift=-202.580pt]current page.north west) {{\fontsize{11.040pt}{11.702pt}\selectfont  \makebox[433.97pt][s]{emphasis was placed on Recall, F1-score, and PR-AUC when assessing fraud-detection}}};
\node[anchor=base west,inner sep=0pt,outer sep=0pt] at ([xshift=82.104pt,yshift=-215.180pt]current page.north west) {{\fontsize{11.040pt}{11.702pt}\selectfont  \makebox[340.73pt][s]{performance. The principal classification metrics were calculated as follows:}}};
\node[anchor=base west,inner sep=0pt,outer sep=0pt] at ([xshift=82.104pt,yshift=-398.450pt]current page.north west) {{\fontsize{11.040pt}{11.702pt}\selectfont  \makebox[433.70pt][s]{where TP, TN, FP, and FN denote true positives, true negatives, false positives, and false}}};
\node[anchor=base west,inner sep=0pt,outer sep=0pt] at ([xshift=82.104pt,yshift=-411.170pt]current page.north west) {{\fontsize{11.040pt}{11.702pt}\selectfont  \makebox[433.71pt][s]{negatives, respectively. ROC-AUC measures the model's discrimination across classification}}};
\node[anchor=base west,inner sep=0pt,outer sep=0pt] at ([xshift=82.104pt,yshift=-423.770pt]current page.north west) {{\fontsize{11.040pt}{11.702pt}\selectfont  \makebox[433.91pt][s]{thresholds, while PR-AUC summarizes the precision--recall trade-off and is particularly}}};
\node[anchor=base west,inner sep=0pt,outer sep=0pt] at ([xshift=82.104pt,yshift=-436.370pt]current page.north west) {{\fontsize{11.040pt}{11.702pt}\selectfont  \makebox[433.92pt][s]{informative for imbalanced fraud-detection problems.The extended agent-aware evaluation}}};
\node[anchor=base west,inner sep=0pt,outer sep=0pt] at ([xshift=82.104pt,yshift=-449.090pt]current page.north west) {{\fontsize{11.040pt}{11.702pt}\selectfont  \makebox[433.74pt][s]{additionally measured attack-specific detection rate, authorization outcomes, legitimate}}};
\node[anchor=base west,inner sep=0pt,outer sep=0pt] at ([xshift=82.104pt,yshift=-461.690pt]current page.north west) {{\fontsize{11.040pt}{11.702pt}\selectfont  \makebox[262.13pt][s]{transaction escalation rate, and transaction-value exposure.}}};
\node[anchor=base west,inner sep=0pt,outer sep=0pt] at ([xshift=82.104pt,yshift=-487.030pt]current page.north west) {{\fontsize{11.040pt}{11.702pt}\selectfont  \makebox[214.47pt][s]{Attack-specific detection rate was calculated as:}}};
\node[anchor=base west,inner sep=0pt,outer sep=0pt] at ([xshift=82.104pt,yshift=-545.950pt]current page.north west) {{\fontsize{11.040pt}{11.702pt}\selectfont  Where}};
\node[anchor=base west,inner sep=0pt,outer sep=0pt] at ([xshift=115.700pt,yshift=-545.950pt]current page.north west) {{\fontsize{11.040pt}{11.702pt}\selectfont  N}};
\node[anchor=base west,inner sep=0pt,outer sep=0pt] at ([xshift=123.260pt,yshift=-548.230pt]current page.north west) {{\fontsize{8.040pt}{8.522pt}\selectfont  detected}};
\node[anchor=base west,inner sep=0pt,outer sep=0pt] at ([xshift=157.220pt,yshift=-545.950pt]current page.north west) {{\fontsize{11.040pt}{11.702pt}\selectfont  \makebox[261.99pt][s]{represents the number of detected attack transactions and}}};
\node[anchor=base west,inner sep=0pt,outer sep=0pt] at ([xshift=419.350pt,yshift=-545.950pt]current page.north west) {{\fontsize{11.040pt}{11.702pt}\selectfont  N}};
\node[anchor=base west,inner sep=0pt,outer sep=0pt] at ([xshift=426.910pt,yshift=-548.230pt]current page.north west) {{\fontsize{8.040pt}{8.522pt}\selectfont  attack}};
\node[anchor=base west,inner sep=0pt,outer sep=0pt] at ([xshift=451.180pt,yshift=-545.950pt]current page.north west) {{\fontsize{11.040pt}{11.702pt}\selectfont  represents the}};
\node[anchor=base west,inner sep=0pt,outer sep=0pt] at ([xshift=82.104pt,yshift=-558.550pt]current page.north west) {{\fontsize{11.040pt}{11.702pt}\selectfont  \makebox[394.76pt][s]{total number of transactions associated with the corresponding simulated attack scenario.}}};
\node[anchor=base west,inner sep=0pt,outer sep=0pt] at ([xshift=82.104pt,yshift=-583.870pt]current page.north west) {{\fontsize{11.040pt}{11.702pt}\selectfont  \makebox[433.68pt][s]{To evaluate the contribution and stability of the feature groups, ablation experiments were}}};
\node[anchor=base west,inner sep=0pt,outer sep=0pt] at ([xshift=82.104pt,yshift=-596.470pt]current page.north west) {{\fontsize{11.040pt}{11.702pt}\selectfont  \makebox[433.62pt][s]{conducted across 10 paired runs using three configurations: base transaction features, base plus}}};
\node[anchor=base west,inner sep=0pt,outer sep=0pt] at ([xshift=82.104pt,yshift=-609.190pt]current page.north west) {{\fontsize{11.040pt}{11.702pt}\selectfont  \makebox[433.89pt][s]{behavioral features, and base plus behavioral and agent-specific features. Differences between}}};
\node[anchor=base west,inner sep=0pt,outer sep=0pt] at ([xshift=82.104pt,yshift=-621.790pt]current page.north west) {{\fontsize{11.040pt}{11.702pt}\selectfont  \makebox[433.86pt][s]{configurations were assessed using paired t-tests and Wilcoxon signed-rank tests, with statistical}}};
\node[anchor=base west,inner sep=0pt,outer sep=0pt] at ([xshift=82.104pt,yshift=-634.510pt]current page.north west) {{\fontsize{11.040pt}{11.702pt}\selectfont  \makebox[147.99pt][s]{significance evaluated at p<0.05.}}};
\node[anchor=base west,inner sep=0pt,outer sep=0pt] at ([xshift=82.104pt,yshift=-660.700pt]current page.north west) {{\fontsize{12.000pt}{12.720pt}\selectfont \bfseries 4.6.}};
\node[anchor=base west,inner sep=0pt,outer sep=0pt] at ([xshift=104.660pt,yshift=-660.700pt]current page.north west) {{\fontsize{12.000pt}{12.720pt}\selectfont \bfseries \makebox[218.69pt][s]{Explainable Artificial Intelligence (SHAP)}}};
\node[anchor=base west,inner sep=0pt,outer sep=0pt] at ([xshift=82.104pt,yshift=-686.260pt]current page.north west) {{\fontsize{11.040pt}{11.702pt}\selectfont  \makebox[433.80pt][s]{To improve transparency in fraud-risk assessment, ACAGS applies SHapley Additive}}};
\node[anchor=base west,inner sep=0pt,outer sep=0pt] at ([xshift=82.104pt,yshift=-698.860pt]current page.north west) {{\fontsize{11.040pt}{11.702pt}\selectfont  \makebox[433.61pt][s]{exPlanations (SHAP) [8] to the selected XGBoost model. SHAP estimates the contribution of}}};
\node[anchor=base west,inner sep=0pt,outer sep=0pt] at ([xshift=82.104pt,yshift=-711.580pt]current page.north west) {{\fontsize{11.040pt}{11.702pt}\selectfont  \makebox[433.62pt][s]{individual features to model predictions, enabling both global interpretation of influential}}};
\node[anchor=base west,inner sep=0pt,outer sep=0pt] at ([xshift=82.104pt,yshift=-724.180pt]current page.north west) {{\fontsize{11.040pt}{11.702pt}\selectfont  \makebox[282.29pt][s]{variables and transaction-level explanation of fraud-risk scores.}}};
\node[inner sep=0pt,outer sep=0pt] at ([xshift=302.600pt,yshift=-247.000pt]current page.north west) {$\displaystyle \mathrm{Accuracy}=\frac{\mathrm{TP}+\mathrm{TN}}{\mathrm{TP}+\mathrm{TN}+\mathrm{FP}+\mathrm{FN}}$};
\node[inner sep=0pt,outer sep=0pt] at ([xshift=302.000pt,yshift=-288.000pt]current page.north west) {$\displaystyle \mathrm{Precision}=\frac{\mathrm{TP}}{\mathrm{TP}+\mathrm{FP}}$};
\node[inner sep=0pt,outer sep=0pt] at ([xshift=302.000pt,yshift=-326.000pt]current page.north west) {$\displaystyle \mathrm{Recall}=\frac{\mathrm{TP}}{\mathrm{TP}+\mathrm{FN}}$};
\node[inner sep=0pt,outer sep=0pt] at ([xshift=301.000pt,yshift=-363.000pt]current page.north west) {$\displaystyle \mathrm{F1}=2\times\frac{\mathrm{Precision}\times\mathrm{Recall}}{\mathrm{Precision}+\mathrm{Recall}}$};
\node[inner sep=0pt,outer sep=0pt] at ([xshift=299.000pt,yshift=-523.000pt]current page.north west) {$\displaystyle \mathrm{Attack\ Detection\ Rate}=\frac{N_{\mathrm{detected}}}{N_{\mathrm{attack}}}\times100$};
\end{tikzpicture}
\null
\newpage
\begin{tikzpicture}[remember picture,overlay]
\node[anchor=base west,inner sep=0pt,outer sep=0pt] at ([xshift=127.580pt,yshift=-81.360pt]current page.north west) {{\fontsize{9.960pt}{10.558pt}\selectfont  \makebox[342.77pt][s]{International Journal on Cybernetics \& Informatics (IJCI) Vol.15, No.4, August 2026}}};
\node[anchor=base west,inner sep=0pt,outer sep=0pt] at ([xshift=503.380pt,yshift=-767.856pt]current page.north west) {{\fontsize{9.960pt}{10.558pt}\selectfont  94}};
\node[anchor=base west,inner sep=0pt,outer sep=0pt] at ([xshift=82.104pt,yshift=-100.220pt]current page.north west) {{\fontsize{11.040pt}{11.702pt}\selectfont  \makebox[433.85pt][s]{In ACAGS, SHAP is used only for interpretation and does not directly determine the payment}}};
\node[anchor=base west,inner sep=0pt,outer sep=0pt] at ([xshift=82.104pt,yshift=-112.820pt]current page.north west) {{\fontsize{11.040pt}{11.702pt}\selectfont  \makebox[433.62pt][s]{authorization outcome. The fraud probability generated by XGBoost is passed separately to the}}};
\node[anchor=base west,inner sep=0pt,outer sep=0pt] at ([xshift=82.104pt,yshift=-125.540pt]current page.north west) {{\fontsize{11.040pt}{11.702pt}\selectfont  \makebox[433.83pt][s]{authorization mechanism, where it is evaluated alongside agent-specific controls such as identity}}};
\node[anchor=base west,inner sep=0pt,outer sep=0pt] at ([xshift=82.104pt,yshift=-138.140pt]current page.north west) {{\fontsize{11.040pt}{11.702pt}\selectfont  \makebox[433.66pt][s]{validation, delegated spending limits, authorization scope, and delegation constraints. This}}};
\node[anchor=base west,inner sep=0pt,outer sep=0pt] at ([xshift=82.104pt,yshift=-150.860pt]current page.north west) {{\fontsize{11.040pt}{11.702pt}\selectfont  \makebox[431.43pt][s]{separation preserves a clear distinction between model explanationandauthorization decision-}}};
\node[anchor=base west,inner sep=0pt,outer sep=0pt] at ([xshift=82.104pt,yshift=-163.460pt]current page.north west) {{\fontsize{11.040pt}{11.702pt}\selectfont  making}};
\node[anchor=base west,inner sep=0pt,outer sep=0pt] at ([xshift=115.220pt,yshift=-163.460pt]current page.north west) {{\fontsize{11.040pt}{11.702pt}\selectfont \bfseries .}};
\node[anchor=base west,inner sep=0pt,outer sep=0pt] at ([xshift=82.104pt,yshift=-190.820pt]current page.north west) {{\fontsize{12.000pt}{12.720pt}\selectfont \bfseries 4.7.}};
\node[anchor=base west,inner sep=0pt,outer sep=0pt] at ([xshift=104.660pt,yshift=-190.820pt]current page.north west) {{\fontsize{12.000pt}{12.720pt}\selectfont \bfseries \makebox[200.07pt][s]{ACAGS Authorization Decision Policy}}};
\node[anchor=base west,inner sep=0pt,outer sep=0pt] at ([xshift=82.104pt,yshift=-216.380pt]current page.north west) {{\fontsize{11.040pt}{11.702pt}\selectfont  \makebox[433.63pt][s]{The final ACAGS authorization mechanism combines the fraud probability generated by the}}};
\node[anchor=base west,inner sep=0pt,outer sep=0pt] at ([xshift=82.104pt,yshift=-228.980pt]current page.north west) {{\fontsize{11.040pt}{11.702pt}\selectfont  \makebox[433.76pt][s]{selected XGBoost model with agent-specific authorization controls. These controls evaluate}}};
\node[anchor=base west,inner sep=0pt,outer sep=0pt] at ([xshift=82.104pt,yshift=-241.700pt]current page.north west) {{\fontsize{11.040pt}{11.702pt}\selectfont  \makebox[433.66pt][s]{agent identity, delegated spending limits, authorization scope, and delegation constraints before}}};
\node[anchor=base west,inner sep=0pt,outer sep=0pt] at ([xshift=82.104pt,yshift=-254.300pt]current page.north west) {{\fontsize{11.040pt}{11.702pt}\selectfont  \makebox[433.64pt][s]{determining the transaction outcome. SHAP explanations remain separate from this decision}}};
\node[anchor=base west,inner sep=0pt,outer sep=0pt] at ([xshift=82.104pt,yshift=-266.900pt]current page.north west) {{\fontsize{11.040pt}{11.702pt}\selectfont  \makebox[319.01pt][s]{logic and are used only to interpret the underlying fraud-risk prediction.}}};
\node[anchor=base west,inner sep=0pt,outer sep=0pt] at ([xshift=82.104pt,yshift=-292.250pt]current page.north west) {{\fontsize{11.040pt}{11.702pt}\selectfont  \makebox[250.19pt][s]{Each transaction is assigned one of three outcomes:}}};
\node[anchor=base west,inner sep=0pt,outer sep=0pt] at ([xshift=335.350pt,yshift=-292.250pt]current page.north west) {{\fontsize{11.040pt}{11.702pt}\selectfont \bfseries \makebox[175.16pt][s]{APPROVE, REVIEW, or REJECT}}};
\node[anchor=base west,inner sep=0pt,outer sep=0pt] at ([xshift=510.580pt,yshift=-292.250pt]current page.north west) {{\fontsize{11.040pt}{11.702pt}\selectfont  .}};
\node[anchor=base west,inner sep=0pt,outer sep=0pt] at ([xshift=82.104pt,yshift=-304.970pt]current page.north west) {{\fontsize{11.040pt}{11.702pt}\selectfont  \makebox[433.75pt][s]{Transactions satisfying the required agent-authorization constraints and exhibiting low fraud risk}}};
\node[anchor=base west,inner sep=0pt,outer sep=0pt] at ([xshift=82.104pt,yshift=-317.570pt]current page.north west) {{\fontsize{11.040pt}{11.702pt}\selectfont  \makebox[433.63pt][s]{are approved, while transactions with elevated behavioral or fraud risk are routed for review.}}};
\node[anchor=base west,inner sep=0pt,outer sep=0pt] at ([xshift=82.104pt,yshift=-330.290pt]current page.north west) {{\fontsize{11.040pt}{11.702pt}\selectfont  \makebox[433.65pt][s]{Transactions involving explicit authorization violations, such as invalid delegation or activity}}};
\node[anchor=base west,inner sep=0pt,outer sep=0pt] at ([xshift=82.104pt,yshift=-342.890pt]current page.north west) {{\fontsize{11.040pt}{11.702pt}\selectfont  \makebox[433.64pt][s]{outside the permitted scope, are rejected. This layered decision mechanism allows ACAGS to}}};
\node[anchor=base west,inner sep=0pt,outer sep=0pt] at ([xshift=82.104pt,yshift=-355.490pt]current page.north west) {{\fontsize{11.040pt}{11.702pt}\selectfont  \makebox[433.64pt][s]{distinguish uncertain behavioral risk from explicit policy violations rather than treating all}}};
\node[anchor=base west,inner sep=0pt,outer sep=0pt] at ([xshift=82.104pt,yshift=-368.210pt]current page.north west) {{\fontsize{11.040pt}{11.702pt}\selectfont  \makebox[155.91pt][s]{suspicious transactions identically.}}};
\node[anchor=base west,inner sep=0pt,outer sep=0pt] at ([xshift=82.104pt,yshift=-393.530pt]current page.north west) {{\fontsize{11.040pt}{11.702pt}\selectfont  \makebox[433.99pt][s]{The authorization logic was applied consistently across the simulated agentic-commerce}}};
\node[anchor=base west,inner sep=0pt,outer sep=0pt] at ([xshift=82.104pt,yshift=-406.130pt]current page.north west) {{\fontsize{11.040pt}{11.702pt}\selectfont  \makebox[433.86pt][s]{transactions, enabling subsequent evaluation of attack-specific authorization outcomes, legitimate}}};
\node[anchor=base west,inner sep=0pt,outer sep=0pt] at ([xshift=82.104pt,yshift=-418.730pt]current page.north west) {{\fontsize{11.040pt}{11.702pt}\selectfont  \makebox[242.93pt][s]{transaction escalation, and transaction-value exposure.}}};
\node[anchor=base west,inner sep=0pt,outer sep=0pt] at ([xshift=82.104pt,yshift=-446.930pt]current page.north west) {{\fontsize{14.040pt}{14.882pt}\selectfont \bfseries 5.}};
\node[anchor=base west,inner sep=0pt,outer sep=0pt] at ([xshift=100.100pt,yshift=-446.930pt]current page.north west) {{\fontsize{14.040pt}{14.882pt}\selectfont \bfseries P}};
\node[anchor=base west,inner sep=0pt,outer sep=0pt] at ([xshift=108.740pt,yshift=-446.930pt]current page.north west) {{\fontsize{11.040pt}{11.702pt}\selectfont \bfseries ERFORMANCE}};
\node[anchor=base west,inner sep=0pt,outer sep=0pt] at ([xshift=191.570pt,yshift=-446.930pt]current page.north west) {{\fontsize{14.040pt}{14.882pt}\selectfont \bfseries E}};
\node[anchor=base west,inner sep=0pt,outer sep=0pt] at ([xshift=200.930pt,yshift=-446.930pt]current page.north west) {{\fontsize{11.040pt}{11.702pt}\selectfont \bfseries VALUATION}};
\node[anchor=base west,inner sep=0pt,outer sep=0pt] at ([xshift=82.104pt,yshift=-472.870pt]current page.north west) {{\fontsize{11.040pt}{11.702pt}\selectfont  \makebox[433.63pt][s]{The performance evaluation examines the effectiveness of ACAGS through comparative model}}};
\node[anchor=base west,inner sep=0pt,outer sep=0pt] at ([xshift=82.104pt,yshift=-485.470pt]current page.north west) {{\fontsize{11.040pt}{11.702pt}\selectfont  \makebox[433.92pt][s]{assessment, behavioral feature analysis, explainability, simulated agent-attack evaluation,}}};
\node[anchor=base west,inner sep=0pt,outer sep=0pt] at ([xshift=82.104pt,yshift=-498.190pt]current page.north west) {{\fontsize{11.040pt}{11.702pt}\selectfont  \makebox[433.63pt][s]{ablation analysis, and authorization outcomes. The evaluated classifiers are assessed using}}};
\node[anchor=base west,inner sep=0pt,outer sep=0pt] at ([xshift=82.104pt,yshift=-510.790pt]current page.north west) {{\fontsize{11.040pt}{11.702pt}\selectfont  \makebox[433.65pt][s]{complementary performance metrics suited to highly imbalanced fraud data, while SHAP}}};
\node[anchor=base west,inner sep=0pt,outer sep=0pt] at ([xshift=82.104pt,yshift=-523.390pt]current page.north west) {{\fontsize{11.040pt}{11.702pt}\selectfont  \makebox[433.80pt][s]{analysis examines the factors influencing fraud-risk predictions. Together, these evaluations}}};
\node[anchor=base west,inner sep=0pt,outer sep=0pt] at ([xshift=82.104pt,yshift=-536.110pt]current page.north west) {{\fontsize{11.040pt}{11.702pt}\selectfont  \makebox[433.81pt][s]{assess the predictive performance, interpretability, and agent-aware authorization behavior of the}}};
\node[anchor=base west,inner sep=0pt,outer sep=0pt] at ([xshift=82.104pt,yshift=-548.710pt]current page.north west) {{\fontsize{11.040pt}{11.702pt}\selectfont  \makebox[96.84pt][s]{proposed framework.}}};
\node[anchor=base west,inner sep=0pt,outer sep=0pt] at ([xshift=82.104pt,yshift=-574.990pt]current page.north west) {{\fontsize{12.000pt}{12.720pt}\selectfont \bfseries 5.1.}};
\node[anchor=base west,inner sep=0pt,outer sep=0pt] at ([xshift=104.660pt,yshift=-574.990pt]current page.north west) {{\fontsize{12.000pt}{12.720pt}\selectfont \bfseries \makebox[202.95pt][s]{Baseline Fraud-Detection Performance}}};
\node[anchor=base west,inner sep=0pt,outer sep=0pt] at ([xshift=82.104pt,yshift=-600.430pt]current page.north west) {{\fontsize{11.040pt}{11.702pt}\selectfont  \makebox[433.74pt][s]{The baseline fraud-detection component of ACAGS was evaluated using four supervised}}};
\node[anchor=base west,inner sep=0pt,outer sep=0pt] at ([xshift=82.104pt,yshift=-613.150pt]current page.north west) {{\fontsize{11.040pt}{11.702pt}\selectfont  \makebox[433.90pt][s]{machine learning classifiers under identical experimental conditions. Model performance was}}};
\node[anchor=base west,inner sep=0pt,outer sep=0pt] at ([xshift=82.104pt,yshift=-625.750pt]current page.north west) {{\fontsize{11.040pt}{11.702pt}\selectfont  \makebox[433.95pt][s]{assessed using Accuracy, Precision, Recall, F1-score, ROC-AUC, and Precision--Recall AUC}}};
\node[anchor=base west,inner sep=0pt,outer sep=0pt] at ([xshift=82.104pt,yshift=-638.470pt]current page.north west) {{\fontsize{11.040pt}{11.702pt}\selectfont  \makebox[284.45pt][s]{(PR-AUC). The comparative results are summarized in Table 2.}}};
\end{tikzpicture}
\null
\newpage
\begin{tikzpicture}[remember picture,overlay]
\fill[black!15] ([xshift=101.420pt,yshift=-115.820pt]current page.north west) rectangle ([xshift=184.220pt,yshift=-138.860pt]current page.north west);
\fill[black!15] ([xshift=106.580pt,yshift=-115.820pt]current page.north west) rectangle ([xshift=179.060pt,yshift=-127.340pt]current page.north west);
\fill[black!15] ([xshift=184.700pt,yshift=-115.820pt]current page.north west) rectangle ([xshift=235.244pt,yshift=-138.860pt]current page.north west);
\fill[black!15] ([xshift=189.860pt,yshift=-115.820pt]current page.north west) rectangle ([xshift=230.084pt,yshift=-127.340pt]current page.north west);
\fill[black!15] ([xshift=189.860pt,yshift=-127.340pt]current page.north west) rectangle ([xshift=230.084pt,yshift=-138.860pt]current page.north west);
\fill[black!15] ([xshift=235.730pt,yshift=-115.820pt]current page.north west) rectangle ([xshift=285.050pt,yshift=-138.860pt]current page.north west);
\fill[black!15] ([xshift=240.890pt,yshift=-115.820pt]current page.north west) rectangle ([xshift=279.890pt,yshift=-127.340pt]current page.north west);
\fill[black!15] ([xshift=240.890pt,yshift=-127.340pt]current page.north west) rectangle ([xshift=279.890pt,yshift=-138.860pt]current page.north west);
\fill[black!15] ([xshift=285.530pt,yshift=-115.820pt]current page.north west) rectangle ([xshift=328.394pt,yshift=-138.860pt]current page.north west);
\fill[black!15] ([xshift=290.690pt,yshift=-115.820pt]current page.north west) rectangle ([xshift=323.234pt,yshift=-127.340pt]current page.north west);
\fill[black!15] ([xshift=328.870pt,yshift=-115.820pt]current page.north west) rectangle ([xshift=378.070pt,yshift=-138.860pt]current page.north west);
\fill[black!15] ([xshift=334.030pt,yshift=-115.820pt]current page.north west) rectangle ([xshift=372.910pt,yshift=-127.340pt]current page.north west);
\fill[black!15] ([xshift=378.550pt,yshift=-115.820pt]current page.north west) rectangle ([xshift=436.030pt,yshift=-138.860pt]current page.north west);
\fill[black!15] ([xshift=383.710pt,yshift=-115.820pt]current page.north west) rectangle ([xshift=430.870pt,yshift=-127.340pt]current page.north west);
\fill[black!15] ([xshift=383.710pt,yshift=-127.340pt]current page.north west) rectangle ([xshift=430.870pt,yshift=-138.860pt]current page.north west);
\fill[black!15] ([xshift=436.510pt,yshift=-115.820pt]current page.north west) rectangle ([xshift=494.014pt,yshift=-138.860pt]current page.north west);
\fill[black!15] ([xshift=441.670pt,yshift=-115.820pt]current page.north west) rectangle ([xshift=488.854pt,yshift=-127.340pt]current page.north west);
\fill[black] ([xshift=100.940pt,yshift=-115.340pt]current page.north west) rectangle ([xshift=101.420pt,yshift=-115.820pt]current page.north west);
\fill[black] ([xshift=100.940pt,yshift=-115.340pt]current page.north west) rectangle ([xshift=101.420pt,yshift=-115.820pt]current page.north west);
\fill[black] ([xshift=101.420pt,yshift=-115.340pt]current page.north west) rectangle ([xshift=184.220pt,yshift=-115.820pt]current page.north west);
\fill[black] ([xshift=184.220pt,yshift=-115.340pt]current page.north west) rectangle ([xshift=184.700pt,yshift=-115.820pt]current page.north west);
\fill[black] ([xshift=184.700pt,yshift=-115.340pt]current page.north west) rectangle ([xshift=235.244pt,yshift=-115.820pt]current page.north west);
\fill[black] ([xshift=235.250pt,yshift=-115.340pt]current page.north west) rectangle ([xshift=235.730pt,yshift=-115.820pt]current page.north west);
\fill[black] ([xshift=235.730pt,yshift=-115.340pt]current page.north west) rectangle ([xshift=285.050pt,yshift=-115.820pt]current page.north west);
\fill[black] ([xshift=285.050pt,yshift=-115.340pt]current page.north west) rectangle ([xshift=285.530pt,yshift=-115.820pt]current page.north west);
\fill[black] ([xshift=285.530pt,yshift=-115.340pt]current page.north west) rectangle ([xshift=328.394pt,yshift=-115.820pt]current page.north west);
\fill[black] ([xshift=328.390pt,yshift=-115.340pt]current page.north west) rectangle ([xshift=328.870pt,yshift=-115.820pt]current page.north west);
\fill[black] ([xshift=328.870pt,yshift=-115.340pt]current page.north west) rectangle ([xshift=378.070pt,yshift=-115.820pt]current page.north west);
\fill[black] ([xshift=378.070pt,yshift=-115.340pt]current page.north west) rectangle ([xshift=378.550pt,yshift=-115.820pt]current page.north west);
\fill[black] ([xshift=378.550pt,yshift=-115.340pt]current page.north west) rectangle ([xshift=436.030pt,yshift=-115.820pt]current page.north west);
\fill[black] ([xshift=436.030pt,yshift=-115.340pt]current page.north west) rectangle ([xshift=436.510pt,yshift=-115.820pt]current page.north west);
\fill[black] ([xshift=436.510pt,yshift=-115.340pt]current page.north west) rectangle ([xshift=494.014pt,yshift=-115.820pt]current page.north west);
\fill[black] ([xshift=494.020pt,yshift=-115.340pt]current page.north west) rectangle ([xshift=494.500pt,yshift=-115.820pt]current page.north west);
\fill[black] ([xshift=494.020pt,yshift=-115.340pt]current page.north west) rectangle ([xshift=494.500pt,yshift=-115.820pt]current page.north west);
\fill[black] ([xshift=100.940pt,yshift=-115.820pt]current page.north west) rectangle ([xshift=101.420pt,yshift=-138.860pt]current page.north west);
\fill[black] ([xshift=184.220pt,yshift=-115.820pt]current page.north west) rectangle ([xshift=184.700pt,yshift=-138.860pt]current page.north west);
\fill[black] ([xshift=235.250pt,yshift=-115.820pt]current page.north west) rectangle ([xshift=235.730pt,yshift=-138.860pt]current page.north west);
\fill[black] ([xshift=285.050pt,yshift=-115.820pt]current page.north west) rectangle ([xshift=285.530pt,yshift=-138.860pt]current page.north west);
\fill[black] ([xshift=328.390pt,yshift=-115.820pt]current page.north west) rectangle ([xshift=328.870pt,yshift=-138.860pt]current page.north west);
\fill[black] ([xshift=378.070pt,yshift=-115.820pt]current page.north west) rectangle ([xshift=378.550pt,yshift=-138.860pt]current page.north west);
\fill[black] ([xshift=436.030pt,yshift=-115.820pt]current page.north west) rectangle ([xshift=436.510pt,yshift=-138.860pt]current page.north west);
\fill[black] ([xshift=494.020pt,yshift=-115.820pt]current page.north west) rectangle ([xshift=494.500pt,yshift=-138.860pt]current page.north west);
\fill[black] ([xshift=100.940pt,yshift=-138.860pt]current page.north west) rectangle ([xshift=101.420pt,yshift=-139.340pt]current page.north west);
\fill[black] ([xshift=101.420pt,yshift=-138.860pt]current page.north west) rectangle ([xshift=184.220pt,yshift=-139.340pt]current page.north west);
\fill[black] ([xshift=184.220pt,yshift=-138.860pt]current page.north west) rectangle ([xshift=184.700pt,yshift=-139.340pt]current page.north west);
\fill[black] ([xshift=184.700pt,yshift=-138.860pt]current page.north west) rectangle ([xshift=235.244pt,yshift=-139.340pt]current page.north west);
\fill[black] ([xshift=235.250pt,yshift=-138.860pt]current page.north west) rectangle ([xshift=235.730pt,yshift=-139.340pt]current page.north west);
\fill[black] ([xshift=235.730pt,yshift=-138.860pt]current page.north west) rectangle ([xshift=285.050pt,yshift=-139.340pt]current page.north west);
\fill[black] ([xshift=285.050pt,yshift=-138.860pt]current page.north west) rectangle ([xshift=285.530pt,yshift=-139.340pt]current page.north west);
\fill[black] ([xshift=285.530pt,yshift=-138.860pt]current page.north west) rectangle ([xshift=328.394pt,yshift=-139.340pt]current page.north west);
\fill[black] ([xshift=328.390pt,yshift=-138.860pt]current page.north west) rectangle ([xshift=328.870pt,yshift=-139.340pt]current page.north west);
\fill[black] ([xshift=328.870pt,yshift=-138.860pt]current page.north west) rectangle ([xshift=378.070pt,yshift=-139.340pt]current page.north west);
\fill[black] ([xshift=378.070pt,yshift=-138.860pt]current page.north west) rectangle ([xshift=378.550pt,yshift=-139.340pt]current page.north west);
\fill[black] ([xshift=378.550pt,yshift=-138.860pt]current page.north west) rectangle ([xshift=436.030pt,yshift=-139.340pt]current page.north west);
\fill[black] ([xshift=436.030pt,yshift=-138.860pt]current page.north west) rectangle ([xshift=436.510pt,yshift=-139.340pt]current page.north west);
\fill[black] ([xshift=436.510pt,yshift=-138.860pt]current page.north west) rectangle ([xshift=494.014pt,yshift=-139.340pt]current page.north west);
\fill[black] ([xshift=494.020pt,yshift=-138.860pt]current page.north west) rectangle ([xshift=494.500pt,yshift=-139.340pt]current page.north west);
\fill[black] ([xshift=100.940pt,yshift=-139.340pt]current page.north west) rectangle ([xshift=101.420pt,yshift=-162.260pt]current page.north west);
\fill[black] ([xshift=184.220pt,yshift=-139.340pt]current page.north west) rectangle ([xshift=184.700pt,yshift=-162.260pt]current page.north west);
\fill[black] ([xshift=235.250pt,yshift=-139.340pt]current page.north west) rectangle ([xshift=235.730pt,yshift=-162.260pt]current page.north west);
\fill[black] ([xshift=285.050pt,yshift=-139.340pt]current page.north west) rectangle ([xshift=285.530pt,yshift=-162.260pt]current page.north west);
\fill[black] ([xshift=328.390pt,yshift=-139.340pt]current page.north west) rectangle ([xshift=328.870pt,yshift=-162.260pt]current page.north west);
\fill[black] ([xshift=378.070pt,yshift=-139.340pt]current page.north west) rectangle ([xshift=378.550pt,yshift=-162.260pt]current page.north west);
\fill[black] ([xshift=436.030pt,yshift=-139.340pt]current page.north west) rectangle ([xshift=436.510pt,yshift=-162.260pt]current page.north west);
\fill[black] ([xshift=494.020pt,yshift=-139.340pt]current page.north west) rectangle ([xshift=494.500pt,yshift=-162.260pt]current page.north west);
\fill[black] ([xshift=100.940pt,yshift=-162.260pt]current page.north west) rectangle ([xshift=101.420pt,yshift=-162.740pt]current page.north west);
\fill[black] ([xshift=101.420pt,yshift=-162.260pt]current page.north west) rectangle ([xshift=184.220pt,yshift=-162.740pt]current page.north west);
\fill[black] ([xshift=184.220pt,yshift=-162.260pt]current page.north west) rectangle ([xshift=184.700pt,yshift=-162.740pt]current page.north west);
\fill[black] ([xshift=184.700pt,yshift=-162.260pt]current page.north west) rectangle ([xshift=235.244pt,yshift=-162.740pt]current page.north west);
\fill[black] ([xshift=235.250pt,yshift=-162.260pt]current page.north west) rectangle ([xshift=235.730pt,yshift=-162.740pt]current page.north west);
\fill[black] ([xshift=235.730pt,yshift=-162.260pt]current page.north west) rectangle ([xshift=285.050pt,yshift=-162.740pt]current page.north west);
\fill[black] ([xshift=285.050pt,yshift=-162.260pt]current page.north west) rectangle ([xshift=285.530pt,yshift=-162.740pt]current page.north west);
\fill[black] ([xshift=285.530pt,yshift=-162.260pt]current page.north west) rectangle ([xshift=328.394pt,yshift=-162.740pt]current page.north west);
\fill[black] ([xshift=328.390pt,yshift=-162.260pt]current page.north west) rectangle ([xshift=328.870pt,yshift=-162.740pt]current page.north west);
\fill[black] ([xshift=328.870pt,yshift=-162.260pt]current page.north west) rectangle ([xshift=378.070pt,yshift=-162.740pt]current page.north west);
\fill[black] ([xshift=378.070pt,yshift=-162.260pt]current page.north west) rectangle ([xshift=378.550pt,yshift=-162.740pt]current page.north west);
\fill[black] ([xshift=378.550pt,yshift=-162.260pt]current page.north west) rectangle ([xshift=436.030pt,yshift=-162.740pt]current page.north west);
\fill[black] ([xshift=436.030pt,yshift=-162.260pt]current page.north west) rectangle ([xshift=436.510pt,yshift=-162.740pt]current page.north west);
\fill[black] ([xshift=436.510pt,yshift=-162.260pt]current page.north west) rectangle ([xshift=494.014pt,yshift=-162.740pt]current page.north west);
\fill[black] ([xshift=494.020pt,yshift=-162.260pt]current page.north west) rectangle ([xshift=494.500pt,yshift=-162.740pt]current page.north west);
\fill[black] ([xshift=100.940pt,yshift=-162.740pt]current page.north west) rectangle ([xshift=101.420pt,yshift=-174.260pt]current page.north west);
\fill[black] ([xshift=184.220pt,yshift=-162.740pt]current page.north west) rectangle ([xshift=184.700pt,yshift=-174.260pt]current page.north west);
\fill[black] ([xshift=235.250pt,yshift=-162.740pt]current page.north west) rectangle ([xshift=235.730pt,yshift=-174.260pt]current page.north west);
\fill[black] ([xshift=285.050pt,yshift=-162.740pt]current page.north west) rectangle ([xshift=285.530pt,yshift=-174.260pt]current page.north west);
\fill[black] ([xshift=328.390pt,yshift=-162.740pt]current page.north west) rectangle ([xshift=328.870pt,yshift=-174.260pt]current page.north west);
\fill[black] ([xshift=378.070pt,yshift=-162.740pt]current page.north west) rectangle ([xshift=378.550pt,yshift=-174.260pt]current page.north west);
\fill[black] ([xshift=436.030pt,yshift=-162.740pt]current page.north west) rectangle ([xshift=436.510pt,yshift=-174.260pt]current page.north west);
\fill[black] ([xshift=494.020pt,yshift=-162.740pt]current page.north west) rectangle ([xshift=494.500pt,yshift=-174.260pt]current page.north west);
\fill[black] ([xshift=100.940pt,yshift=-174.260pt]current page.north west) rectangle ([xshift=101.420pt,yshift=-174.740pt]current page.north west);
\fill[black] ([xshift=101.420pt,yshift=-174.260pt]current page.north west) rectangle ([xshift=184.220pt,yshift=-174.740pt]current page.north west);
\fill[black] ([xshift=184.220pt,yshift=-174.260pt]current page.north west) rectangle ([xshift=184.700pt,yshift=-174.740pt]current page.north west);
\fill[black] ([xshift=184.700pt,yshift=-174.260pt]current page.north west) rectangle ([xshift=235.244pt,yshift=-174.740pt]current page.north west);
\fill[black] ([xshift=235.250pt,yshift=-174.260pt]current page.north west) rectangle ([xshift=235.730pt,yshift=-174.740pt]current page.north west);
\fill[black] ([xshift=235.730pt,yshift=-174.260pt]current page.north west) rectangle ([xshift=285.050pt,yshift=-174.740pt]current page.north west);
\fill[black] ([xshift=285.050pt,yshift=-174.260pt]current page.north west) rectangle ([xshift=285.530pt,yshift=-174.740pt]current page.north west);
\fill[black] ([xshift=285.530pt,yshift=-174.260pt]current page.north west) rectangle ([xshift=328.394pt,yshift=-174.740pt]current page.north west);
\fill[black] ([xshift=328.390pt,yshift=-174.260pt]current page.north west) rectangle ([xshift=328.870pt,yshift=-174.740pt]current page.north west);
\fill[black] ([xshift=328.870pt,yshift=-174.260pt]current page.north west) rectangle ([xshift=378.070pt,yshift=-174.740pt]current page.north west);
\fill[black] ([xshift=378.070pt,yshift=-174.260pt]current page.north west) rectangle ([xshift=378.550pt,yshift=-174.740pt]current page.north west);
\fill[black] ([xshift=378.550pt,yshift=-174.260pt]current page.north west) rectangle ([xshift=436.030pt,yshift=-174.740pt]current page.north west);
\fill[black] ([xshift=436.030pt,yshift=-174.260pt]current page.north west) rectangle ([xshift=436.510pt,yshift=-174.740pt]current page.north west);
\fill[black] ([xshift=436.510pt,yshift=-174.260pt]current page.north west) rectangle ([xshift=494.014pt,yshift=-174.740pt]current page.north west);
\fill[black] ([xshift=494.020pt,yshift=-174.260pt]current page.north west) rectangle ([xshift=494.500pt,yshift=-174.740pt]current page.north west);
\fill[black] ([xshift=100.940pt,yshift=-174.740pt]current page.north west) rectangle ([xshift=101.420pt,yshift=-186.260pt]current page.north west);
\fill[black] ([xshift=184.220pt,yshift=-174.740pt]current page.north west) rectangle ([xshift=184.700pt,yshift=-186.260pt]current page.north west);
\fill[black] ([xshift=235.250pt,yshift=-174.740pt]current page.north west) rectangle ([xshift=235.730pt,yshift=-186.260pt]current page.north west);
\fill[black] ([xshift=285.050pt,yshift=-174.740pt]current page.north west) rectangle ([xshift=285.530pt,yshift=-186.260pt]current page.north west);
\fill[black] ([xshift=328.390pt,yshift=-174.740pt]current page.north west) rectangle ([xshift=328.870pt,yshift=-186.260pt]current page.north west);
\fill[black] ([xshift=378.070pt,yshift=-174.740pt]current page.north west) rectangle ([xshift=378.550pt,yshift=-186.260pt]current page.north west);
\fill[black] ([xshift=436.030pt,yshift=-174.740pt]current page.north west) rectangle ([xshift=436.510pt,yshift=-186.260pt]current page.north west);
\fill[black] ([xshift=494.020pt,yshift=-174.740pt]current page.north west) rectangle ([xshift=494.500pt,yshift=-186.260pt]current page.north west);
\fill[black] ([xshift=100.940pt,yshift=-186.260pt]current page.north west) rectangle ([xshift=101.420pt,yshift=-186.740pt]current page.north west);
\fill[black] ([xshift=101.420pt,yshift=-186.260pt]current page.north west) rectangle ([xshift=184.220pt,yshift=-186.740pt]current page.north west);
\fill[black] ([xshift=184.220pt,yshift=-186.260pt]current page.north west) rectangle ([xshift=184.700pt,yshift=-186.740pt]current page.north west);
\fill[black] ([xshift=184.700pt,yshift=-186.260pt]current page.north west) rectangle ([xshift=235.244pt,yshift=-186.740pt]current page.north west);
\fill[black] ([xshift=235.250pt,yshift=-186.260pt]current page.north west) rectangle ([xshift=235.730pt,yshift=-186.740pt]current page.north west);
\fill[black] ([xshift=235.730pt,yshift=-186.260pt]current page.north west) rectangle ([xshift=285.050pt,yshift=-186.740pt]current page.north west);
\fill[black] ([xshift=285.050pt,yshift=-186.260pt]current page.north west) rectangle ([xshift=285.530pt,yshift=-186.740pt]current page.north west);
\fill[black] ([xshift=285.530pt,yshift=-186.260pt]current page.north west) rectangle ([xshift=328.394pt,yshift=-186.740pt]current page.north west);
\fill[black] ([xshift=328.390pt,yshift=-186.260pt]current page.north west) rectangle ([xshift=328.870pt,yshift=-186.740pt]current page.north west);
\fill[black] ([xshift=328.870pt,yshift=-186.260pt]current page.north west) rectangle ([xshift=378.070pt,yshift=-186.740pt]current page.north west);
\fill[black] ([xshift=378.070pt,yshift=-186.260pt]current page.north west) rectangle ([xshift=378.550pt,yshift=-186.740pt]current page.north west);
\fill[black] ([xshift=378.550pt,yshift=-186.260pt]current page.north west) rectangle ([xshift=436.030pt,yshift=-186.740pt]current page.north west);
\fill[black] ([xshift=436.030pt,yshift=-186.260pt]current page.north west) rectangle ([xshift=436.510pt,yshift=-186.740pt]current page.north west);
\fill[black] ([xshift=436.510pt,yshift=-186.260pt]current page.north west) rectangle ([xshift=494.014pt,yshift=-186.740pt]current page.north west);
\fill[black] ([xshift=494.020pt,yshift=-186.260pt]current page.north west) rectangle ([xshift=494.500pt,yshift=-186.740pt]current page.north west);
\fill[black] ([xshift=100.940pt,yshift=-186.740pt]current page.north west) rectangle ([xshift=101.420pt,yshift=-198.500pt]current page.north west);
\fill[black] ([xshift=100.940pt,yshift=-198.500pt]current page.north west) rectangle ([xshift=101.420pt,yshift=-198.980pt]current page.north west);
\fill[black] ([xshift=100.940pt,yshift=-198.500pt]current page.north west) rectangle ([xshift=101.420pt,yshift=-198.980pt]current page.north west);
\fill[black] ([xshift=101.420pt,yshift=-198.500pt]current page.north west) rectangle ([xshift=184.220pt,yshift=-198.980pt]current page.north west);
\fill[black] ([xshift=184.220pt,yshift=-186.740pt]current page.north west) rectangle ([xshift=184.700pt,yshift=-198.500pt]current page.north west);
\fill[black] ([xshift=184.220pt,yshift=-198.500pt]current page.north west) rectangle ([xshift=184.700pt,yshift=-198.980pt]current page.north west);
\fill[black] ([xshift=184.700pt,yshift=-198.500pt]current page.north west) rectangle ([xshift=235.244pt,yshift=-198.980pt]current page.north west);
\fill[black] ([xshift=235.250pt,yshift=-186.740pt]current page.north west) rectangle ([xshift=235.730pt,yshift=-198.500pt]current page.north west);
\fill[black] ([xshift=235.250pt,yshift=-198.500pt]current page.north west) rectangle ([xshift=235.730pt,yshift=-198.980pt]current page.north west);
\fill[black] ([xshift=235.730pt,yshift=-198.500pt]current page.north west) rectangle ([xshift=285.050pt,yshift=-198.980pt]current page.north west);
\fill[black] ([xshift=285.050pt,yshift=-186.740pt]current page.north west) rectangle ([xshift=285.530pt,yshift=-198.500pt]current page.north west);
\fill[black] ([xshift=285.050pt,yshift=-198.500pt]current page.north west) rectangle ([xshift=285.530pt,yshift=-198.980pt]current page.north west);
\fill[black] ([xshift=285.530pt,yshift=-198.500pt]current page.north west) rectangle ([xshift=328.394pt,yshift=-198.980pt]current page.north west);
\fill[black] ([xshift=328.390pt,yshift=-186.740pt]current page.north west) rectangle ([xshift=328.870pt,yshift=-198.500pt]current page.north west);
\fill[black] ([xshift=328.390pt,yshift=-198.500pt]current page.north west) rectangle ([xshift=328.870pt,yshift=-198.980pt]current page.north west);
\fill[black] ([xshift=328.870pt,yshift=-198.500pt]current page.north west) rectangle ([xshift=378.070pt,yshift=-198.980pt]current page.north west);
\fill[black] ([xshift=378.070pt,yshift=-186.740pt]current page.north west) rectangle ([xshift=378.550pt,yshift=-198.500pt]current page.north west);
\fill[black] ([xshift=378.070pt,yshift=-198.500pt]current page.north west) rectangle ([xshift=378.550pt,yshift=-198.980pt]current page.north west);
\fill[black] ([xshift=378.550pt,yshift=-198.500pt]current page.north west) rectangle ([xshift=436.030pt,yshift=-198.980pt]current page.north west);
\fill[black] ([xshift=436.030pt,yshift=-186.740pt]current page.north west) rectangle ([xshift=436.510pt,yshift=-198.500pt]current page.north west);
\fill[black] ([xshift=436.030pt,yshift=-198.500pt]current page.north west) rectangle ([xshift=436.510pt,yshift=-198.980pt]current page.north west);
\fill[black] ([xshift=436.510pt,yshift=-198.500pt]current page.north west) rectangle ([xshift=494.014pt,yshift=-198.980pt]current page.north west);
\fill[black] ([xshift=494.020pt,yshift=-186.740pt]current page.north west) rectangle ([xshift=494.500pt,yshift=-198.500pt]current page.north west);
\fill[black] ([xshift=494.020pt,yshift=-198.500pt]current page.north west) rectangle ([xshift=494.500pt,yshift=-198.980pt]current page.north west);
\fill[black] ([xshift=494.020pt,yshift=-198.500pt]current page.north west) rectangle ([xshift=494.500pt,yshift=-198.980pt]current page.north west);
\node[anchor=base west,inner sep=0pt,outer sep=0pt] at ([xshift=127.580pt,yshift=-81.360pt]current page.north west) {{\fontsize{9.960pt}{10.558pt}\selectfont  \makebox[342.77pt][s]{International Journal on Cybernetics \& Informatics (IJCI) Vol.15, No.4, August 2026}}};
\node[anchor=base west,inner sep=0pt,outer sep=0pt] at ([xshift=503.380pt,yshift=-767.856pt]current page.north west) {{\fontsize{9.960pt}{10.558pt}\selectfont  95}};
\node[anchor=base west,inner sep=0pt,outer sep=0pt] at ([xshift=140.660pt,yshift=-99.380pt]current page.north west) {{\fontsize{9.960pt}{10.558pt}\selectfont  \makebox[316.49pt][s]{Table 2. Comparative Performance of Evaluated Machine Learning Classifiers}}};
\node[anchor=base west,inner sep=0pt,outer sep=0pt] at ([xshift=106.580pt,yshift=-125.180pt]current page.north west) {{\fontsize{9.960pt}{10.558pt}\selectfont \bfseries Model}};
\node[anchor=base west,inner sep=0pt,outer sep=0pt] at ([xshift=189.860pt,yshift=-125.180pt]current page.north west) {{\fontsize{9.960pt}{10.558pt}\selectfont \bfseries Accurac}};
\node[anchor=base west,inner sep=0pt,outer sep=0pt] at ([xshift=189.860pt,yshift=-136.700pt]current page.north west) {{\fontsize{9.960pt}{10.558pt}\selectfont \bfseries y}};
\node[anchor=base west,inner sep=0pt,outer sep=0pt] at ([xshift=240.890pt,yshift=-125.180pt]current page.north west) {{\fontsize{9.960pt}{10.558pt}\selectfont \bfseries Precisio}};
\node[anchor=base west,inner sep=0pt,outer sep=0pt] at ([xshift=240.890pt,yshift=-136.700pt]current page.north west) {{\fontsize{9.960pt}{10.558pt}\selectfont \bfseries n}};
\node[anchor=base west,inner sep=0pt,outer sep=0pt] at ([xshift=290.690pt,yshift=-125.180pt]current page.north west) {{\fontsize{9.960pt}{10.558pt}\selectfont \bfseries Recall}};
\node[anchor=base west,inner sep=0pt,outer sep=0pt] at ([xshift=334.030pt,yshift=-125.180pt]current page.north west) {{\fontsize{9.960pt}{10.558pt}\selectfont \bfseries F1-Score}};
\node[anchor=base west,inner sep=0pt,outer sep=0pt] at ([xshift=383.710pt,yshift=-125.180pt]current page.north west) {{\fontsize{9.960pt}{10.558pt}\selectfont \bfseries ROC-}};
\node[anchor=base west,inner sep=0pt,outer sep=0pt] at ([xshift=383.710pt,yshift=-136.700pt]current page.north west) {{\fontsize{9.960pt}{10.558pt}\selectfont \bfseries AUC}};
\node[anchor=base west,inner sep=0pt,outer sep=0pt] at ([xshift=441.670pt,yshift=-125.180pt]current page.north west) {{\fontsize{9.960pt}{10.558pt}\selectfont \bfseries PR-AUC}};
\node[anchor=base west,inner sep=0pt,outer sep=0pt] at ([xshift=106.580pt,yshift=-148.700pt]current page.north west) {{\fontsize{9.960pt}{10.558pt}\selectfont  Logistic}};
\node[anchor=base west,inner sep=0pt,outer sep=0pt] at ([xshift=106.580pt,yshift=-160.100pt]current page.north west) {{\fontsize{9.960pt}{10.558pt}\selectfont  Regression}};
\node[anchor=base west,inner sep=0pt,outer sep=0pt] at ([xshift=189.860pt,yshift=-148.700pt]current page.north west) {{\fontsize{9.960pt}{10.558pt}\selectfont  0.9441}};
\node[anchor=base west,inner sep=0pt,outer sep=0pt] at ([xshift=240.890pt,yshift=-148.700pt]current page.north west) {{\fontsize{9.960pt}{10.558pt}\selectfont  0.0736}};
\node[anchor=base west,inner sep=0pt,outer sep=0pt] at ([xshift=290.690pt,yshift=-148.700pt]current page.north west) {{\fontsize{9.960pt}{10.558pt}\selectfont  0.7535}};
\node[anchor=base west,inner sep=0pt,outer sep=0pt] at ([xshift=334.030pt,yshift=-148.700pt]current page.north west) {{\fontsize{9.960pt}{10.558pt}\selectfont  0.1342}};
\node[anchor=base west,inner sep=0pt,outer sep=0pt] at ([xshift=383.710pt,yshift=-148.700pt]current page.north west) {{\fontsize{9.960pt}{10.558pt}\selectfont  0.8776}};
\node[anchor=base west,inner sep=0pt,outer sep=0pt] at ([xshift=441.670pt,yshift=-148.700pt]current page.north west) {{\fontsize{9.960pt}{10.558pt}\selectfont  0.1938}};
\node[anchor=base west,inner sep=0pt,outer sep=0pt] at ([xshift=106.580pt,yshift=-172.100pt]current page.north west) {{\fontsize{9.960pt}{10.558pt}\selectfont  Random Forest}};
\node[anchor=base west,inner sep=0pt,outer sep=0pt] at ([xshift=189.860pt,yshift=-172.100pt]current page.north west) {{\fontsize{9.960pt}{10.558pt}\selectfont  0.9972}};
\node[anchor=base west,inner sep=0pt,outer sep=0pt] at ([xshift=240.890pt,yshift=-172.100pt]current page.north west) {{\fontsize{9.960pt}{10.558pt}\selectfont  0.7221}};
\node[anchor=base west,inner sep=0pt,outer sep=0pt] at ([xshift=290.690pt,yshift=-172.100pt]current page.north west) {{\fontsize{9.960pt}{10.558pt}\selectfont  0.8221}};
\node[anchor=base west,inner sep=0pt,outer sep=0pt] at ([xshift=334.030pt,yshift=-172.100pt]current page.north west) {{\fontsize{9.960pt}{10.558pt}\selectfont  0.7688}};
\node[anchor=base west,inner sep=0pt,outer sep=0pt] at ([xshift=383.710pt,yshift=-172.100pt]current page.north west) {{\fontsize{9.960pt}{10.558pt}\selectfont  0.9946}};
\node[anchor=base west,inner sep=0pt,outer sep=0pt] at ([xshift=441.670pt,yshift=-172.100pt]current page.north west) {{\fontsize{9.960pt}{10.558pt}\selectfont  0.8494}};
\node[anchor=base west,inner sep=0pt,outer sep=0pt] at ([xshift=106.580pt,yshift=-184.100pt]current page.north west) {{\fontsize{9.960pt}{10.558pt}\selectfont  XGBoost}};
\node[anchor=base west,inner sep=0pt,outer sep=0pt] at ([xshift=189.860pt,yshift=-184.100pt]current page.north west) {{\fontsize{9.960pt}{10.558pt}\selectfont  0.9970}};
\node[anchor=base west,inner sep=0pt,outer sep=0pt] at ([xshift=240.890pt,yshift=-184.100pt]current page.north west) {{\fontsize{9.960pt}{10.558pt}\selectfont  0.6698}};
\node[anchor=base west,inner sep=0pt,outer sep=0pt] at ([xshift=290.690pt,yshift=-184.100pt]current page.north west) {{\fontsize{9.960pt}{10.558pt}\selectfont  0.9514}};
\node[anchor=base west,inner sep=0pt,outer sep=0pt] at ([xshift=334.030pt,yshift=-184.100pt]current page.north west) {{\fontsize{9.960pt}{10.558pt}\selectfont  0.7861}};
\node[anchor=base west,inner sep=0pt,outer sep=0pt] at ([xshift=383.710pt,yshift=-184.100pt]current page.north west) {{\fontsize{9.960pt}{10.558pt}\selectfont  0.9991}};
\node[anchor=base west,inner sep=0pt,outer sep=0pt] at ([xshift=441.670pt,yshift=-184.100pt]current page.north west) {{\fontsize{9.960pt}{10.558pt}\selectfont \bfseries 0.9459}};
\node[anchor=base west,inner sep=0pt,outer sep=0pt] at ([xshift=106.580pt,yshift=-196.100pt]current page.north west) {{\fontsize{9.960pt}{10.558pt}\selectfont  LightGBM}};
\node[anchor=base west,inner sep=0pt,outer sep=0pt] at ([xshift=189.860pt,yshift=-196.100pt]current page.north west) {{\fontsize{9.960pt}{10.558pt}\selectfont  0.9917}};
\node[anchor=base west,inner sep=0pt,outer sep=0pt] at ([xshift=240.890pt,yshift=-196.100pt]current page.north west) {{\fontsize{9.960pt}{10.558pt}\selectfont  0.4082}};
\node[anchor=base west,inner sep=0pt,outer sep=0pt] at ([xshift=290.690pt,yshift=-196.100pt]current page.north west) {{\fontsize{9.960pt}{10.558pt}\selectfont  0.9773}};
\node[anchor=base west,inner sep=0pt,outer sep=0pt] at ([xshift=334.030pt,yshift=-196.100pt]current page.north west) {{\fontsize{9.960pt}{10.558pt}\selectfont  0.5759}};
\node[anchor=base west,inner sep=0pt,outer sep=0pt] at ([xshift=383.710pt,yshift=-196.100pt]current page.north west) {{\fontsize{9.960pt}{10.558pt}\selectfont  0.9991}};
\node[anchor=base west,inner sep=0pt,outer sep=0pt] at ([xshift=441.670pt,yshift=-196.100pt]current page.north west) {{\fontsize{9.960pt}{10.558pt}\selectfont  0.9376}};
\node[anchor=base west,inner sep=0pt,outer sep=0pt] at ([xshift=82.104pt,yshift=-224.060pt]current page.north west) {{\fontsize{11.040pt}{11.702pt}\selectfont  \makebox[433.81pt][s]{As shown in Table 2, the ensemble-based classifiers substantially outperformed Logistic}}};
\node[anchor=base west,inner sep=0pt,outer sep=0pt] at ([xshift=82.104pt,yshift=-236.780pt]current page.north west) {{\fontsize{11.040pt}{11.702pt}\selectfont  \makebox[433.92pt][s]{Regression. XGBoost provided the strongest overall balance, achieving an F1-score of 0.7861,}}};
\node[anchor=base west,inner sep=0pt,outer sep=0pt] at ([xshift=82.104pt,yshift=-249.380pt]current page.north west) {{\fontsize{11.040pt}{11.702pt}\selectfont  \makebox[433.80pt][s]{ROC-AUC of 0.9991, and the highest PR-AUC of 0.9459. Random Forest achieved the highest}}};
\node[anchor=base west,inner sep=0pt,outer sep=0pt] at ([xshift=82.104pt,yshift=-262.100pt]current page.north west) {{\fontsize{11.040pt}{11.702pt}\selectfont  \makebox[433.65pt][s]{Precision (0.7221), while LightGBM produced the highest Recall (0.9773) but substantially}}};
\node[anchor=base west,inner sep=0pt,outer sep=0pt] at ([xshift=82.104pt,yshift=-274.700pt]current page.north west) {{\fontsize{11.040pt}{11.702pt}\selectfont  \makebox[433.61pt][s]{lower Precision. Based on its balanced performance, XGBoost was selected as the primary}}};
\node[anchor=base west,inner sep=0pt,outer sep=0pt] at ([xshift=82.104pt,yshift=-287.330pt]current page.north west) {{\fontsize{11.040pt}{11.702pt}\selectfont  \makebox[281.09pt][s]{classifier for the subsequent agent-aware ACAGS experiments.}}};
\node[anchor=base west,inner sep=0pt,outer sep=0pt] at ([xshift=174.020pt,yshift=-498.790pt]current page.north west) {{\fontsize{9.960pt}{10.558pt}\selectfont  \makebox[249.74pt][s]{Figure 4. Precision--Recall curves for the evaluated classifiers}}};
\node[anchor=base west,inner sep=0pt,outer sep=0pt] at ([xshift=82.104pt,yshift=-523.870pt]current page.north west) {{\fontsize{11.040pt}{11.702pt}\selectfont  \makebox[433.87pt][s]{As illustrated in Figure 4, XGBoost achieved the strongest precision--recall performance among}}};
\node[anchor=base west,inner sep=0pt,outer sep=0pt] at ([xshift=82.104pt,yshift=-536.470pt]current page.north west) {{\fontsize{11.040pt}{11.702pt}\selectfont  \makebox[433.86pt][s]{the evaluated classifiers, consistent with its highest PR-AUC of 0.9459. This result is particularly}}};
\node[anchor=base west,inner sep=0pt,outer sep=0pt] at ([xshift=82.104pt,yshift=-549.190pt]current page.north west) {{\fontsize{11.040pt}{11.702pt}\selectfont  \makebox[315.65pt][s]{relevant given the substantial class imbalance in the evaluation dataset.}}};
\node[anchor=north west,inner sep=0pt,outer sep=0pt] at ([xshift=179.000pt,yshift=-300.000pt]current page.north west) {\includegraphics[width=238.000pt,height=180.000pt]{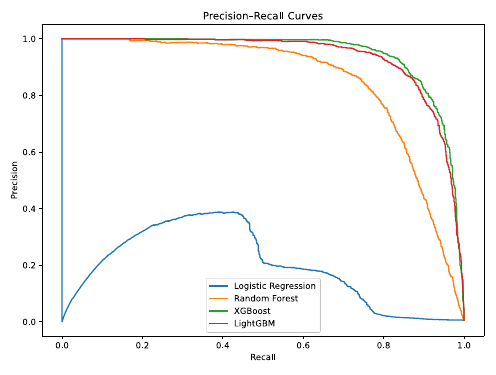}};
\end{tikzpicture}
\null
\newpage
\begin{tikzpicture}[remember picture,overlay]
\node[anchor=base west,inner sep=0pt,outer sep=0pt] at ([xshift=127.580pt,yshift=-81.360pt]current page.north west) {{\fontsize{9.960pt}{10.558pt}\selectfont  \makebox[342.77pt][s]{International Journal on Cybernetics \& Informatics (IJCI) Vol.15, No.4, August 2026}}};
\node[anchor=base west,inner sep=0pt,outer sep=0pt] at ([xshift=503.380pt,yshift=-767.856pt]current page.north west) {{\fontsize{9.960pt}{10.558pt}\selectfont  96}};
\node[anchor=base west,inner sep=0pt,outer sep=0pt] at ([xshift=173.540pt,yshift=-306.410pt]current page.north west) {{\fontsize{9.960pt}{10.558pt}\selectfont  \makebox[247.91pt][s]{Figure 5. Confusion matrix of the selected XGBoost classifier}}};
\node[anchor=base west,inner sep=0pt,outer sep=0pt] at ([xshift=82.104pt,yshift=-331.490pt]current page.north west) {{\fontsize{11.040pt}{11.702pt}\selectfont  \makebox[433.98pt][s]{Figure 5 demonstrates that the selected XGBoost classifier correctly identified 1,429 fraudulent}}};
\node[anchor=base west,inner sep=0pt,outer sep=0pt] at ([xshift=82.104pt,yshift=-344.090pt]current page.north west) {{\fontsize{11.040pt}{11.702pt}\selectfont  \makebox[433.73pt][s]{transactions, while 72 fraudulent transactions remained undetected. The model also correctly}}};
\node[anchor=base west,inner sep=0pt,outer sep=0pt] at ([xshift=82.104pt,yshift=-356.690pt]current page.north west) {{\fontsize{11.040pt}{11.702pt}\selectfont  \makebox[433.86pt][s]{classified 258,965 legitimate transactions, with 720 false-positive predictions. This balance is}}};
\node[anchor=base west,inner sep=0pt,outer sep=0pt] at ([xshift=82.104pt,yshift=-369.410pt]current page.north west) {{\fontsize{11.040pt}{11.702pt}\selectfont  \makebox[433.63pt][s]{particularly desirable in payment authorization systems, where reducing missed fraud while}}};
\node[anchor=base west,inner sep=0pt,outer sep=0pt] at ([xshift=82.104pt,yshift=-382.010pt]current page.north west) {{\fontsize{11.040pt}{11.702pt}\selectfont  \makebox[433.63pt][s]{limiting unnecessary false alerts is important for maintaining both security and customer}}};
\node[anchor=base west,inner sep=0pt,outer sep=0pt] at ([xshift=82.104pt,yshift=-394.730pt]current page.north west) {{\fontsize{11.040pt}{11.702pt}\selectfont  experience.}};
\node[anchor=base west,inner sep=0pt,outer sep=0pt] at ([xshift=82.104pt,yshift=-422.090pt]current page.north west) {{\fontsize{12.000pt}{12.720pt}\selectfont \bfseries 5.2.}};
\node[anchor=base west,inner sep=0pt,outer sep=0pt] at ([xshift=104.660pt,yshift=-422.090pt]current page.north west) {{\fontsize{12.000pt}{12.720pt}\selectfont \bfseries \makebox[199.76pt][s]{Explainability and Behavioral Analysis}}};
\node[anchor=base west,inner sep=0pt,outer sep=0pt] at ([xshift=82.104pt,yshift=-447.530pt]current page.north west) {{\fontsize{11.040pt}{11.702pt}\selectfont  \makebox[431.31pt][s]{SHAP analysis was applied to examine the contribution of transaction, behavioral, and agent-}}};
\node[anchor=base west,inner sep=0pt,outer sep=0pt] at ([xshift=82.104pt,yshift=-460.250pt]current page.north west) {{\fontsize{11.040pt}{11.702pt}\selectfont  \makebox[433.76pt][s]{specific features to XGBoost fraud-risk predictions. Transaction amount remained the dominant}}};
\node[anchor=base west,inner sep=0pt,outer sep=0pt] at ([xshift=82.104pt,yshift=-472.870pt]current page.north west) {{\fontsize{11.040pt}{11.702pt}\selectfont  \makebox[433.88pt][s]{predictor, while temporal and historical behavioral characteristics---including nighttime activity,}}};
\node[anchor=base west,inner sep=0pt,outer sep=0pt] at ([xshift=82.104pt,yshift=-485.590pt]current page.north west) {{\fontsize{11.040pt}{11.702pt}\selectfont  \makebox[431.48pt][s]{prior transaction count, time since the previous transaction, and deviation from prior spending---}}};
\node[anchor=base west,inner sep=0pt,outer sep=0pt] at ([xshift=82.104pt,yshift=-498.190pt]current page.north west) {{\fontsize{11.040pt}{11.702pt}\selectfont  \makebox[433.80pt][s]{also influenced model output. Agent-specific variables, including delegated spending limit, agent}}};
\node[anchor=base west,inner sep=0pt,outer sep=0pt] at ([xshift=82.104pt,yshift=-510.910pt]current page.north west) {{\fontsize{11.040pt}{11.702pt}\selectfont  \makebox[433.77pt][s]{trust score, transaction-to-agent-limit ratio, and agent identity score, provided additional}}};
\node[anchor=base west,inner sep=0pt,outer sep=0pt] at ([xshift=82.104pt,yshift=-523.510pt]current page.north west) {{\fontsize{11.040pt}{11.702pt}\selectfont  \makebox[433.62pt][s]{information in the extended model. These results demonstrate that ACAGS combines}}};
\node[anchor=base west,inner sep=0pt,outer sep=0pt] at ([xshift=82.104pt,yshift=-536.110pt]current page.north west) {{\fontsize{11.040pt}{11.702pt}\selectfont  \makebox[433.97pt][s]{conventional transaction characteristics with historical behavior and delegated-agent context}}};
\node[anchor=base west,inner sep=0pt,outer sep=0pt] at ([xshift=82.104pt,yshift=-548.830pt]current page.north west) {{\fontsize{11.040pt}{11.702pt}\selectfont  \makebox[123.15pt][s]{when estimating fraud risk.}}};
\node[anchor=north west,inner sep=0pt,outer sep=0pt] at ([xshift=180.000pt,yshift=-88.000pt]current page.north west) {\includegraphics[width=236.000pt,height=198.000pt]{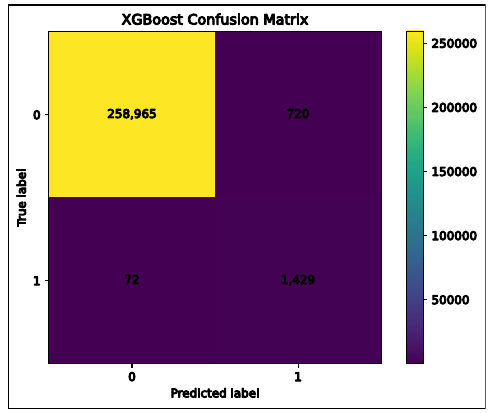}};
\end{tikzpicture}
\null
\newpage
\begin{tikzpicture}[remember picture,overlay]
\node[anchor=base west,inner sep=0pt,outer sep=0pt] at ([xshift=127.580pt,yshift=-81.360pt]current page.north west) {{\fontsize{9.960pt}{10.558pt}\selectfont  \makebox[342.77pt][s]{International Journal on Cybernetics \& Informatics (IJCI) Vol.15, No.4, August 2026}}};
\node[anchor=base west,inner sep=0pt,outer sep=0pt] at ([xshift=503.380pt,yshift=-767.856pt]current page.north west) {{\fontsize{9.960pt}{10.558pt}\selectfont  97}};
\node[anchor=base west,inner sep=0pt,outer sep=0pt] at ([xshift=134.660pt,yshift=-319.010pt]current page.north west) {{\fontsize{9.960pt}{10.558pt}\selectfont  \makebox[325.98pt][s]{Figure 6. SHAP analysis of fraud-risk predictions in the extended ACAGS model}}};
\node[anchor=base west,inner sep=0pt,outer sep=0pt] at ([xshift=82.104pt,yshift=-345.050pt]current page.north west) {{\fontsize{12.000pt}{12.720pt}\selectfont \bfseries 5.3.}};
\node[anchor=base west,inner sep=0pt,outer sep=0pt] at ([xshift=104.660pt,yshift=-345.050pt]current page.north west) {{\fontsize{12.000pt}{12.720pt}\selectfont \bfseries \makebox[186.27pt][s]{Simulated Agent-Attack Evaluation}}};
\node[anchor=base west,inner sep=0pt,outer sep=0pt] at ([xshift=82.104pt,yshift=-370.490pt]current page.north west) {{\fontsize{11.040pt}{11.702pt}\selectfont  \makebox[433.82pt][s]{ACAGS was evaluated against the five simulated agent-specific attack scenarios defined in the}}};
\node[anchor=base west,inner sep=0pt,outer sep=0pt] at ([xshift=82.104pt,yshift=-383.090pt]current page.north west) {{\fontsize{11.040pt}{11.702pt}\selectfont  \makebox[433.91pt][s]{threat model. At the selected agent-risk threshold, the framework detected 99.77\% of abnormal}}};
\node[anchor=base west,inner sep=0pt,outer sep=0pt] at ([xshift=82.104pt,yshift=-395.810pt]current page.north west) {{\fontsize{11.040pt}{11.702pt}\selectfont  \makebox[433.63pt][s]{agent behavior, 99.73\% of agent impersonation, 100\% of malicious delegation, 100\% of scope}}};
\node[anchor=base west,inner sep=0pt,outer sep=0pt] at ([xshift=82.104pt,yshift=-408.410pt]current page.north west) {{\fontsize{11.040pt}{11.702pt}\selectfont  \makebox[433.85pt][s]{violations, and 72.27\% of compromised-agent attacks.The comparatively lower detection rate for}}};
\node[anchor=base west,inner sep=0pt,outer sep=0pt] at ([xshift=82.104pt,yshift=-421.130pt]current page.north west) {{\fontsize{11.040pt}{11.702pt}\selectfont  \makebox[433.68pt][s]{compromised agents indicates that malicious activity conducted through otherwise legitimate}}};
\node[anchor=base west,inner sep=0pt,outer sep=0pt] at ([xshift=82.104pt,yshift=-433.730pt]current page.north west) {{\fontsize{11.040pt}{11.702pt}\selectfont  \makebox[433.60pt][s]{agent credentials presents a more difficult detection problem than malicious delegation or scope}}};
\node[anchor=base west,inner sep=0pt,outer sep=0pt] at ([xshift=82.104pt,yshift=-446.330pt]current page.north west) {{\fontsize{11.040pt}{11.702pt}\selectfont  violations.}};
\node[anchor=base west,inner sep=0pt,outer sep=0pt] at ([xshift=162.980pt,yshift=-657.550pt]current page.north west) {{\fontsize{9.960pt}{10.558pt}\selectfont  \makebox[269.18pt][s]{Figure 7. ACAGS detection rate by simulated agent attack scenario}}};
\node[anchor=base west,inner sep=0pt,outer sep=0pt] at ([xshift=82.104pt,yshift=-683.500pt]current page.north west) {{\fontsize{12.000pt}{12.720pt}\selectfont \bfseries 5.4.}};
\node[anchor=base west,inner sep=0pt,outer sep=0pt] at ([xshift=104.660pt,yshift=-683.500pt]current page.north west) {{\fontsize{12.000pt}{12.720pt}\selectfont \bfseries \makebox[179.43pt][s]{Ablation and Statistical Validation}}};
\node[anchor=base west,inner sep=0pt,outer sep=0pt] at ([xshift=82.104pt,yshift=-708.940pt]current page.north west) {{\fontsize{11.040pt}{11.702pt}\selectfont  \makebox[433.99pt][s]{The ablation analysis evaluated the incremental contribution of behavioral and agent-specific}}};
\node[anchor=base west,inner sep=0pt,outer sep=0pt] at ([xshift=82.104pt,yshift=-721.660pt]current page.north west) {{\fontsize{11.040pt}{11.702pt}\selectfont  \makebox[433.97pt][s]{features across 10 paired experimental runs. Mean F1-score increased from 0.5212 using base}}};
\node[anchor=base west,inner sep=0pt,outer sep=0pt] at ([xshift=82.104pt,yshift=-734.260pt]current page.north west) {{\fontsize{11.040pt}{11.702pt}\selectfont  \makebox[431.43pt][s]{transaction features to 0.6749 with behavioral features and 0.6946 after incorporating agent-}}};
\node[anchor=base west,inner sep=0pt,outer sep=0pt] at ([xshift=82.104pt,yshift=-746.976pt]current page.north west) {{\fontsize{11.040pt}{11.702pt}\selectfont  \makebox[434.00pt][s]{specific features. Mean PR-AUC similarly increased from 0.5282to 0.7455 and 0.7723,}}};
\node[anchor=north west,inner sep=0pt,outer sep=0pt] at ([xshift=156.000pt,yshift=-87.000pt]current page.north west) {\includegraphics[width=284.000pt,height=213.000pt]{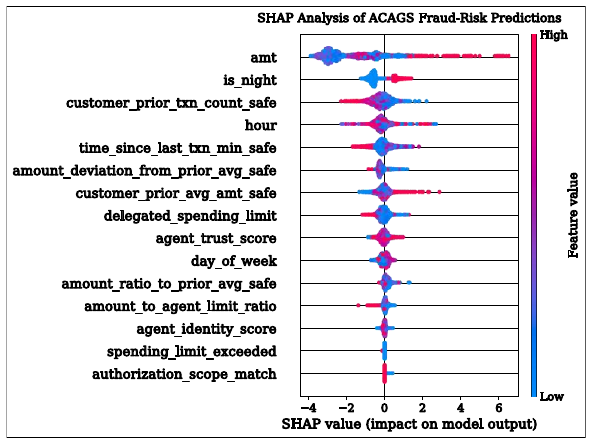}};
\node[anchor=north west,inner sep=0pt,outer sep=0pt] at ([xshift=156.000pt,yshift=-458.000pt]current page.north west) {\includegraphics[width=285.000pt,height=180.000pt]{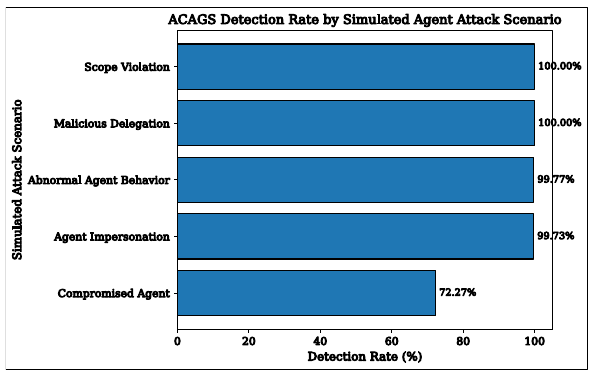}};
\end{tikzpicture}
\null
\newpage
\begin{tikzpicture}[remember picture,overlay]
\node[anchor=base west,inner sep=0pt,outer sep=0pt] at ([xshift=127.580pt,yshift=-81.360pt]current page.north west) {{\fontsize{9.960pt}{10.558pt}\selectfont  \makebox[342.77pt][s]{International Journal on Cybernetics \& Informatics (IJCI) Vol.15, No.4, August 2026}}};
\node[anchor=base west,inner sep=0pt,outer sep=0pt] at ([xshift=503.380pt,yshift=-767.856pt]current page.north west) {{\fontsize{9.960pt}{10.558pt}\selectfont  98}};
\node[anchor=base west,inner sep=0pt,outer sep=0pt] at ([xshift=82.104pt,yshift=-100.220pt]current page.north west) {{\fontsize{11.040pt}{11.702pt}\selectfont  \makebox[433.68pt][s]{respectively. Paired t-tests showed statistically significant improvements from base to behavioral}}};
\node[anchor=base west,inner sep=0pt,outer sep=0pt] at ([xshift=82.104pt,yshift=-112.820pt]current page.north west) {{\fontsize{11.040pt}{11.702pt}\selectfont  \makebox[433.97pt][s]{features and from behavioral to behavioral-plus-agent features across F1-score, ROC-AUC, and}}};
\node[anchor=base west,inner sep=0pt,outer sep=0pt] at ([xshift=82.104pt,yshift=-125.540pt]current page.north west) {{\fontsize{11.040pt}{11.702pt}\selectfont  \makebox[431.43pt][s]{PR-AUC (p < 0.001). Wilcoxon signed-rank tests also produced p < 0.05, providing non-}}};
\node[anchor=base west,inner sep=0pt,outer sep=0pt] at ([xshift=82.104pt,yshift=-138.140pt]current page.north west) {{\fontsize{11.040pt}{11.702pt}\selectfont  \makebox[228.15pt][s]{parametric support for the observed improvements.}}};
\node[anchor=base west,inner sep=0pt,outer sep=0pt] at ([xshift=160.820pt,yshift=-342.650pt]current page.north west) {{\fontsize{9.960pt}{10.558pt}\selectfont  \makebox[273.60pt][s]{Figure 8. Ablation analysis of behavioral and agent-specific features}}};
\node[anchor=base west,inner sep=0pt,outer sep=0pt] at ([xshift=82.104pt,yshift=-368.690pt]current page.north west) {{\fontsize{12.000pt}{12.720pt}\selectfont \bfseries 5.5.}};
\node[anchor=base west,inner sep=0pt,outer sep=0pt] at ([xshift=104.660pt,yshift=-368.690pt]current page.north west) {{\fontsize{12.000pt}{12.720pt}\selectfont \bfseries \makebox[261.17pt][s]{ACAGS Authorization and Operational Outcomes}}};
\node[anchor=base west,inner sep=0pt,outer sep=0pt] at ([xshift=82.104pt,yshift=-394.130pt]current page.north west) {{\fontsize{11.040pt}{11.702pt}\selectfont  \makebox[433.88pt][s]{The final ACAGS policy translated fraud-risk estimates and agent-specific authorization controls}}};
\node[anchor=base west,inner sep=0pt,outer sep=0pt] at ([xshift=82.104pt,yshift=-406.730pt]current page.north west) {{\fontsize{11.040pt}{11.702pt}\selectfont  \makebox[433.82pt][s]{into APPROVE, REVIEW, or REJECT outcomes. Scope violations and malicious delegation}}};
\node[anchor=base west,inner sep=0pt,outer sep=0pt] at ([xshift=82.104pt,yshift=-419.450pt]current page.north west) {{\fontsize{11.040pt}{11.702pt}\selectfont  \makebox[433.70pt][s]{were rejected in 100\% of simulated cases, while abnormal agent behavior was predominantly}}};
\node[anchor=base west,inner sep=0pt,outer sep=0pt] at ([xshift=82.104pt,yshift=-432.050pt]current page.north west) {{\fontsize{11.040pt}{11.702pt}\selectfont  \makebox[433.92pt][s]{routed to review (97.79\%). Agent impersonation was primarily reviewed (83.46\%) or rejected}}};
\node[anchor=base west,inner sep=0pt,outer sep=0pt] at ([xshift=82.104pt,yshift=-444.770pt]current page.north west) {{\fontsize{11.040pt}{11.702pt}\selectfont  \makebox[433.70pt][s]{(16.15\%), whereas compromised-agent transactions were more difficult to resolve automatically,}}};
\node[anchor=base west,inner sep=0pt,outer sep=0pt] at ([xshift=82.104pt,yshift=-457.370pt]current page.north west) {{\fontsize{11.040pt}{11.702pt}\selectfont  \makebox[433.82pt][s]{with 72.52\% routed to review and 27.29\% approved. These results demonstrate that explicit}}};
\node[anchor=base west,inner sep=0pt,outer sep=0pt] at ([xshift=82.104pt,yshift=-470.090pt]current page.north west) {{\fontsize{11.040pt}{11.702pt}\selectfont  \makebox[433.56pt][s]{authorization violations can be rejected directly, while behaviorally uncertain activity can be}}};
\node[anchor=base west,inner sep=0pt,outer sep=0pt] at ([xshift=82.104pt,yshift=-482.710pt]current page.north west) {{\fontsize{11.040pt}{11.702pt}\selectfont  \makebox[140.91pt][s]{escalated for additional review.}}};
\node[anchor=base west,inner sep=0pt,outer sep=0pt] at ([xshift=143.300pt,yshift=-695.620pt]current page.north west) {{\fontsize{9.960pt}{10.558pt}\selectfont  \makebox[308.53pt][s]{Figure 9. ACAGS authorization outcomes by simulated agent attack scenario}}};
\node[anchor=base west,inner sep=0pt,outer sep=0pt] at ([xshift=82.104pt,yshift=-720.580pt]current page.north west) {{\fontsize{11.040pt}{11.702pt}\selectfont  \makebox[297.00pt][s]{Across 259,335evaluated transactions, ACAGS approved 98.599\%}}};
\node[anchor=base west,inner sep=0pt,outer sep=0pt] at ([xshift=379.150pt,yshift=-720.580pt]current page.north west) {{\fontsize{11.040pt}{11.702pt}\selectfont \bfseries ,}};
\node[anchor=base west,inner sep=0pt,outer sep=0pt] at ([xshift=381.910pt,yshift=-720.580pt]current page.north west) {{\fontsize{11.040pt}{11.702pt}\selectfont  \makebox[134.15pt][s]{routed 0.731\%to review, and}}};
\node[anchor=base west,inner sep=0pt,outer sep=0pt] at ([xshift=82.104pt,yshift=-733.300pt]current page.north west) {{\fontsize{11.040pt}{11.702pt}\selectfont  \makebox[72.55pt][s]{rejected 0.669\%}}};
\node[anchor=base west,inner sep=0pt,outer sep=0pt] at ([xshift=154.700pt,yshift=-733.300pt]current page.north west) {{\fontsize{11.040pt}{11.702pt}\selectfont \bfseries .}};
\node[anchor=base west,inner sep=0pt,outer sep=0pt] at ([xshift=157.460pt,yshift=-733.300pt]current page.north west) {{\fontsize{11.040pt}{11.702pt}\selectfont  \makebox[142.10pt][s]{Among fraudulent transactions}}};
\node[anchor=base west,inner sep=0pt,outer sep=0pt] at ([xshift=299.690pt,yshift=-733.300pt]current page.north west) {{\fontsize{11.040pt}{11.702pt}\selectfont \bfseries ,}};
\node[anchor=base west,inner sep=0pt,outer sep=0pt] at ([xshift=306.170pt,yshift=-733.300pt]current page.north west) {{\fontsize{11.040pt}{11.702pt}\selectfont  \makebox[209.79pt][s]{65.09\% were escalated to review or rejection,}}};
\node[anchor=base west,inner sep=0pt,outer sep=0pt] at ([xshift=82.104pt,yshift=-745.896pt]current page.north west) {{\fontsize{11.040pt}{11.702pt}\selectfont  \makebox[433.69pt][s]{while 34.91\% were automatically approved. In comparison, the escalation rate for legitimate}}};
\node[anchor=north west,inner sep=0pt,outer sep=0pt] at ([xshift=155.000pt,yshift=-150.000pt]current page.north west) {\includegraphics[width=287.000pt,height=174.000pt]{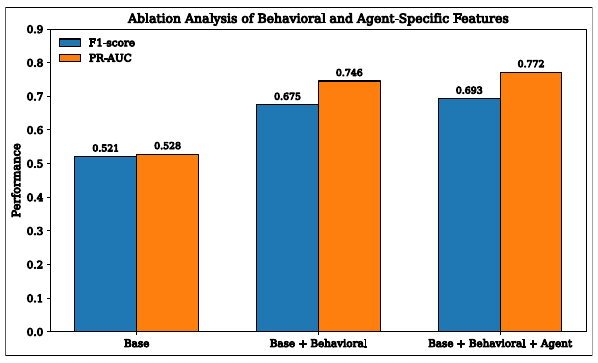}};
\node[anchor=north west,inner sep=0pt,outer sep=0pt] at ([xshift=143.000pt,yshift=-495.000pt]current page.north west) {\includegraphics[width=311.000pt,height=183.000pt]{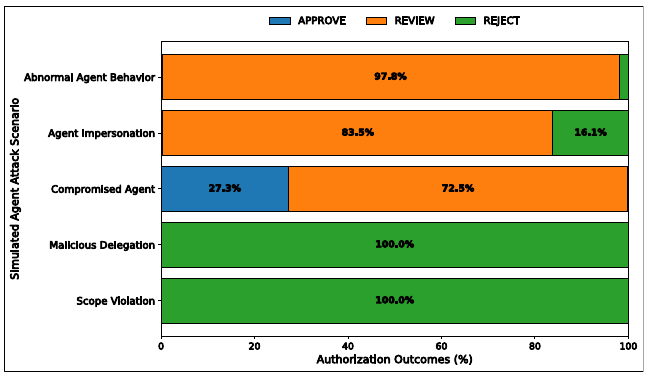}};
\end{tikzpicture}
\null
\newpage
\begin{tikzpicture}[remember picture,overlay]
\node[anchor=base west,inner sep=0pt,outer sep=0pt] at ([xshift=127.580pt,yshift=-81.360pt]current page.north west) {{\fontsize{9.960pt}{10.558pt}\selectfont  \makebox[342.77pt][s]{International Journal on Cybernetics \& Informatics (IJCI) Vol.15, No.4, August 2026}}};
\node[anchor=base west,inner sep=0pt,outer sep=0pt] at ([xshift=503.380pt,yshift=-767.856pt]current page.north west) {{\fontsize{9.960pt}{10.558pt}\selectfont  99}};
\node[anchor=base west,inner sep=0pt,outer sep=0pt] at ([xshift=82.104pt,yshift=-100.220pt]current page.north west) {{\fontsize{11.040pt}{11.702pt}\selectfont  \makebox[433.69pt][s]{transactions was 1.03\%, indicating a relatively low operational burden on legitimate activity. The}}};
\node[anchor=base west,inner sep=0pt,outer sep=0pt] at ([xshift=82.104pt,yshift=-112.820pt]current page.north west) {{\fontsize{11.040pt}{11.702pt}\selectfont  \makebox[433.99pt][s]{total value of fraudulent transactions was \$812,865.18, of which \$671,908.03 (82.66\%) was}}};
\node[anchor=base west,inner sep=0pt,outer sep=0pt] at ([xshift=82.104pt,yshift=-125.540pt]current page.north west) {{\fontsize{11.040pt}{11.702pt}\selectfont  \makebox[431.43pt][s]{associated with transactions escalated for review or rejection. These values represent transaction-}}};
\node[anchor=base west,inner sep=0pt,outer sep=0pt] at ([xshift=82.104pt,yshift=-138.140pt]current page.north west) {{\fontsize{11.040pt}{11.702pt}\selectfont  \makebox[433.71pt][s]{value exposure under the experimental authorization policy and should not be interpreted as}}};
\node[anchor=base west,inner sep=0pt,outer sep=0pt] at ([xshift=82.104pt,yshift=-150.860pt]current page.north west) {{\fontsize{11.040pt}{11.702pt}\selectfont  \makebox[168.63pt][s]{realized or prevented financial losses.}}};
\node[anchor=base west,inner sep=0pt,outer sep=0pt] at ([xshift=82.104pt,yshift=-178.940pt]current page.north west) {{\fontsize{14.040pt}{14.882pt}\selectfont \bfseries 6.}};
\node[anchor=base west,inner sep=0pt,outer sep=0pt] at ([xshift=100.100pt,yshift=-178.940pt]current page.north west) {{\fontsize{14.040pt}{14.882pt}\selectfont \bfseries D}};
\node[anchor=base west,inner sep=0pt,outer sep=0pt] at ([xshift=110.300pt,yshift=-178.940pt]current page.north west) {{\fontsize{11.040pt}{11.702pt}\selectfont \bfseries ISCUSSION}};
\node[anchor=base west,inner sep=0pt,outer sep=0pt] at ([xshift=82.104pt,yshift=-204.860pt]current page.north west) {{\fontsize{11.040pt}{11.702pt}\selectfont  \makebox[433.88pt][s]{The experimental findings demonstrate that behavioral and agent-specific information can}}};
\node[anchor=base west,inner sep=0pt,outer sep=0pt] at ([xshift=82.104pt,yshift=-217.460pt]current page.north west) {{\fontsize{11.040pt}{11.702pt}\selectfont  \makebox[433.71pt][s]{strengthen fraud-risk assessment within the evaluated ACAGS framework. The baseline}}};
\node[anchor=base west,inner sep=0pt,outer sep=0pt] at ([xshift=82.104pt,yshift=-230.180pt]current page.north west) {{\fontsize{11.040pt}{11.702pt}\selectfont  \makebox[433.65pt][s]{experiments establish performance on conventional payment transactions, while the simulated}}};
\node[anchor=base west,inner sep=0pt,outer sep=0pt] at ([xshift=82.104pt,yshift=-242.780pt]current page.north west) {{\fontsize{11.040pt}{11.702pt}\selectfont  \makebox[433.73pt][s]{agentic-commerce experiments provide controlled evidence across agent impersonation,}}};
\node[anchor=base west,inner sep=0pt,outer sep=0pt] at ([xshift=82.104pt,yshift=-255.500pt]current page.north west) {{\fontsize{11.040pt}{11.702pt}\selectfont  \makebox[433.70pt][s]{compromised-agent activity, malicious delegation, scope violations, and abnormal agent}}};
\node[anchor=base west,inner sep=0pt,outer sep=0pt] at ([xshift=82.104pt,yshift=-268.100pt]current page.north west) {{\fontsize{11.040pt}{11.702pt}\selectfont  \makebox[433.76pt][s]{behavior. The ablation and multi-run analyses further demonstrate measurable contributions from}}};
\node[anchor=base west,inner sep=0pt,outer sep=0pt] at ([xshift=82.104pt,yshift=-280.700pt]current page.north west) {{\fontsize{11.040pt}{11.702pt}\selectfont  \makebox[433.65pt][s]{behavioral and agent-specific features. These findings should be interpreted within the simulated}}};
\node[anchor=base west,inner sep=0pt,outer sep=0pt] at ([xshift=82.104pt,yshift=-293.450pt]current page.north west) {{\fontsize{11.040pt}{11.702pt}\selectfont  \makebox[431.43pt][s]{experimental setting and do not constitute evidence of performance in production agentic-}}};
\node[anchor=base west,inner sep=0pt,outer sep=0pt] at ([xshift=82.104pt,yshift=-306.050pt]current page.north west) {{\fontsize{11.040pt}{11.702pt}\selectfont  \makebox[114.03pt][s]{commerce environments.}}};
\node[anchor=base west,inner sep=0pt,outer sep=0pt] at ([xshift=82.104pt,yshift=-332.330pt]current page.north west) {{\fontsize{12.000pt}{12.720pt}\selectfont \bfseries 6.1.}};
\node[anchor=base west,inner sep=0pt,outer sep=0pt] at ([xshift=104.660pt,yshift=-332.330pt]current page.north west) {{\fontsize{12.000pt}{12.720pt}\selectfont \bfseries \makebox[129.51pt][s]{Interpretation of Results}}};
\node[anchor=base west,inner sep=0pt,outer sep=0pt] at ([xshift=82.104pt,yshift=-357.770pt]current page.north west) {{\fontsize{11.040pt}{11.702pt}\selectfont  \makebox[433.92pt][s]{The baseline results show that ensemble classifiers provide strong fraud-detection performance,}}};
\node[anchor=base west,inner sep=0pt,outer sep=0pt] at ([xshift=82.104pt,yshift=-370.490pt]current page.north west) {{\fontsize{11.040pt}{11.702pt}\selectfont  \makebox[433.63pt][s]{with XGBoost offering the most balanced performance across the evaluated metrics. The}}};
\node[anchor=base west,inner sep=0pt,outer sep=0pt] at ([xshift=82.104pt,yshift=-383.090pt]current page.north west) {{\fontsize{11.040pt}{11.702pt}\selectfont  \makebox[433.63pt][s]{extended experiments further show that behavioral features substantially improve predictive}}};
\node[anchor=base west,inner sep=0pt,outer sep=0pt] at ([xshift=82.104pt,yshift=-395.810pt]current page.north west) {{\fontsize{11.040pt}{11.702pt}\selectfont  \makebox[433.84pt][s]{performance over base transaction attributes, while agent-specific features provide additional}}};
\node[anchor=base west,inner sep=0pt,outer sep=0pt] at ([xshift=82.104pt,yshift=-408.410pt]current page.north west) {{\fontsize{11.040pt}{11.702pt}\selectfont  \makebox[433.86pt][s]{gains. Across 10 runs, the combined behavioral and agent-specific configuration achieved higher}}};
\node[anchor=base west,inner sep=0pt,outer sep=0pt] at ([xshift=82.104pt,yshift=-421.130pt]current page.north west) {{\fontsize{11.040pt}{11.702pt}\selectfont  \makebox[433.93pt][s]{mean F1-score, ROC-AUC, and PR-AUC than either the base or behavioral-only configurations,}}};
\node[anchor=base west,inner sep=0pt,outer sep=0pt] at ([xshift=82.104pt,yshift=-433.730pt]current page.north west) {{\fontsize{11.040pt}{11.702pt}\selectfont  \makebox[275.69pt][s]{with statistical testing supporting the observed improvements.}}};
\node[anchor=base west,inner sep=0pt,outer sep=0pt] at ([xshift=82.104pt,yshift=-459.050pt]current page.north west) {{\fontsize{11.040pt}{11.702pt}\selectfont  \makebox[433.90pt][s]{Attack-specific results also reveal differences in detection difficulty. Malicious delegation and}}};
\node[anchor=base west,inner sep=0pt,outer sep=0pt] at ([xshift=82.104pt,yshift=-471.670pt]current page.north west) {{\fontsize{11.040pt}{11.702pt}\selectfont  \makebox[433.61pt][s]{scope violations were detected in 100\% of simulated cases, while abnormal agent behavior and}}};
\node[anchor=base west,inner sep=0pt,outer sep=0pt] at ([xshift=82.104pt,yshift=-484.390pt]current page.north west) {{\fontsize{11.040pt}{11.702pt}\selectfont  \makebox[433.63pt][s]{agent impersonation achieved detection rates of 99.77\% and 99.73\%, respectively.}}};
\node[anchor=base west,inner sep=0pt,outer sep=0pt] at ([xshift=82.104pt,yshift=-496.990pt]current page.north west) {{\fontsize{11.040pt}{11.702pt}\selectfont  \makebox[402.67pt][s]{Compromised-agent attacks were more challenging, with a detection rate of 72.27\%}}};
\node[anchor=base west,inner sep=0pt,outer sep=0pt] at ([xshift=485.140pt,yshift=-496.990pt]current page.north west) {{\fontsize{11.040pt}{11.702pt}\selectfont \bfseries .}};
\node[anchor=base west,inner sep=0pt,outer sep=0pt] at ([xshift=487.900pt,yshift=-496.990pt]current page.north west) {{\fontsize{11.040pt}{11.702pt}\selectfont  This}};
\node[anchor=base west,inner sep=0pt,outer sep=0pt] at ([xshift=82.104pt,yshift=-509.590pt]current page.north west) {{\fontsize{11.040pt}{11.702pt}\selectfont  \makebox[433.77pt][s]{distinction suggests that agent-aware payment security should combine explicit authorization}}};
\node[anchor=base west,inner sep=0pt,outer sep=0pt] at ([xshift=82.104pt,yshift=-522.310pt]current page.north west) {{\fontsize{11.040pt}{11.702pt}\selectfont  \makebox[433.66pt][s]{controls with behavioral risk assessment, particularly when malicious activity occurs through}}};
\node[anchor=base west,inner sep=0pt,outer sep=0pt] at ([xshift=82.104pt,yshift=-534.910pt]current page.north west) {{\fontsize{11.040pt}{11.702pt}\selectfont  \makebox[150.03pt][s]{otherwise valid agent credentials.}}};
\node[anchor=base west,inner sep=0pt,outer sep=0pt] at ([xshift=82.104pt,yshift=-560.230pt]current page.north west) {{\fontsize{11.040pt}{11.702pt}\selectfont  \makebox[433.81pt][s]{From a practical perspective, the agent-aware design complements established authentication and}}};
\node[anchor=base west,inner sep=0pt,outer sep=0pt] at ([xshift=82.104pt,yshift=-572.830pt]current page.north west) {{\fontsize{11.040pt}{11.702pt}\selectfont  \makebox[433.91pt][s]{authorization mechanisms by introducing behavioral and delegated-authorization controls for}}};
\node[anchor=base west,inner sep=0pt,outer sep=0pt] at ([xshift=82.104pt,yshift=-585.550pt]current page.north west) {{\fontsize{11.040pt}{11.702pt}\selectfont  \makebox[433.80pt][s]{evaluating autonomous transactions. SHAP-based explanations further support interpretation of}}};
\node[anchor=base west,inner sep=0pt,outer sep=0pt] at ([xshift=82.104pt,yshift=-598.150pt]current page.north west) {{\fontsize{11.040pt}{11.702pt}\selectfont  \makebox[345.34pt][s]{fraud-risk predictions without directly determining authorization outcomes}}};
\node[anchor=base west,inner sep=0pt,outer sep=0pt] at ([xshift=427.750pt,yshift=-598.150pt]current page.north west) {{\fontsize{11.040pt}{11.702pt}\selectfont \bfseries .}};
\node[anchor=base west,inner sep=0pt,outer sep=0pt] at ([xshift=430.510pt,yshift=-598.150pt]current page.north west) {{\fontsize{11.040pt}{11.702pt}\selectfont  \makebox[85.52pt][s]{Such capabilities}}};
\node[anchor=base west,inner sep=0pt,outer sep=0pt] at ([xshift=82.104pt,yshift=-610.870pt]current page.north west) {{\fontsize{11.040pt}{11.702pt}\selectfont  \makebox[433.61pt][s]{may become increasingly relevant as AI agents assume greater responsibility for initiating and}}};
\node[anchor=base west,inner sep=0pt,outer sep=0pt] at ([xshift=82.104pt,yshift=-623.470pt]current page.north west) {{\fontsize{11.040pt}{11.702pt}\selectfont  \makebox[260.57pt][s]{executing commercial transactions on behalf of users [16].}}};
\node[anchor=base west,inner sep=0pt,outer sep=0pt] at ([xshift=82.104pt,yshift=-649.750pt]current page.north west) {{\fontsize{12.000pt}{12.720pt}\selectfont \bfseries 6.2.}};
\node[anchor=base west,inner sep=0pt,outer sep=0pt] at ([xshift=104.660pt,yshift=-649.750pt]current page.north west) {{\fontsize{12.000pt}{12.720pt}\selectfont \bfseries \makebox[311.21pt][s]{Practical Implications for Agentic-Commerce Cybersecurity}}};
\node[anchor=base west,inner sep=0pt,outer sep=0pt] at ([xshift=82.104pt,yshift=-675.220pt]current page.north west) {{\fontsize{11.040pt}{11.702pt}\selectfont  \makebox[433.57pt][s]{The results indicate that ACAGS can support a layered authorization approach in which fraud}}};
\node[anchor=base west,inner sep=0pt,outer sep=0pt] at ([xshift=82.104pt,yshift=-687.940pt]current page.north west) {{\fontsize{11.040pt}{11.702pt}\selectfont  \makebox[433.81pt][s]{probability, behavioral deviations, and agent-specific policy constraints contribute to transaction}}};
\node[anchor=base west,inner sep=0pt,outer sep=0pt] at ([xshift=82.104pt,yshift=-700.540pt]current page.north west) {{\fontsize{11.040pt}{11.702pt}\selectfont  \makebox[433.64pt][s]{screening. Rather than treating all suspicious activity uniformly, the framework distinguishes}}};
\node[anchor=base west,inner sep=0pt,outer sep=0pt] at ([xshift=82.104pt,yshift=-713.140pt]current page.north west) {{\fontsize{11.040pt}{11.702pt}\selectfont  \makebox[433.66pt][s]{explicit authorization violations from uncertain behavioral risk, allowing transactions to be}}};
\node[anchor=base west,inner sep=0pt,outer sep=0pt] at ([xshift=82.104pt,yshift=-725.860pt]current page.north west) {{\fontsize{11.040pt}{11.702pt}\selectfont  \makebox[372.44pt][s]{approved, escalated for review, or rejected according to the observed risk condition.}}};
\end{tikzpicture}
\null
\newpage
\begin{tikzpicture}[remember picture,overlay]
\node[anchor=base west,inner sep=0pt,outer sep=0pt] at ([xshift=127.580pt,yshift=-81.360pt]current page.north west) {{\fontsize{9.960pt}{10.558pt}\selectfont  \makebox[342.77pt][s]{International Journal on Cybernetics \& Informatics (IJCI) Vol.15, No.4, August 2026}}};
\node[anchor=base west,inner sep=0pt,outer sep=0pt] at ([xshift=498.340pt,yshift=-767.856pt]current page.north west) {{\fontsize{9.960pt}{10.558pt}\selectfont  100}};
\node[anchor=base west,inner sep=0pt,outer sep=0pt] at ([xshift=82.104pt,yshift=-100.220pt]current page.north west) {{\fontsize{11.040pt}{11.702pt}\selectfont  \makebox[433.87pt][s]{The low 1.03\% escalation rate for legitimate transactions suggests that this approach may limit}}};
\node[anchor=base west,inner sep=0pt,outer sep=0pt] at ([xshift=82.104pt,yshift=-112.820pt]current page.north west) {{\fontsize{11.040pt}{11.702pt}\selectfont  \makebox[433.62pt][s]{unnecessary operational intervention while retaining stronger controls for suspicious activity.}}};
\node[anchor=base west,inner sep=0pt,outer sep=0pt] at ([xshift=82.104pt,yshift=-125.540pt]current page.north west) {{\fontsize{11.040pt}{11.702pt}\selectfont  \makebox[433.97pt][s]{However, the 27.29\% approval ratefor simulated compromised-agent transactions demonstrates}}};
\node[anchor=base west,inner sep=0pt,outer sep=0pt] at ([xshift=82.104pt,yshift=-138.140pt]current page.north west) {{\fontsize{11.040pt}{11.702pt}\selectfont  \makebox[433.70pt][s]{that valid but compromised agent credentials remain an important challenge. Operational}}};
\node[anchor=base west,inner sep=0pt,outer sep=0pt] at ([xshift=82.104pt,yshift=-150.860pt]current page.north west) {{\fontsize{11.040pt}{11.702pt}\selectfont  \makebox[433.86pt][s]{deployment would therefore require institution-specific threshold calibration, continuous}}};
\node[anchor=base west,inner sep=0pt,outer sep=0pt] at ([xshift=82.104pt,yshift=-163.460pt]current page.north west) {{\fontsize{11.040pt}{11.702pt}\selectfont  \makebox[433.90pt][s]{monitoring, latency testing, governance controls, and validation against real-world transaction}}};
\node[anchor=base west,inner sep=0pt,outer sep=0pt] at ([xshift=82.104pt,yshift=-176.060pt]current page.north west) {{\fontsize{11.040pt}{11.702pt}\selectfont  patterns.}};
\node[anchor=base west,inner sep=0pt,outer sep=0pt] at ([xshift=82.104pt,yshift=-202.340pt]current page.north west) {{\fontsize{12.000pt}{12.720pt}\selectfont \bfseries 6.3.}};
\node[anchor=base west,inner sep=0pt,outer sep=0pt] at ([xshift=104.660pt,yshift=-202.340pt]current page.north west) {{\fontsize{12.000pt}{12.720pt}\selectfont \bfseries Limitations}};
\node[anchor=base west,inner sep=0pt,outer sep=0pt] at ([xshift=82.104pt,yshift=-227.900pt]current page.north west) {{\fontsize{11.040pt}{11.702pt}\selectfont  \makebox[433.81pt][s]{This study has several limitations. First, the agentic-commerce environment and attack scenarios}}};
\node[anchor=base west,inner sep=0pt,outer sep=0pt] at ([xshift=82.104pt,yshift=-240.500pt]current page.north west) {{\fontsize{11.040pt}{11.702pt}\selectfont  \makebox[433.91pt][s]{were simulated, rather than derived from production autonomous-payment systems. Although the}}};
\node[anchor=base west,inner sep=0pt,outer sep=0pt] at ([xshift=82.104pt,yshift=-253.100pt]current page.north west) {{\fontsize{11.040pt}{11.702pt}\selectfont  \makebox[433.78pt][s]{underlying transaction dataset provides a large-scale basis for fraud evaluation, the simulated}}};
\node[anchor=base west,inner sep=0pt,outer sep=0pt] at ([xshift=82.104pt,yshift=-265.820pt]current page.north west) {{\fontsize{11.040pt}{11.702pt}\selectfont  \makebox[433.62pt][s]{agent identities, delegated authorization attributes, and attack behaviors may not capture the full}}};
\node[anchor=base west,inner sep=0pt,outer sep=0pt] at ([xshift=82.104pt,yshift=-278.420pt]current page.north west) {{\fontsize{11.040pt}{11.702pt}\selectfont  \makebox[198.63pt][s]{complexity of real-world agentic commerce.}}};
\node[anchor=base west,inner sep=0pt,outer sep=0pt] at ([xshift=82.104pt,yshift=-303.770pt]current page.north west) {{\fontsize{11.040pt}{11.702pt}\selectfont  \makebox[433.68pt][s]{Second, the authorization thresholds and policy rules were evaluated under controlled}}};
\node[anchor=base west,inner sep=0pt,outer sep=0pt] at ([xshift=82.104pt,yshift=-316.490pt]current page.north west) {{\fontsize{11.040pt}{11.702pt}\selectfont  \makebox[433.85pt][s]{experimental conditions and may require institution-specific calibration in operational}}};
\node[anchor=base west,inner sep=0pt,outer sep=0pt] at ([xshift=82.104pt,yshift=-329.090pt]current page.north west) {{\fontsize{11.040pt}{11.702pt}\selectfont  \makebox[434.00pt][s]{environments. The study also does not evaluate production-level factors such as end-to-end}}};
\node[anchor=base west,inner sep=0pt,outer sep=0pt] at ([xshift=82.104pt,yshift=-341.690pt]current page.north west) {{\fontsize{11.040pt}{11.702pt}\selectfont  \makebox[433.66pt][s]{authorization latency, system scalability, evolving adversarial behavior, or integration with live}}};
\node[anchor=base west,inner sep=0pt,outer sep=0pt] at ([xshift=82.104pt,yshift=-354.410pt]current page.north west) {{\fontsize{11.040pt}{11.702pt}\selectfont  \makebox[433.97pt][s]{payment infrastructure. Future work should therefore validate ACAGS using real-world or}}};
\node[anchor=base west,inner sep=0pt,outer sep=0pt] at ([xshift=82.104pt,yshift=-367.010pt]current page.north west) {{\fontsize{11.040pt}{11.702pt}\selectfont  \makebox[433.67pt][s]{industry-derived agentic transaction data, evaluate adaptive attack strategies, and assess}}};
\node[anchor=base west,inner sep=0pt,outer sep=0pt] at ([xshift=82.104pt,yshift=-379.730pt]current page.north west) {{\fontsize{11.040pt}{11.702pt}\selectfont  \makebox[285.89pt][s]{performance under production-scale authorization requirements.}}};
\node[anchor=base west,inner sep=0pt,outer sep=0pt] at ([xshift=82.104pt,yshift=-407.810pt]current page.north west) {{\fontsize{14.040pt}{14.882pt}\selectfont \bfseries 7.}};
\node[anchor=base west,inner sep=0pt,outer sep=0pt] at ([xshift=100.100pt,yshift=-407.810pt]current page.north west) {{\fontsize{14.040pt}{14.882pt}\selectfont \bfseries C}};
\node[anchor=base west,inner sep=0pt,outer sep=0pt] at ([xshift=110.300pt,yshift=-407.810pt]current page.north west) {{\fontsize{11.040pt}{11.702pt}\selectfont \bfseries ONCLUSION AND}};
\node[anchor=base west,inner sep=0pt,outer sep=0pt] at ([xshift=205.610pt,yshift=-407.810pt]current page.north west) {{\fontsize{14.040pt}{14.882pt}\selectfont \bfseries F}};
\node[anchor=base west,inner sep=0pt,outer sep=0pt] at ([xshift=214.250pt,yshift=-407.810pt]current page.north west) {{\fontsize{11.040pt}{11.702pt}\selectfont \bfseries UTURE}};
\node[anchor=base west,inner sep=0pt,outer sep=0pt] at ([xshift=255.050pt,yshift=-407.810pt]current page.north west) {{\fontsize{14.040pt}{14.882pt}\selectfont \bfseries W}};
\node[anchor=base west,inner sep=0pt,outer sep=0pt] at ([xshift=269.090pt,yshift=-407.810pt]current page.north west) {{\fontsize{11.040pt}{11.702pt}\selectfont \bfseries ORK}};
\node[anchor=base west,inner sep=0pt,outer sep=0pt] at ([xshift=82.104pt,yshift=-433.730pt]current page.north west) {{\fontsize{11.040pt}{11.702pt}\selectfont  \makebox[433.90pt][s]{This study presented ACAGS, a cybersecurity framework for behavior-aware fraud detection and}}};
\node[anchor=base west,inner sep=0pt,outer sep=0pt] at ([xshift=82.104pt,yshift=-446.330pt]current page.north west) {{\fontsize{11.040pt}{11.702pt}\selectfont  \makebox[433.69pt][s]{trusted AI-agent payment authorization in agentic commerce. The framework integrates}}};
\node[anchor=base west,inner sep=0pt,outer sep=0pt] at ([xshift=82.104pt,yshift=-459.050pt]current page.north west) {{\fontsize{11.040pt}{11.702pt}\selectfont  \makebox[433.99pt][s]{behavioral feature engineering, machine-learning-based fraud assessment, agent-specific}}};
\node[anchor=base west,inner sep=0pt,outer sep=0pt] at ([xshift=82.104pt,yshift=-471.670pt]current page.north west) {{\fontsize{11.040pt}{11.702pt}\selectfont  \makebox[433.66pt][s]{authorization controls, and explainable AI to evaluate autonomous payment transactions. Among}}};
\node[anchor=base west,inner sep=0pt,outer sep=0pt] at ([xshift=82.104pt,yshift=-484.390pt]current page.north west) {{\fontsize{11.040pt}{11.702pt}\selectfont  \makebox[433.62pt][s]{the baseline classifiers, XGBoost achieved the strongest overall performance, while the extended}}};
\node[anchor=base west,inner sep=0pt,outer sep=0pt] at ([xshift=82.104pt,yshift=-496.990pt]current page.north west) {{\fontsize{11.040pt}{11.702pt}\selectfont  \makebox[433.67pt][s]{experiments demonstrated that behavioral features substantially improved predictive performance}}};
\node[anchor=base west,inner sep=0pt,outer sep=0pt] at ([xshift=82.104pt,yshift=-509.590pt]current page.north west) {{\fontsize{11.040pt}{11.702pt}\selectfont  \makebox[340.01pt][s]{and agent-specific features provided additional statistically significant gains.}}};
\node[anchor=base west,inner sep=0pt,outer sep=0pt] at ([xshift=82.104pt,yshift=-534.910pt]current page.north west) {{\fontsize{11.040pt}{11.702pt}\selectfont  \makebox[433.71pt][s]{Evaluation across five simulated agent-specific attack scenarios showed that ACAGS effectively}}};
\node[anchor=base west,inner sep=0pt,outer sep=0pt] at ([xshift=82.104pt,yshift=-547.630pt]current page.north west) {{\fontsize{11.040pt}{11.702pt}\selectfont  \makebox[433.98pt][s]{identified explicit policy violations and abnormal agent activity, although compromised-agent}}};
\node[anchor=base west,inner sep=0pt,outer sep=0pt] at ([xshift=82.104pt,yshift=-560.230pt]current page.north west) {{\fontsize{11.040pt}{11.702pt}\selectfont  \makebox[433.63pt][s]{attacks remained comparatively difficult to detect. The authorization mechanism further}}};
\node[anchor=base west,inner sep=0pt,outer sep=0pt] at ([xshift=82.104pt,yshift=-572.830pt]current page.north west) {{\fontsize{11.040pt}{11.702pt}\selectfont  \makebox[433.57pt][s]{demonstrated the value of distinguishing uncertain behavioral risk from explicit authorization}}};
\node[anchor=base west,inner sep=0pt,outer sep=0pt] at ([xshift=82.104pt,yshift=-585.550pt]current page.north west) {{\fontsize{11.040pt}{11.702pt}\selectfont  \makebox[83.84pt][s]{violations through\ }}};
\node[anchor=base west,inner sep=0pt,outer sep=0pt] at ([xshift=169.200pt,yshift=-585.550pt]current page.north west) {{\fontsize{11.040pt}{11.702pt}\selectfont \bfseries \makebox[173.25pt][s]{APPROVE, REVIEW, and REJECT}}};
\node[anchor=base west,inner sep=0pt,outer sep=0pt] at ([xshift=344.550pt,yshift=-585.550pt]current page.north west) {{\fontsize{11.040pt}{11.702pt}\selectfont  \makebox[171.44pt][s]{\hspace{1pt}outcomes. These findings support the}}};
\node[anchor=base west,inner sep=0pt,outer sep=0pt] at ([xshift=82.104pt,yshift=-598.150pt]current page.north west) {{\fontsize{11.040pt}{11.702pt}\selectfont  \makebox[433.90pt][s]{potential of combining behavioral risk assessment with delegated-authorization controls to}}};
\node[anchor=base west,inner sep=0pt,outer sep=0pt] at ([xshift=82.104pt,yshift=-610.870pt]current page.north west) {{\fontsize{11.040pt}{11.702pt}\selectfont  \makebox[356.21pt][s]{strengthen cybersecurity in emerging agentic-commerce payment environments.}}};
\node[anchor=base west,inner sep=0pt,outer sep=0pt] at ([xshift=82.104pt,yshift=-636.070pt]current page.north west) {{\fontsize{11.040pt}{11.702pt}\selectfont  \makebox[433.62pt][s]{The results should be interpreted within the study's controlled simulation setting rather than as}}};
\node[anchor=base west,inner sep=0pt,outer sep=0pt] at ([xshift=82.104pt,yshift=-648.790pt]current page.north west) {{\fontsize{11.040pt}{11.702pt}\selectfont  \makebox[433.63pt][s]{evidence of production deployment performance. Future work should evaluate ACAGS using}}};
\node[anchor=base west,inner sep=0pt,outer sep=0pt] at ([xshift=82.104pt,yshift=-661.420pt]current page.north west) {{\fontsize{11.040pt}{11.702pt}\selectfont  \makebox[433.68pt][s]{real-world agentic transaction data, investigate adaptive and coordinated agent attacks, optimize}}};
\node[anchor=base west,inner sep=0pt,outer sep=0pt] at ([xshift=82.104pt,yshift=-674.140pt]current page.north west) {{\fontsize{11.040pt}{11.702pt}\selectfont  \makebox[433.65pt][s]{authorization thresholds, and assess latency, scalability, and integration with production payment}}};
\node[anchor=base west,inner sep=0pt,outer sep=0pt] at ([xshift=82.104pt,yshift=-686.740pt]current page.north west) {{\fontsize{11.040pt}{11.702pt}\selectfont  infrastructure.}};
\end{tikzpicture}
\null
\newpage
\begin{tikzpicture}[remember picture,overlay]
\node[anchor=base west,inner sep=0pt,outer sep=0pt] at ([xshift=127.580pt,yshift=-81.360pt]current page.north west) {{\fontsize{9.960pt}{10.558pt}\selectfont  \makebox[342.77pt][s]{International Journal on Cybernetics \& Informatics (IJCI) Vol.15, No.4, August 2026}}};
\node[anchor=base west,inner sep=0pt,outer sep=0pt] at ([xshift=498.340pt,yshift=-767.856pt]current page.north west) {{\fontsize{9.960pt}{10.558pt}\selectfont  101}};
\node[anchor=base west,inner sep=0pt,outer sep=0pt] at ([xshift=82.104pt,yshift=-101.180pt]current page.north west) {{\fontsize{12.000pt}{12.720pt}\selectfont \bfseries R}};
\node[anchor=base west,inner sep=0pt,outer sep=0pt] at ([xshift=90.744pt,yshift=-101.180pt]current page.north west) {{\fontsize{9.480pt}{10.049pt}\selectfont \bfseries EFERENCES}};
\node[anchor=base west,inner sep=0pt,outer sep=0pt] at ([xshift=82.104pt,yshift=-125.780pt]current page.north west) {{\fontsize{9.960pt}{10.558pt}\selectfont  [1]}};
\node[anchor=base west,inner sep=0pt,outer sep=0pt] at ([xshift=104.660pt,yshift=-125.780pt]current page.north west) {{\fontsize{9.960pt}{10.558pt}\selectfont  OpenAI,}};
\node[anchor=base west,inner sep=0pt,outer sep=0pt] at ([xshift=158.005pt,yshift=-125.780pt]current page.north west) {{\fontsize{9.960pt}{10.558pt}\selectfont  "Introducing}};
\node[anchor=base west,inner sep=0pt,outer sep=0pt] at ([xshift=227.157pt,yshift=-125.780pt]current page.north west) {{\fontsize{9.960pt}{10.558pt}\selectfont  Operator,"}};
\node[anchor=base west,inner sep=0pt,outer sep=0pt] at ([xshift=287.933pt,yshift=-125.780pt]current page.north west) {{\fontsize{9.960pt}{10.558pt}\selectfont  OpenAI,}};
\node[anchor=base west,inner sep=0pt,outer sep=0pt] at ([xshift=341.159pt,yshift=-125.780pt]current page.north west) {{\fontsize{9.960pt}{10.558pt}\selectfont  Jan.}};
\node[anchor=base west,inner sep=0pt,outer sep=0pt] at ([xshift=375.451pt,yshift=-125.780pt]current page.north west) {{\fontsize{9.960pt}{10.558pt}\selectfont  2025.}};
\node[anchor=base west,inner sep=0pt,outer sep=0pt] at ([xshift=416.437pt,yshift=-125.780pt]current page.north west) {{\fontsize{9.960pt}{10.558pt}\selectfont  [Online].}};
\node[anchor=base west,inner sep=0pt,outer sep=0pt] at ([xshift=471.326pt,yshift=-125.780pt]current page.north west) {{\fontsize{9.960pt}{10.558pt}\selectfont  Available:}};
\node[anchor=base west,inner sep=0pt,outer sep=0pt] at ([xshift=104.660pt,yshift=-137.300pt]current page.north west) {{\fontsize{9.960pt}{10.558pt}\selectfont  \makebox[282.86pt][s]{https://openai.com/index/introducing-operator/. Accessed: Aug. 2026.}}};
\node[anchor=base west,inner sep=0pt,outer sep=0pt] at ([xshift=82.104pt,yshift=-148.820pt]current page.north west) {{\fontsize{9.960pt}{10.558pt}\selectfont  [2]}};
\node[anchor=base west,inner sep=0pt,outer sep=0pt] at ([xshift=104.660pt,yshift=-148.820pt]current page.north west) {{\fontsize{9.960pt}{10.558pt}\selectfont  Anthropic,}};
\node[anchor=base west,inner sep=0pt,outer sep=0pt] at ([xshift=159.311pt,yshift=-148.820pt]current page.north west) {{\fontsize{9.960pt}{10.558pt}\selectfont  "Introducing}};
\node[anchor=base west,inner sep=0pt,outer sep=0pt] at ([xshift=221.401pt,yshift=-148.820pt]current page.north west) {{\fontsize{9.960pt}{10.558pt}\selectfont  Computer}};
\node[anchor=base west,inner sep=0pt,outer sep=0pt] at ([xshift=272.944pt,yshift=-148.820pt]current page.north west) {{\fontsize{9.960pt}{10.558pt}\selectfont  Use,"}};
\node[anchor=base west,inner sep=0pt,outer sep=0pt] at ([xshift=306.529pt,yshift=-148.820pt]current page.north west) {{\fontsize{9.960pt}{10.558pt}\selectfont  Anthropic,}};
\node[anchor=base west,inner sep=0pt,outer sep=0pt] at ([xshift=361.060pt,yshift=-148.820pt]current page.north west) {{\fontsize{9.960pt}{10.558pt}\selectfont  Oct.}};
\node[anchor=base west,inner sep=0pt,outer sep=0pt] at ([xshift=389.486pt,yshift=-148.820pt]current page.north west) {{\fontsize{9.960pt}{10.558pt}\selectfont  2024.}};
\node[anchor=base west,inner sep=0pt,outer sep=0pt] at ([xshift=423.510pt,yshift=-148.820pt]current page.north west) {{\fontsize{9.960pt}{10.558pt}\selectfont  [Online].}};
\node[anchor=base west,inner sep=0pt,outer sep=0pt] at ([xshift=471.338pt,yshift=-148.820pt]current page.north west) {{\fontsize{9.960pt}{10.558pt}\selectfont  Available:}};
\node[anchor=base west,inner sep=0pt,outer sep=0pt] at ([xshift=104.660pt,yshift=-160.340pt]current page.north west) {{\fontsize{9.960pt}{10.558pt}\selectfont  \makebox[352.01pt][s]{https://www.anthropic.com/news/3-5-models-and-computer-use. Accessed: Aug. 2026.}}};
\node[anchor=base west,inner sep=0pt,outer sep=0pt] at ([xshift=82.104pt,yshift=-179.060pt]current page.north west) {{\fontsize{9.960pt}{10.558pt}\selectfont  [3]}};
\node[anchor=base west,inner sep=0pt,outer sep=0pt] at ([xshift=104.660pt,yshift=-179.060pt]current page.north west) {{\fontsize{9.960pt}{10.558pt}\selectfont  \makebox[410.97pt][s]{S. Davidovic and H. Tourpe, How Agentic AI Will Reshape Payments, IMF Notes, no. 2026/004,}}};
\node[anchor=base west,inner sep=0pt,outer sep=0pt] at ([xshift=104.660pt,yshift=-190.580pt]current page.north west) {{\fontsize{9.960pt}{10.558pt}\selectfont  \makebox[404.09pt][s]{International Monetary Fund, Washington, DC, USA, Apr. 2026, doi: 10.5089/9781513533308.068.}}};
\node[anchor=base west,inner sep=0pt,outer sep=0pt] at ([xshift=82.104pt,yshift=-243.740pt]current page.north west) {{\fontsize{9.960pt}{10.558pt}\selectfont  [5]}};
\node[anchor=base west,inner sep=0pt,outer sep=0pt] at ([xshift=104.660pt,yshift=-243.740pt]current page.north west) {{\fontsize{9.960pt}{10.558pt}\selectfont  \makebox[330.38pt][s]{L. Breiman, "Random Forests," Machine Learning, vol. 45, no. 1, pp. 5--32, 2001.}}};
\node[anchor=base west,inner sep=0pt,outer sep=0pt] at ([xshift=82.104pt,yshift=-255.140pt]current page.north west) {{\fontsize{9.960pt}{10.558pt}\selectfont  [6]}};
\node[anchor=base west,inner sep=0pt,outer sep=0pt] at ([xshift=104.660pt,yshift=-255.140pt]current page.north west) {{\fontsize{9.960pt}{10.558pt}\selectfont  \makebox[411.01pt][s]{T. Chen and C. Guestrin, "XGBoost: A Scalable Tree Boosting System," Proc. 22nd ACM SIGKDD}}};
\node[anchor=base west,inner sep=0pt,outer sep=0pt] at ([xshift=104.660pt,yshift=-266.660pt]current page.north west) {{\fontsize{9.960pt}{10.558pt}\selectfont  \makebox[286.10pt][s]{Int. Conf. Knowledge Discovery and Data Mining, pp. 785--794, 2016.}}};
\node[anchor=base west,inner sep=0pt,outer sep=0pt] at ([xshift=82.104pt,yshift=-278.180pt]current page.north west) {{\fontsize{9.960pt}{10.558pt}\selectfont  [7]}};
\node[anchor=base west,inner sep=0pt,outer sep=0pt] at ([xshift=104.660pt,yshift=-278.180pt]current page.north west) {{\fontsize{9.960pt}{10.558pt}\selectfont  \makebox[411.07pt][s]{G. Keet al., "LightGBM: A Highly Efficient Gradient Boosting Decision Tree," Advances in Neural}}};
\node[anchor=base west,inner sep=0pt,outer sep=0pt] at ([xshift=104.660pt,yshift=-289.730pt]current page.north west) {{\fontsize{9.960pt}{10.558pt}\selectfont  \makebox[192.24pt][s]{Information Processing Systems, vol. 30, 2017.}}};
\node[anchor=base west,inner sep=0pt,outer sep=0pt] at ([xshift=82.104pt,yshift=-301.250pt]current page.north west) {{\fontsize{9.960pt}{10.558pt}\selectfont  [8]}};
\node[anchor=base west,inner sep=0pt,outer sep=0pt] at ([xshift=104.660pt,yshift=-301.250pt]current page.north west) {{\fontsize{9.960pt}{10.558pt}\selectfont  \makebox[411.13pt][s]{S. M. Lundberg and S.-I. Lee, "A Unified Approach to Interpreting Model Predictions," Advances in}}};
\node[anchor=base west,inner sep=0pt,outer sep=0pt] at ([xshift=104.660pt,yshift=-312.770pt]current page.north west) {{\fontsize{9.960pt}{10.558pt}\selectfont  \makebox[222.02pt][s]{Neural Information Processing Systems, vol. 30, 2017.}}};
\node[anchor=base west,inner sep=0pt,outer sep=0pt] at ([xshift=82.104pt,yshift=-381.770pt]current page.north west) {{\fontsize{9.960pt}{10.558pt}\selectfont  [11]}};
\node[anchor=base west,inner sep=0pt,outer sep=0pt] at ([xshift=104.660pt,yshift=-381.770pt]current page.north west) {{\fontsize{9.960pt}{10.558pt}\selectfont  \makebox[411.17pt][s]{K. See and X. W. Tan, "Compliance-Aware Agentic Payments on Stablecoin Rails," arXiv preprint}}};
\node[anchor=base west,inner sep=0pt,outer sep=0pt] at ([xshift=104.660pt,yshift=-393.170pt]current page.north west) {{\fontsize{9.960pt}{10.558pt}\selectfont  \makebox[103.08pt][s]{arXiv:2605.00071, 2026.}}};
\node[anchor=base west,inner sep=0pt,outer sep=0pt] at ([xshift=82.104pt,yshift=-404.690pt]current page.north west) {{\fontsize{9.960pt}{10.558pt}\selectfont  [12]}};
\node[anchor=base west,inner sep=0pt,outer sep=0pt] at ([xshift=104.660pt,yshift=-404.690pt]current page.north west) {{\fontsize{9.960pt}{10.558pt}\selectfont  \makebox[411.03pt][s]{A. de Valois-Franklin and A. Bogdan, "RAILS: Verification-Native Clearing for Agentic}}};
\node[anchor=base west,inner sep=0pt,outer sep=0pt] at ([xshift=104.660pt,yshift=-416.210pt]current page.north west) {{\fontsize{9.960pt}{10.558pt}\selectfont  \makebox[215.42pt][s]{Commerce," arXiv preprint arXiv:2606.08790, 2026.}}};
\node[anchor=base west,inner sep=0pt,outer sep=0pt] at ([xshift=82.104pt,yshift=-427.730pt]current page.north west) {{\fontsize{9.960pt}{10.558pt}\selectfont  [13]}};
\node[anchor=base west,inner sep=0pt,outer sep=0pt] at ([xshift=104.660pt,yshift=-427.730pt]current page.north west) {{\fontsize{9.960pt}{10.558pt}\selectfont  \makebox[410.69pt][s]{S. Fan et al., "Agentic Commerce World: An Auditable and Verifiable Environment for Vibe}}};
\node[anchor=base west,inner sep=0pt,outer sep=0pt] at ([xshift=104.660pt,yshift=-439.250pt]current page.north west) {{\fontsize{9.960pt}{10.558pt}\selectfont  \makebox[215.42pt][s]{Commerce," arXiv preprint arXiv:2608.02441, 2026.}}};
\node[anchor=base west,inner sep=0pt,outer sep=0pt] at ([xshift=82.104pt,yshift=-450.770pt]current page.north west) {{\fontsize{9.960pt}{10.558pt}\selectfont  [14]}};
\node[anchor=base west,inner sep=0pt,outer sep=0pt] at ([xshift=104.660pt,yshift=-450.770pt]current page.north west) {{\fontsize{9.960pt}{10.558pt}\selectfont  \makebox[411.03pt][s]{E. Tabassi, Artificial Intelligence Risk Management Framework (AI RMF 1.0), National Institute of}}};
\node[anchor=base west,inner sep=0pt,outer sep=0pt] at ([xshift=104.660pt,yshift=-462.170pt]current page.north west) {{\fontsize{9.960pt}{10.558pt}\selectfont  \makebox[411.02pt][s]{Standards and Technology, Gaithersburg, MD, USA, NIST AI 100-1, Jan. 2023, doi:}}};
\node[anchor=base west,inner sep=0pt,outer sep=0pt] at ([xshift=104.660pt,yshift=-473.710pt]current page.north west) {{\fontsize{9.960pt}{10.558pt}\selectfont  10.6028/NIST.AI.100-1.}};
\node[anchor=base west,inner sep=0pt,outer sep=0pt] at ([xshift=82.104pt,yshift=-485.230pt]current page.north west) {{\fontsize{9.960pt}{10.558pt}\selectfont  [15]}};
\node[anchor=base west,inner sep=0pt,outer sep=0pt] at ([xshift=104.660pt,yshift=-485.230pt]current page.north west) {{\fontsize{9.960pt}{10.558pt}\selectfont  \makebox[410.97pt][s]{OWASP Foundation, OWASP Top 10 for LLM Applications 2025, Version 2025, Nov. 2024.}}};
\node[anchor=base west,inner sep=0pt,outer sep=0pt] at ([xshift=104.660pt,yshift=-496.750pt]current page.north west) {{\fontsize{9.960pt}{10.558pt}\selectfont  [Online].}};
\node[anchor=base west,inner sep=0pt,outer sep=0pt] at ([xshift=156.552pt,yshift=-496.750pt]current page.north west) {{\fontsize{9.960pt}{10.558pt}\selectfont  Available:}};
\node[anchor=base west,inner sep=0pt,outer sep=0pt] at ([xshift=213.770pt,yshift=-496.750pt]current page.north west) {{\fontsize{9.960pt}{10.558pt}\selectfont  https://genai.owasp.org/resource/owasp-top-10-for-llm-applications-2025/.}};
\node[anchor=base west,inner sep=0pt,outer sep=0pt] at ([xshift=104.660pt,yshift=-508.270pt]current page.north west) {{\fontsize{9.960pt}{10.558pt}\selectfont  \makebox[90.24pt][s]{Accessed: Aug. 2026.}}};
\node[anchor=base west,inner sep=0pt,outer sep=0pt] at ([xshift=82.104pt,yshift=-519.670pt]current page.north west) {{\fontsize{9.960pt}{10.558pt}\selectfont  [16]}};
\node[anchor=base west,inner sep=0pt,outer sep=0pt] at ([xshift=104.660pt,yshift=-519.670pt]current page.north west) {{\fontsize{9.960pt}{10.558pt}\selectfont  \makebox[411.10pt][s]{ACI Worldwide, Agentic Commerce: The Payments Shift Merchants Can't Ignore, White Paper,}}};
\node[anchor=base west,inner sep=0pt,outer sep=0pt] at ([xshift=104.660pt,yshift=-531.190pt]current page.north west) {{\fontsize{9.960pt}{10.558pt}\selectfont  2026.}};
\node[anchor=base west,inner sep=0pt,outer sep=0pt] at ([xshift=82.10pt,yshift=-209.20pt]current page.north west) {\fontsize{9.96pt}{10pt}\selectfont [4]};
\node[anchor=base west,inner sep=0pt,outer sep=0pt] at ([xshift=104.66pt,yshift=-209.20pt]current page.north west) {\fontsize{9.96pt}{10pt}\selectfont Visa, ``Agentic Payments: What Onchain Data Reveals Commerce, Jul. 14, 2026. [Online].};
\node[anchor=base west,inner sep=0pt,outer sep=0pt] at ([xshift=104.66pt,yshift=-220.70pt]current page.north west) {\fontsize{9.96pt}{10pt}\selectfont Available: https://www.visa.com/en-us/thought-leadership/innovation/agentic-payments-from-the-};
\node[anchor=base west,inner sep=0pt,outer sep=0pt] at ([xshift=104.66pt,yshift=-232.20pt]current page.north west) {\fontsize{9.96pt}{10pt}\selectfont ground-up. Accessed: Jul. 22, 2026.};
\node[anchor=base west,inner sep=0pt,outer sep=0pt] at ([xshift=82.10pt,yshift=-324.20pt]current page.north west) {\fontsize{9.96pt}{10pt}\selectfont [9]};
\node[anchor=base west,inner sep=0pt,outer sep=0pt] at ([xshift=104.66pt,yshift=-324.20pt]current page.north west) {\fontsize{9.96pt}{10pt}\selectfont A. Dal Pozzolo, O. Caelen, Y.-A. Le Borgne, S. Waterschoot, and G. Bontempi, ``Learned Lessons in};
\node[anchor=base west,inner sep=0pt,outer sep=0pt] at ([xshift=104.66pt,yshift=-335.70pt]current page.north west) {\fontsize{9.96pt}{10pt}\selectfont Credit Card Fraud Detection from a Practitioner Perspective, Expert Systems with Applications, vol.};
\node[anchor=base west,inner sep=0pt,outer sep=0pt] at ([xshift=104.66pt,yshift=-347.20pt]current page.north west) {\fontsize{9.96pt}{10pt}\selectfont 41, no. 10, pp. 4915--4928, 2014, doi: 10.1016/j.eswa.2014.02.026.};
\node[anchor=base west,inner sep=0pt,outer sep=0pt] at ([xshift=82.10pt,yshift=-358.70pt]current page.north west) {\fontsize{9.96pt}{10pt}\selectfont [10]};
\node[anchor=base west,inner sep=0pt,outer sep=0pt] at ([xshift=104.66pt,yshift=-358.70pt]current page.north west) {\fontsize{9.96pt}{10pt}\selectfont A. Carcillo et al., ``Combining Unsupervised and Supervised Learning in Credit Card Fraud};
\node[anchor=base west,inner sep=0pt,outer sep=0pt] at ([xshift=104.66pt,yshift=-370.20pt]current page.north west) {\fontsize{9.96pt}{10pt}\selectfont Detection, Information Sciences, vol. 557, pp. 317--331, 2021, doi: 10.1016/j.ins.2019.05.042.};
\node[anchor=base west,inner sep=0pt,outer sep=0pt] at ([xshift=82.10pt,yshift=-542.70pt]current page.north west) {\fontsize{9.96pt}{10pt}\selectfont [17]};
\node[anchor=base west,inner sep=0pt,outer sep=0pt] at ([xshift=104.66pt,yshift=-542.70pt]current page.north west) {\fontsize{9.96pt}{10pt}\selectfont K. Singh, ``Credit Card Transactions Fraud Detection Dataset, Kaggle. [Online]. Available:};
\node[anchor=base west,inner sep=0pt,outer sep=0pt] at ([xshift=104.66pt,yshift=-554.20pt]current page.north west) {\fontsize{9.96pt}{10pt}\selectfont https://www.kaggle.com/datasets/kartik2112/fraud-detection. Accessed: Jul. 25, 2026.};
\node[anchor=base west,inner sep=0pt,outer sep=0pt] at ([xshift=82.10pt,yshift=-565.80pt]current page.north west) {\fontsize{9.96pt}{10pt}\selectfont [18]};
\node[anchor=base west,inner sep=0pt,outer sep=0pt] at ([xshift=104.66pt,yshift=-565.80pt]current page.north west) {\fontsize{9.96pt}{10pt}\selectfont F. Rakhmatov and N. Karimov, ``Anomaly Detection in Network Traffic Using Selected Statistical};
\node[anchor=base west,inner sep=0pt,outer sep=0pt] at ([xshift=104.66pt,yshift=-577.30pt]current page.north west) {\fontsize{9.96pt}{10pt}\selectfont and Entropy-Based Features, International Journal on Cybernetics \& Informatics, vol. 14, no. 5, pp.};
\node[anchor=base west,inner sep=0pt,outer sep=0pt] at ([xshift=104.66pt,yshift=-588.70pt]current page.north west) {\fontsize{9.96pt}{10pt}\selectfont 1--12, 2025, doi: 10.5121/ijci.2025.140501.};
\node[anchor=base west,inner sep=0pt,outer sep=0pt] at ([xshift=82.10pt,yshift=-600.20pt]current page.north west) {\fontsize{9.96pt}{10pt}\selectfont [19]};
\node[anchor=base west,inner sep=0pt,outer sep=0pt] at ([xshift=104.66pt,yshift=-600.20pt]current page.north west) {\fontsize{9.96pt}{10pt}\selectfont A. H. Wheeb and M. F. Khan, ``A Survey of Several Machine Learning (ML) Algorithms for Security};
\node[anchor=base west,inner sep=0pt,outer sep=0pt] at ([xshift=104.66pt,yshift=-611.70pt]current page.north west) {\fontsize{9.96pt}{10pt}\selectfont Solution in Internet of Things (IoT) Networks, Journal of Artificial Intelligence Research \&};
\node[anchor=base west,inner sep=0pt,outer sep=0pt] at ([xshift=104.66pt,yshift=-623.20pt]current page.north west) {\fontsize{9.96pt}{10pt}\selectfont Advances, vol. 12, no. 1, pp. 1--11, 2025.};
\end{tikzpicture}
\null
\newpage
\begin{tikzpicture}[remember picture,overlay]
\node[anchor=base west,inner sep=0pt,outer sep=0pt] at ([xshift=127.580pt,yshift=-81.360pt]current page.north west) {{\fontsize{9.960pt}{10.558pt}\selectfont  \makebox[342.77pt][s]{International Journal on Cybernetics \& Informatics (IJCI) Vol.15, No.4, August 2026}}};
\node[anchor=base west,inner sep=0pt,outer sep=0pt] at ([xshift=498.340pt,yshift=-767.856pt]current page.north west) {{\fontsize{9.960pt}{10.558pt}\selectfont  102}};
\node[anchor=base west,inner sep=0pt,outer sep=0pt] at ([xshift=82.104pt,yshift=-101.180pt]current page.north west) {{\fontsize{12.000pt}{12.720pt}\selectfont \bfseries A}};
\node[anchor=base west,inner sep=0pt,outer sep=0pt] at ([xshift=90.744pt,yshift=-101.180pt]current page.north west) {{\fontsize{9.480pt}{10.049pt}\selectfont \bfseries UTHOR}};
\node[anchor=north west,inner sep=0pt,outer sep=0pt] at ([xshift=430.700pt,yshift=-117.000pt]current page.north west) {\includegraphics[width=83.100pt,height=88.200pt]{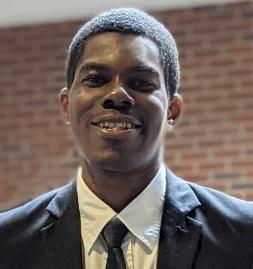}};
\node[anchor=base west,inner sep=0pt,outer sep=0pt] at ([xshift=82.10pt,yshift=-127.90pt]current page.north west) {\fontsize{9.96pt}{10pt}\selectfont \textbf{Kayode Lateef Ogunsusi} holds a bachelor's degree in Industrial Mathematics, a};
\node[anchor=base west,inner sep=0pt,outer sep=0pt] at ([xshift=82.10pt,yshift=-139.40pt]current page.north west) {\fontsize{9.96pt}{10pt}\selectfont master's degree in Information Technology (Nigeria), and a master's degree in};
\node[anchor=base west,inner sep=0pt,outer sep=0pt] at ([xshift=82.10pt,yshift=-150.90pt]current page.north west) {\fontsize{9.96pt}{10pt}\selectfont Computer Science and Quantitative Methods from Austin Peay State University,};
\node[anchor=base west,inner sep=0pt,outer sep=0pt] at ([xshift=82.10pt,yshift=-162.50pt]current page.north west) {\fontsize{9.96pt}{10pt}\selectfont USA. He is a product and risk management professional with experience in fraud};
\node[anchor=base west,inner sep=0pt,outer sep=0pt] at ([xshift=82.10pt,yshift=-174.00pt]current page.north west) {\fontsize{9.96pt}{10pt}\selectfont mitigation, data analytics, and financial decision systems. His research interests};
\node[anchor=base west,inner sep=0pt,outer sep=0pt] at ([xshift=82.10pt,yshift=-185.40pt]current page.north west) {\fontsize{9.96pt}{10pt}\selectfont include artificial intelligence, machine learning, agentic AI, financial risk};
\node[anchor=base west,inner sep=0pt,outer sep=0pt] at ([xshift=82.10pt,yshift=-196.90pt]current page.north west) {\fontsize{9.96pt}{10pt}\selectfont management, and intelligent decision systems. He is currently a credit and fraud};
\node[anchor=base west,inner sep=0pt,outer sep=0pt] at ([xshift=82.10pt,yshift=-208.50pt]current page.north west) {\fontsize{9.96pt}{10pt}\selectfont risk manager at American Express, where he uses his expertise in data analytics and};
\node[anchor=base west,inner sep=0pt,outer sep=0pt] at ([xshift=82.10pt,yshift=-220.00pt]current page.north west) {\fontsize{9.96pt}{10pt}\selectfont advanced risk strategies to enhance solutions that strengthen fraud risk management within the financial};
\node[anchor=base west,inner sep=0pt,outer sep=0pt] at ([xshift=82.10pt,yshift=-231.50pt]current page.north west) {\fontsize{9.96pt}{10pt}\selectfont services industry.};
\end{tikzpicture}
\null
\end{document}